\documentclass[10pt,twocolumn,letterpaper]{article}
\usepackage[pagenumbers]{cvpr} 

\usepackage{graphicx}
\usepackage{amsmath}
\usepackage{amssymb}
\usepackage{booktabs}
\usepackage{enumitem}
\usepackage{colortbl}
\usepackage{color, xcolor}
\usepackage{multirow}
\usepackage{makecell}
\usepackage{array}
\usepackage{arydshln}
\usepackage{adjustbox}
\usepackage{nicefrac}
\usepackage[accsupp]{axessibility} 

\usepackage[pagebackref,breaklinks,colorlinks]{hyperref}

\usepackage[capitalize]{cleveref}
\crefname{section}{Sec.}{Secs.}
\Crefname{section}{Section}{Sections}
\Crefname{table}{Table}{Tables}
\crefname{table}{Tab.}{Tabs.}

\def\confName{CVPR}
\def\confYear{2023}

\definecolor{color3}{rgb}{0.95,0.95,0.95}

\begin{document}

\title{Masked Image Training for Generalizable Deep Image Denoising}

\author{
Haoyu Chen$^{1*}$,\quad Jinjin Gu$^{2,3}$\thanks{Haoyu Chen and Jinjin Gu contribute equally to this work.},\quad Yihao Liu$^{2,4,5}$,\quad Salma Abdel Magid$^{6}$,\\ \vspace{1mm} Chao Dong$^{2,4}$,\quad Qiong Wang$^{4}$,\quad Hanspeter Pfister$^{6}$,\quad Lei Zhu$^{1,7}$\thanks{Lei Zhu (leizhu@ust.hk) is the corresponding author.}\\ \vspace{-0.5mm}
\small $^{1}$The Hong Kong University of Science and Technology (Guangzhou)\quad
\small $^{2}$Shanghai AI Lab\quad $^{3}$The University of Sydney\\ 
\small $^{4}$Shenzhen Institutes of Advanced Technology, Chinese Academy of Sciences\quad
\small $^{5}$University of Chinese Academy of Sciences\\ 
\small $^{6}$Harvard University\quad
\small $^{7}$The Hong Kong University of Science and Technology\\
{\tt\small Project page: \url{https://github.com/haoyuc/MaskedDenoising}}}

\maketitle


\begin{abstract}
When capturing and storing images, devices inevitably introduce noise.
Reducing this noise is a critical task called image denoising.
Deep learning has become the de facto method for image denoising, especially with the emergence of Transformer-based models that have achieved notable state-of-the-art results on various image tasks.
However, deep learning-based methods often suffer from a lack of generalization ability.
For example, deep models trained on Gaussian noise may perform poorly when tested on other noise distributions.
To address this issue, we present a novel approach to enhance the generalization performance of denoising networks, known as masked training.
Our method involves masking random pixels of the input image and reconstructing the missing information during training.
We also mask out the features in the self-attention layers to avoid the impact of training-testing inconsistency.
Our approach exhibits better generalization ability than other deep learning models and is directly applicable to real-world scenarios.
Additionally, our interpretability analysis demonstrates the superiority of our method.

\end{abstract}

\vspace{-4mm}
\section{Introduction}
\label{Introduction}
\vspace{-1mm}
Image denoising is a crucial research area that aims to recover clean images from noisy observations.
Due to the rapid advancements in deep learning, many promising image denoising networks have been developed.
These networks are typically trained using images synthesized from a pre-defined noise distribution and can achieve remarkable performance in removing the corresponding noise.
However, a significant challenge in applying these deep models to real-world scenarios is their generalization ability.
Since the real-world noise distribution can differ from that observed during training, these models often struggle to generalize to such scenarios.

\begin{figure}[t]
  \centering
   \includegraphics[width=\linewidth]{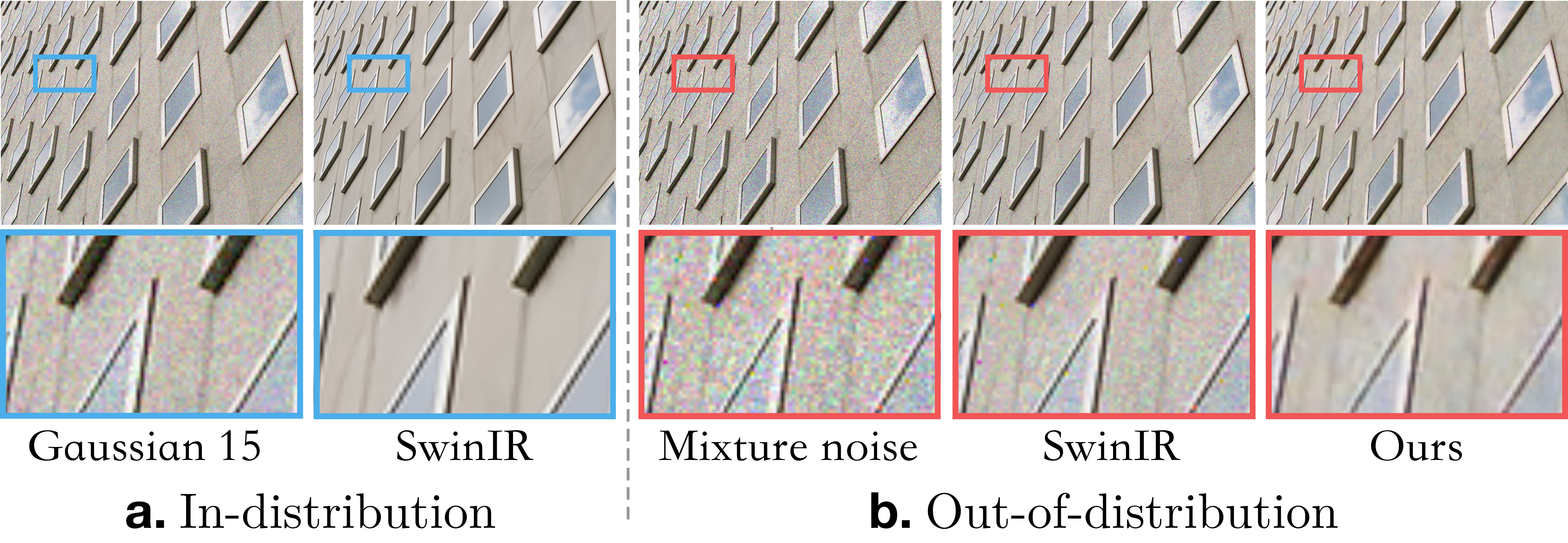}
   \vspace{-6mm}
   \caption{We illustrate the generalization problem of denoising networks. We train a SwinIR model on Gaussian noise with $\sigma=15$. When tested on the same noise, SwinIR demonstrates outstanding performance. However, when applied to out-of-distribution noise, \eg, the mixture of various noise. SwinIR suffers from a huge performance drop. The model trained by the proposed \emph{masked training} method maintains a reasonable denoising effect, despite aldo being trained on Gaussian noise.}
   \label{fig:teaser}
\vspace{-4mm}
\end{figure}

More specifically, most existing denoising works train and evaluate models on images corrupted with Gaussian noise, limiting their performance to a single noise distribution.
When these models are applied to remove noise drawn from other distributions, their performance drastically drops.
\figurename~\ref{fig:teaser} shows an example.
The research community has become increasingly aware of this generalization issue of deep models in recent years.
As a countermeasure, some methods \cite{zhang2017beyond} assume that the noise level of a particular noise type is unknown, while others \cite{brooks2019unprocessing,wei2020physics} attempt to improve the performance in real-world scenarios by synthesizing or collecting training data closer to the target noise or directly performing unsupervised training on the target noise \cite{chen2018image,yuan2018unsupervised}.
However, none of these methods substantially improve the generalization performance of denoising networks, and they still struggle when the noise distribution is mismatched \cite{abdelhamed2018high}.
The generalization issue of deep denoising still poses challenges to making these methods broadly applicable.

In this work, we focus on improving the generalization ability of deep denoising models.
We define generalization ability as the model's performance on noise different from what it observed during training.
We argue that the generalization issue of deep denoising is due to the overfitting of training noise.
The existing training strategy directly optimizes the similarity between the denoised image and the ground truth.
The intention behind this is that the network should learn to reconstruct the texture and semantics of natural images correctly.
However, what is often overlooked is that the network can also reduce the loss simply by overfitting the noise pattern, which is easier than learning the image content.
This is at the heart of the generalization problem.
Even many popular deep learning methods exacerbate this overfitting problem.
When it comes to noise different from that observed during training, the network exhibits this same behavior, resulting in poor performance.

In light of the preceding discussion, our study seeks to improve the generalization performance of deep denoising networks by directing them to learn image content reconstruction instead of overfitting to training noise.
Drawing inspiration from recent masked modeling methods \cite{devlin2018bert,he2022masked,xie2022simmim,bao2021beit}, we employ a masked training strategy to explicitly learn representations for image content reconstruction, as opposed to training noise.
Leveraging the properties of image processing Transformers \cite{liang2021swinir,zhang2022accurate,chen2022cross}, we introduce two masking mechanisms: the \emph{input mask} and the \emph{attention mask}.
During training, the input mask removes input image pixels randomly, and the network reconstructs the removed pixels.
The attention mask is implemented in each self-attention layer of the Transformer, enabling it to learn the completion of masked features dynamically and mitigate the distribution shift between training and testing in masked learning.
Although we use Gaussian noise for training -- similar to previous works -- our method demonstrates significant performance improvements on various noise types, such as speckle noise, Poisson noise, salt and pepper noise, spatially correlated Gaussian noise, Monte Carlo-rendered image noise, ISP noise, and complex mixtures of multiple noise sources.
Existing methods and models have yet to effectively and accurately remove all these diverse noise patterns.

\section{Related Works}
\label{RelatedWorks}

\noindent\textbf{Image Denoising}
approaches very broadly lie in two categories: traditional model-based and data-driven deep-learning-based.
Traditional methods are usually based on modeling image priors to recover image content contaminated by noise \cite{buades2005non,dabov2007image,gu2014weighted,mairal2009non,elad2006image}.
These methods usually do not impose too many constraints on the type of noise, and have been proven to be applicable to a variety of noise, with good generalization performance \cite{abdelhamed2018high}.
However, these methods are not satisfactory for the reconstruction of image content.
In recent years, the paradigm of denoising has gradually shifted to data-driven methods based on deep learning methods \cite{chen2016trainable}.
Many techniques have been proposed to improve the capabilities of the denoising networks continuously, \eg, residual networks \cite{zhang2017beyond,kokkinos2018deep,zhang2018ffdnet}, dense networks \cite{jia2019focnet,zhang2020residual}, recursive networks \cite{tai2017memnet,chen2018deep,liu2018non}, multi-scale \cite{divakar2017image,gu2019self,zamir2021multi}, encoder-decoder \cite{mao2016image,cheng2021nbnet,yue2020dual}, attention operations \cite{zhang2019residual,zhang2021accurate}, self-similarity \cite{hu2021pseudo}, and non-local operations \cite{lefkimmiatis2017non,lefkimmiatis2018universal,plotz2018neural}.
Since 2020, the paradigm of vision network design has gradually shifted from CNNs to Transformers \cite{dosovitskiy2020image}.
Vision Transformers treat input pixels as tokens and use self-attention operations to process interactions between these tokens.
Inspired by the success of vision Transformers, many attempts have been made to employ Transformers for low-level vision tasks \cite{yang2020learning,zamir2021restormer,chen2021pre,wang2021uformer,zhang2022accurate,chen2022cross,liang2021swinir,shi2022rethinking,chen2023recursive,zhang2023xformer}
During the development of these models, the noise pattern used for training is often consistent with the testing one.
The factor that determines its denoising performance is the fitting ability of the network, in other words, the ability of the network to overfit to the training noise.
However, a better network does not mean a better generalization ability of the denoising model.
As we will show in the experiment section, a more efficient network even indicates worse generalization performance.

\vspace{-4mm}
\paragraph{Generalization Problem}
in low-level vision often arises when the testing degradation does not match the training degradation, \eg, different downsampling kernel in super-resolution \cite{gu2019blind,liu2022blind,kong2022reflash}.
We typically develop deep denoising models based on Gaussian noise in the laboratory setting.
However, noise in the real-world is mostly non-Gaussian.
Models trained on Gaussian noise fail in these non-Gaussian scenarios.
There are two main categories of solutions to this problem.
The first is to make training datasets with noise modeling as close to reality as possible during development, \eg, synthesizing real noise according to physical system modeling \cite{brooks2019unprocessing,wei2020physics}, learning to generate real noise \cite{chen2018image,feng2019suppressing,yuan2018unsupervised}, collecting real noise -- clean image pairs for training \cite{plotz2017benchmarking,abdelhamed2018high,guo2019toward,krull2019noise2void}.
Although the models obtained by these methods can improve the effect on the target noise, they still cannot generalize to out-of-distribution noise.
Another category of solutions is to develop ``blind'' denoising models, which are supposed to deal with unknown noise \cite{zhang2017beyond,yue2019variational,krull2019noise2void}.
These methods usually simply assume that the noise level is unknown, or train on a large amount of noise types \cite{zhang2022practical}, which also fails to generalize to other noise not present in the training set.
Few workd have been proposed to study the reasons for the lack of generalization ability in low-level vision \cite{kong2022reflash}.
Liu \etal \cite{liu2021discovering} argue that networks tend to overfit to degradations and show degradation ``semantics'' inside the network.
The presence of these representations often means a decrease in generalization ability.
The utilization of this knowledge can guide us to analyze and evaluate the generalization performance \cite{liu2022evaluating}.
Apart from that, few works have been proposed to improve the generalization ability of denoising models.

\vspace{-4mm}
\paragraph{Masked modeling}
for language \cite{devlin2018bert,radford2018improving,radford2019language,brown2020language} is successful for learning pre-trained representations that generalize well to various downstream tasks.
These methods mask out a portion of the input sequence and train models to predict the missing content.
A similar approach can also be applied to the vision model pre-training.
Masked image models learn representations from corrupted images. The earliest attempts in this regard can be traced back at least to the denoising auto-encoder \cite{vincent2010stacked}.
Since then, many works have used predicting missing parts of images to learn efficient image representations \cite{pathak2016context,bao2021beit,chen2020generative,he2022masked,xie2022simmim}.
However, there have been few successful attempts to apply masked image modeling to low-level vision, even though the masked pre-training method is in the form of low-level vision tasks.

\begin{figure}[t]
  \centering
   \includegraphics[width=\linewidth]{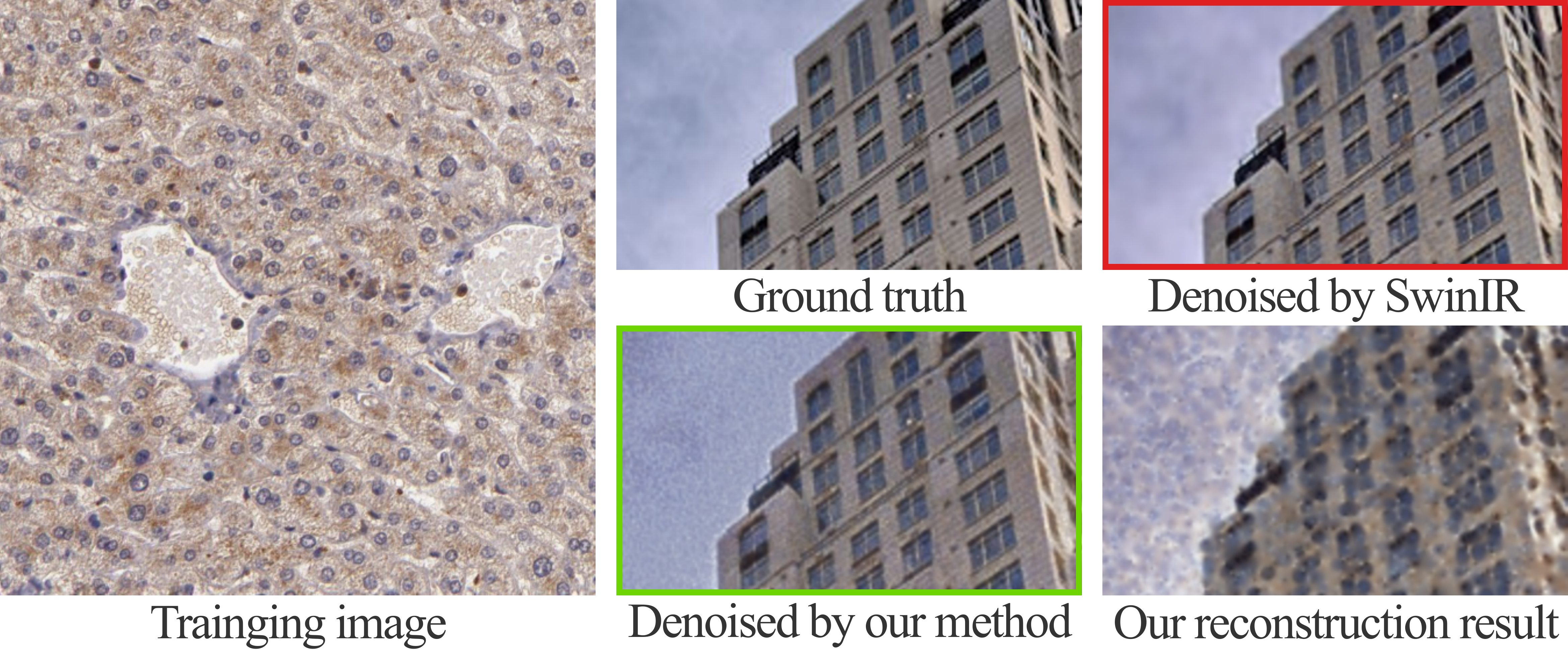}
   \vspace{-5mm}
   \caption{
   SwinIR, when trained solely on immunohistochemistry images with Gaussian noise, can still denoise natural images. This observation supports the assertion that most existing methods perform denoising primarily through overfitting the training noise. In contrast, our approach emphasizes reconstructing natural image textures and edges observed in the training set on natural images, rather than relying on noise overfitting for denoising. This distinction underlines the fundamental difference between our method and previous approaches. ``Our reconstruction result'' refers to using our model but taking masked images as input.
   }
   \label{fig:motivation}
   \vspace{-4mm}
\end{figure}

\section{Method}
Our objective is to create denoising models capable of generalizing to noise not encountered in the training set. In this section, we first discuss our motivation before delving into the specifics of our masked training method.

\begin{figure}[t]
  \centering
   \includegraphics[width=\linewidth]{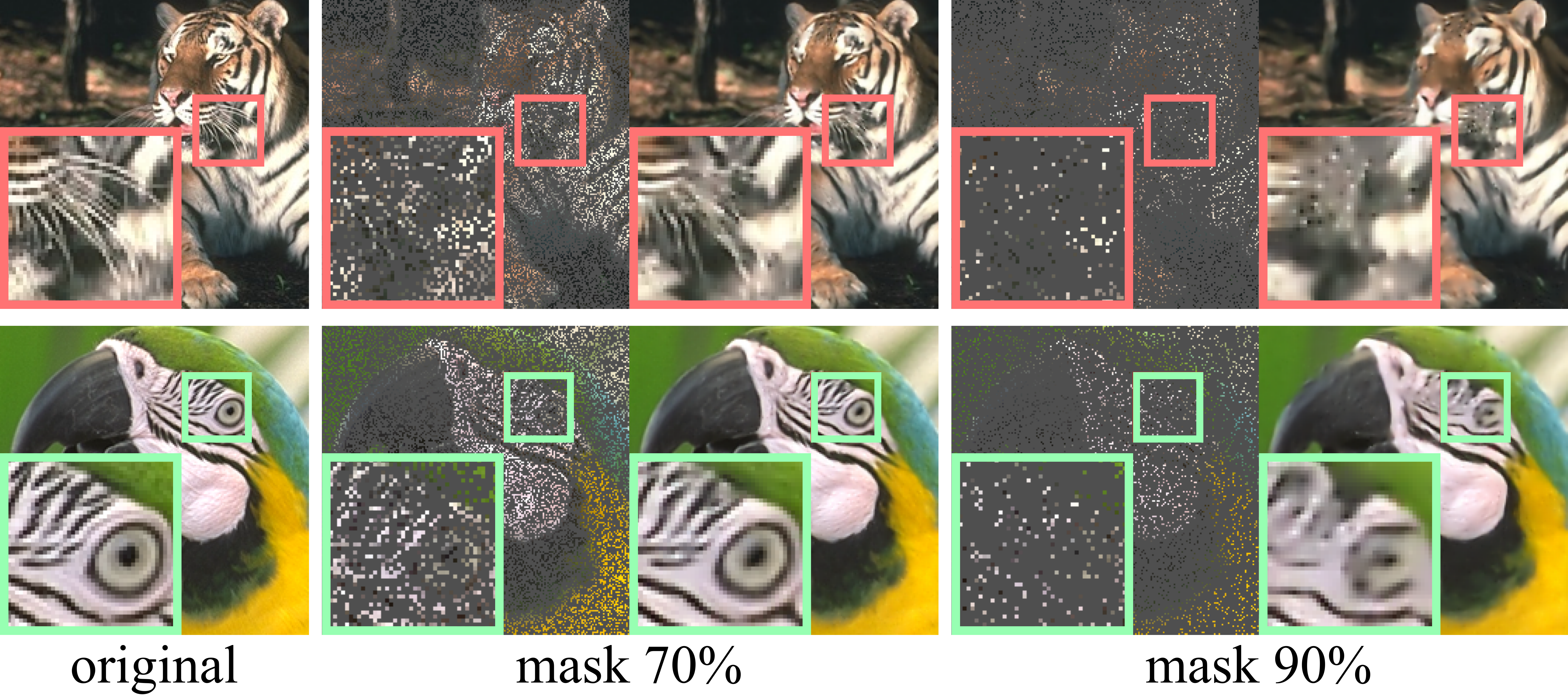}
   \vspace{-5mm}
   \caption{The illustration of the proposed mask-and-complete training strategy. Even if a large number of pixels are masked, the model can still reconstruct the input to some extent.}
   \label{fig:inpainting}
   \vspace{-5mm}
\end{figure}

\begin{figure*}[t]
    \includegraphics[width=\textwidth]{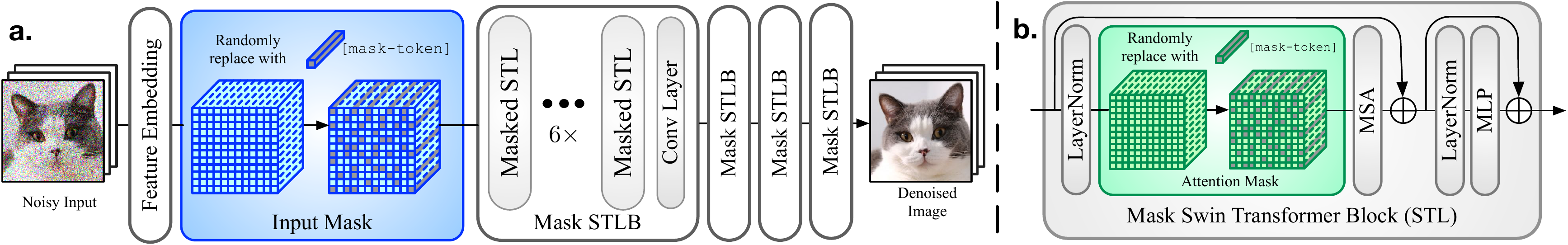}
    \vspace{-4mm}
    \caption{The transformer architecture of our proposed masked image training. We make a minimal change to the original SwinIR architecture -- the \textcolor{blue}{input mask} operation and the \textcolor{green}{attention masks}. Other micro-designs are not essentially different from other Transformers.}
    \label{fig:method}
    \vspace{-3mm}
\end{figure*}

\vspace{-4mm}
\paragraph{Motivation.}
When training a deep network on a large number of images, the expectation is for the network to learn to discern the rich semantics of natural images from noise-contaminated test cases.
However, several studies have noted that the semantics and knowledge acquired by low-level vision networks differ significantly from our expectations \cite{liu2021discovering,gu2021interpreting,liu2022evaluating,magid2022texture}.
We argue that the poor generalization ability of denoising models results from our training method, which leads the model to \emph{focus on overfitting the training noise rather than learning image reconstruction}.
We conduct a simple experiment for verification.
We trained a SwinIR denoising network \cite{liang2021swinir} using images that greatly differ from natural images (immunohistochemistry images \cite{uhlen2010towards}).
We synthesized training data pairs using Gaussian noise, and then assessed the model's performance on \emph{natural images} with Gaussian noise.
According to our hypothesis, if the model learns the content and reconstruction of image semantics from the training set, it should not perform well on natural images, as it has not been exposed to any.
If the model is simply overfitting the noise, the model can remove the noise even if the images are different, as the model mainly relies on detecting the noise for denoising.

The results are presented in Figure~\ref{fig:motivation}.
As observed, the SwinIR trained on immunohistochemistry images can still denoise and reproduce the natural image.
This supports our conjecture regarding generalization ability, indicating that most existing methods perform denoising by overfitting the training noise.
Consequently, when the noise deviates from the training conditions, the denoising performance of these models declines significantly.

This observation also inspires our approach to developing deep denoising models with improved generalization ability.
We aim for the model to learn the reconstruction of image textures and structures, rather than focusing only on noise.
In this paper, we propose a new masked training strategy for denoising networks.
During training, we mask out a portion of the input pixels and then train the deep network to complete them, as shown in \figurename~\ref{fig:inpainting}.
Our approach emphasizes reconstructing natural image textures and edges observed in the image, rather than overfitting noise.
In \figurename~\ref{fig:motivation} we also show the results of our method.
It is evident that our approach seeks to reconstruct the immunohistochemistry image texture from the training set on the testing natural image, instead of relying on noise overfitting for denoising.
This demonstrates the potential of this idea in improving generalization performance. By training our method on natural images, it will concentrate on reconstructing the content of natural images, aligning with our core concept of employing deep learning for low-level vision tasks.

\begin{figure}[t]
  \centering
   \includegraphics[width=\linewidth]{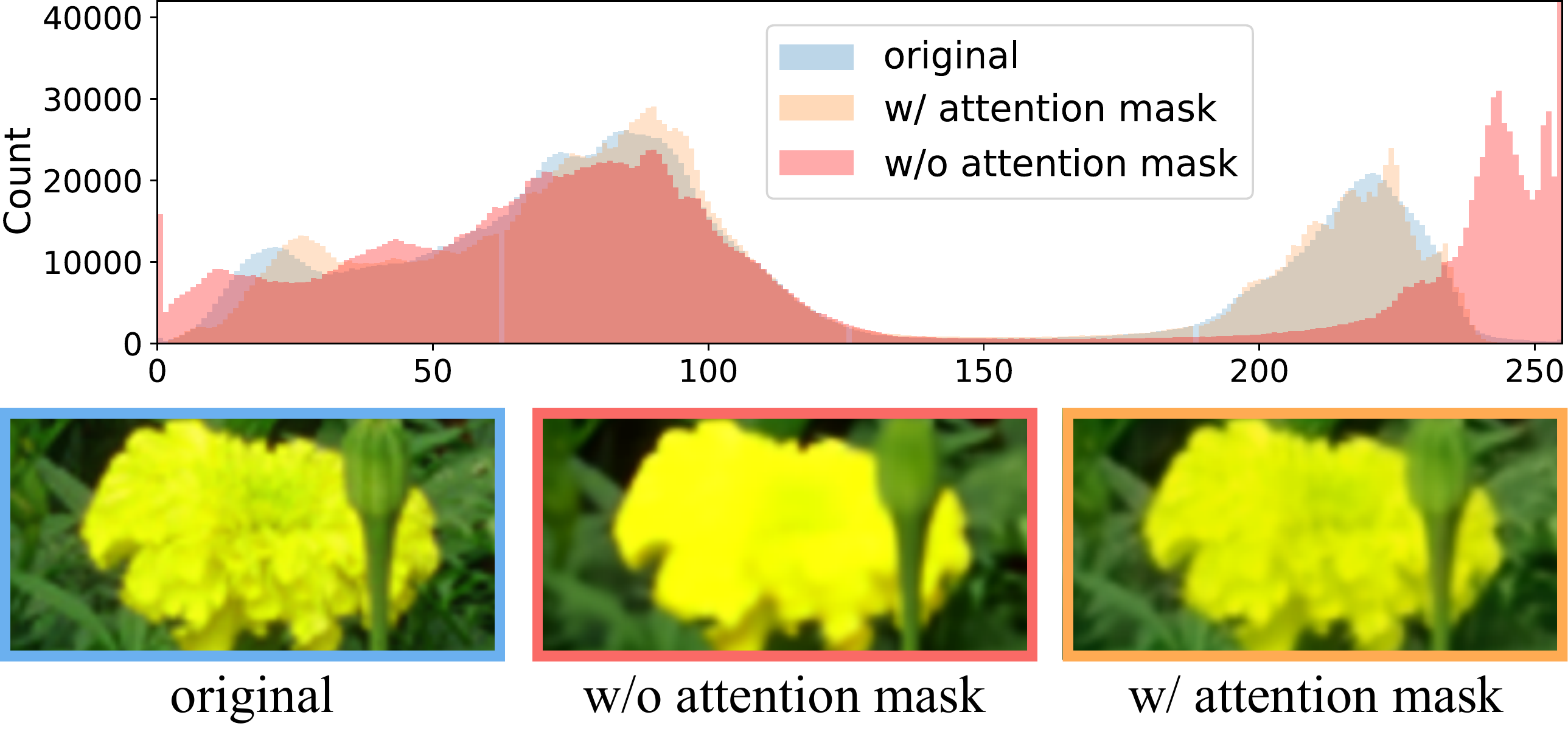}
   \vspace{-7mm}
   \caption{Quantitative effect of the attention mask. The histogram differences are also shown above.}
   \label{fig:attnmask}
   \vspace{-5mm}
\end{figure}

\vspace{-4mm}
\paragraph{The Transformer Architecture.}
Our approach exploits the excellent properties of visual Transformers, so we first describe the basic Transformer backbone used in this study.
The shifted window mechanism is proven to be flexible and effective for image processing tasks \cite{liang2021swinir,chen2022cross,zhang2022accurate}.
We only make minimal changes when applying it to the proposed masked training method without the loss of generality.
This model is illustrated in \figurename~\ref{fig:method}.
Transformers divide the input signal into tokens and process spatial information using self-attention layers.
In our method, a convolution layer with kernel size $1$ is used as the feature embedding module to project the 3-channel pixel values into $C$-dimensional feature tokens.
The $1\times 1$ convolution layer ensures that pixels do not affect each other during feature embedding, which facilitates subsequent masking operations.
These feature tokens are gathered with shape $H\times W\times C$, where $H$, $W$ and $C$ are the height, width and feature dimension.
The shifted window mechanism first reshapes the feature maps of each frame to $\frac{HW}{M^2}\times M^2\times C$ features by partitioning the input into non-overlapping $M\times M$ local windows, where $\frac{HW}{M^2}$ is the total number of windows.
We calculate self-attention on the feature tokens within the same window.
Therefore, $M^2$ tokens are involved in each standard self-attention operation, and we produce the local window feature $X\in\mathbb{R}^{M^2\times C}$.
In each self-attention layer, the query $Q$, key $K$ and value $V$ are calculated as 
$
    Q= XW^{Q},\quad K=XW^{K},\quad V=XW^{V},
$
where $W^{Q},W^{K},W^{V}\in\mathbb{R}^{C\times D}$ are weight matrices, and $D$ is the dimension of projected vectors.
Then, we use $Q$ to query $K$ to generate the attention map
$
    A=\mathtt{softmax}( \nicefrac{QK^T}{\sqrt{D}}+ B)\in\mathbb{R}^{M^2\times M^2},
$
where $B$ is the learnable relative positional encoding.
This attention map $A$ is then used for the weighted sum of $M^2$ vectors in $V$.
The multi-head settings are aligned with SwinIR \cite{liang2021swinir} and ViT \cite{dosovitskiy2020image}.

\begin{figure}[t]
  \centering
   \includegraphics[width=\linewidth]{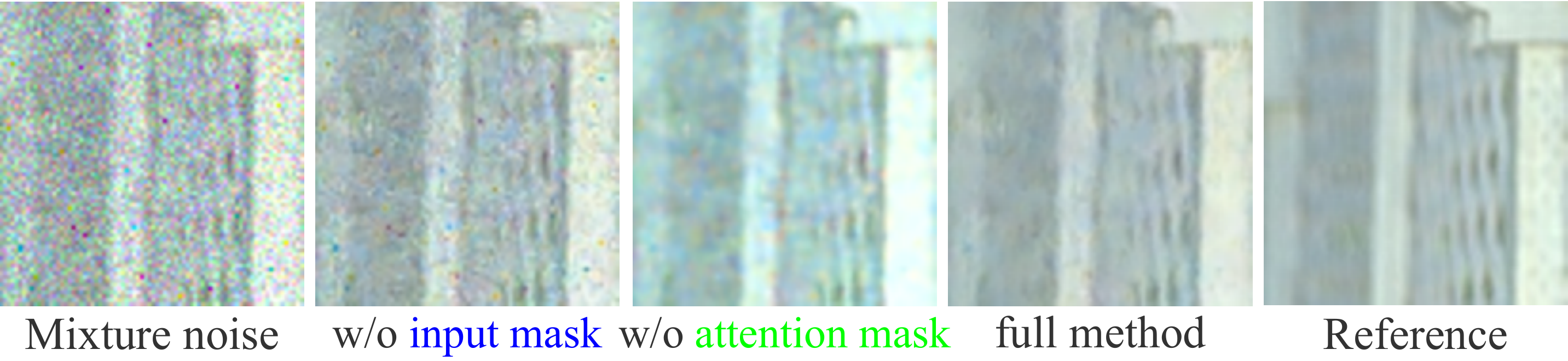}
   \vspace{-7mm}
   \caption{The effectiveness of the input mask and attention mask. Note that the brightness of the image is wrong \textit{w/o} attention mask.}
   \label{fig:ablation}
   \vspace{-2mm}
\end{figure}

\begin{figure}
\begin{minipage}{.535\linewidth}
\scalebox{0.98}{

\centering
\vspace{-1mm}
\resizebox{\textwidth}{!}{
\begin{tabular}{cccc}
\toprule
\rowcolor{color3}  \begin{tabular}[c]{@{}c@{}} \rowcolor{color3}  Input\\ Mask\end{tabular}   & \begin{tabular}[c]{@{}c@{}} \rowcolor{color3}  Attention\\ Mask\end{tabular} & PSNR & SSIM \\ \midrule
             & $\checkmark$ &  29.17 & 0.8227 \\
$\checkmark$ &              &  26.96 & 0.8202  \\
$\checkmark$ & $\checkmark$ &  \textbf{29.74} & \textbf{0.8672}  \\ 
\bottomrule
\end{tabular}
}
}
\vspace{-3mm}
\captionof{table}{The importance of using different mask operations.}\label{tab:differentmask}
\end{minipage}
\begin{minipage}{.45\linewidth}
\scalebox{0.76}{

\centering
\begin{tabular}{ccc}
        \toprule
        \rowcolor{color3}\multicolumn{3}{c}{\textbf{Mix. noise on CBSD68~\cite{martin2001database}}} \\
           \rowcolor{color3} Ratio (\%) & PSNR & SSIM \\
        \midrule
            65       & 29.57 & 0.8657         \\
            75       & \textbf{29.76} & \textbf{0.8678}         \\
            85       & 28.84 & 0.8548         \\
        \bottomrule
        \end{tabular} 
}
\vspace{-3mm}
\captionof{table}{Ablation on the attention mask ratio.}\label{tab:attentionmask}
\end{minipage}
\vspace{-5mm}
\end{figure}

\vspace{-3mm}
\paragraph{Masked Training.}
Our masked training mainly consists of two aspects, the input mask and the attention mask. Although both are mask operations, the purpose of these two masks is different. We describe them separately.

\textcolor{blue}{\emph{The Input Mask}} randomly masks out the feature tokens embedded by the first convolution layer, and encourages the network to complete the masked information during training.
The input mask explicitly constructs a very challenging inpainting problem, as shown in \figurename~\ref{fig:inpainting}.
It can be seen that even if up to 90\% of the pixel information is destroyed, the network can still reconstruct the target image to a certain extent.
The method is very simple.
Given the feature token tensor $\mathbf{f}\in\mathbb{R}^{H\times W\times C}$, we randomly replace the token with a \texttt{[mask token]}$\in\mathbb{R}^C$ with a probability $p_{\mathrm{IM}}$, where $p_{\mathrm{IM}}$ is called the input mask ratio.
The network is trained under the supervision of the $l_1$-norm of the reconstructed image and the ground truth.
The \texttt{[mask token]} can be learnable and initialized with a $\mathbf{0}$ vector. But we actually found that the $\mathbf{0}$ vector itself is already a suitable choice.
The existence of the input mask forces the network to learn to recognize and reconstruct the content of the image from very limited information.

\textcolor{green}{\emph{The Attention Mask.}}
We cannot build usable image processing networks relying solely on the input mask operation.
Because during testing, we will input uncorrupted images to retain enough information.
At this time, due to the inconsistency between training and testing, the network will tend to increase the brightness of the output image.
Such as the example in \figurename~\ref{fig:attnmask}.
Since Transformer uses the self-attention operation to process spatial information, we can narrow the gap between training and testing by performing the same mask operation during the self-attention process.
The specific mask operation is similar to the input mask, but a different attention mask ratio $p_{\mathrm{AM}}$ and \texttt{[mask token]} are used.
When some tokens in the self-attention are masked, the attention operation will adjust to the fact that the information of these tokens is no longer reliable.
Self-attention will focus on unmasked tokens in each layer and complete the masked information.
This operation is difficult to implement on convolutional networks.
\figurename~\ref{fig:attnmask} shows the effect of the attention mask.
As can be seen, the attention mask successfully makes the masked trained network work on the unmasked input image.

\begin{figure}[t]
  \centering
   \includegraphics[width=1\linewidth]{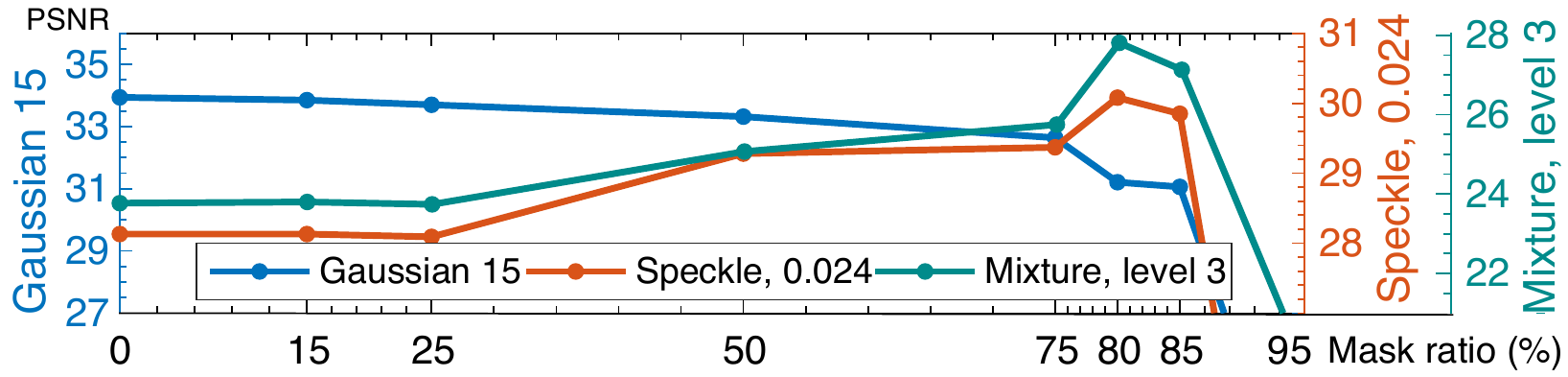}
   \vspace{-6mm}
    \caption{
The trade-off of choosing different mask ratios.
The performance drop on training noise is not significant until 75\%  masking ratio. 
Our performance gain on the noise outside the training set is greater than the performance loss on the training set.}
\label{fig:rebuttal_ratio}
\vspace{-4mm}
\end{figure}

\begin{figure*}[t]
\scriptsize
\centering
\resizebox{\textwidth}{!}{
\begin{tabular}{ccc}
\hspace{-0.45cm}
\begin{adjustbox}{valign=t}
\begin{tabular}{c}
\includegraphics[width=0.253\textwidth]{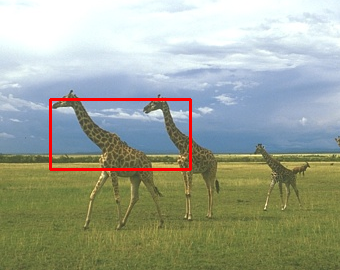}
\\
CBSD68: img\_053
\end{tabular}
\end{adjustbox}
\hspace{-0.46cm}
\begin{adjustbox}{valign=t}
\begin{tabular}{cccccc}
\includegraphics[width=0.18\textwidth]{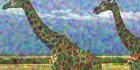} \hspace{-4mm} &
\includegraphics[width=0.18\textwidth]{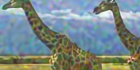} \hspace{-4mm} &
\includegraphics[width=0.18\textwidth]{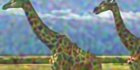} \hspace{-4mm} &
\includegraphics[width=0.18\textwidth]{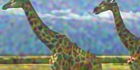} \hspace{-4mm} 
\\
{\fontsize{5pt}{0}\selectfont {Spatially correlated Gaussian, $\sigma=50$}}  \hspace{-4mm} &
DnCNN~\cite{zhang2017beyond} \hspace{-4mm} &
RIDNet~\cite{anwar2019real} \hspace{-4mm} &
RNAN~\cite{zhang2019residual} \hspace{-4mm}
\\
\includegraphics[width=0.18\textwidth]{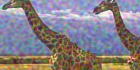} \hspace{-4mm} &
\includegraphics[width=0.18\textwidth]{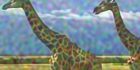} \hspace{-4mm} &
\includegraphics[width=0.18\textwidth]{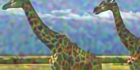} \hspace{-4mm} &
\includegraphics[width=0.18\textwidth]{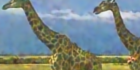} \hspace{-4mm}  
\\ 
Restormer~\cite{zamir2022restormer} \hspace{-4mm} &
SwinIR~\cite{liang2021swinir} \hspace{-4mm} &
baseline  \hspace{-4mm} &
Masked Training \hspace{-4mm}
\\
\end{tabular}
\end{adjustbox}
\\


\hspace{-0.45cm}
\begin{adjustbox}{valign=t}
\begin{tabular}{c}
\includegraphics[width=0.253\textwidth]{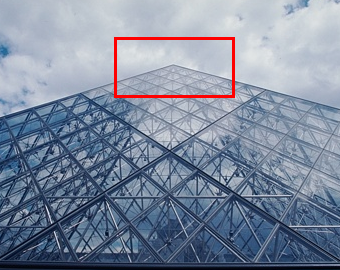}
\\
CBSD68: img\_0046
\end{tabular}
\end{adjustbox}
\hspace{-0.46cm}
\begin{adjustbox}{valign=t}
\begin{tabular}{cccccc}
\includegraphics[width=0.18\textwidth]{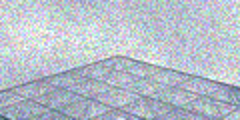} \hspace{-4mm} &
\includegraphics[width=0.18\textwidth]{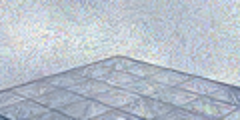} \hspace{-4mm} &
\includegraphics[width=0.18\textwidth]{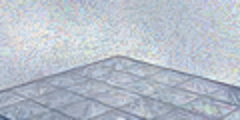} \hspace{-4mm} &
\includegraphics[width=0.18\textwidth]{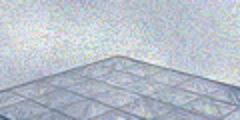} \hspace{-4mm} 
\\
Speckle noise, $\sigma^2=0.013$ \hspace{-4mm} &
DnCNN~\cite{zhang2017beyond} \hspace{-4mm} &
RIDNet~\cite{anwar2019real} \hspace{-4mm} &
RNAN~\cite{zhang2019residual} \hspace{-4mm}
\\
\includegraphics[width=0.18\textwidth]{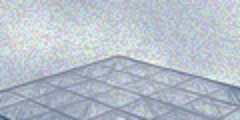} \hspace{-4mm} &
\includegraphics[width=0.18\textwidth]{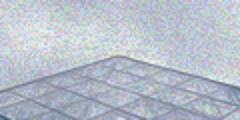} \hspace{-4mm} &
\includegraphics[width=0.18\textwidth]{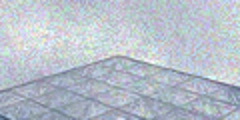} \hspace{-4mm} &
\includegraphics[width=0.18\textwidth]{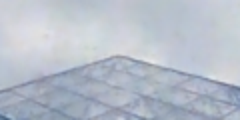} \hspace{-4mm}  
\\ 
Restormer~\cite{zamir2022restormer} \hspace{-4mm} &
SwinIR~\cite{liang2021swinir} \hspace{-4mm} &
baseline  \hspace{-4mm} &
Masked Training \hspace{-4mm}
\\
\end{tabular}
\end{adjustbox}
\\

\hspace{-0.45cm}
\begin{adjustbox}{valign=t}
\begin{tabular}{c}
\includegraphics[width=0.253\textwidth]{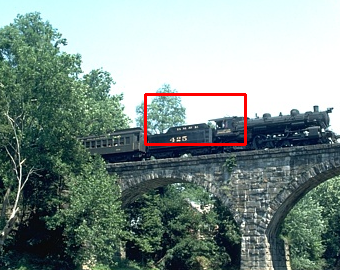}
\\
CBSD68: img\_0067
\end{tabular}
\end{adjustbox}
\hspace{-0.46cm}
\begin{adjustbox}{valign=t}
\begin{tabular}{cccccc}
\includegraphics[width=0.18\textwidth]{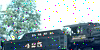} \hspace{-4mm} &
\includegraphics[width=0.18\textwidth]{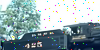} \hspace{-4mm} &
\includegraphics[width=0.18\textwidth]{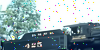} \hspace{-4mm} &
\includegraphics[width=0.18\textwidth]{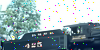} \hspace{-4mm} 
\\
{\fontsize{6pt}{0}\selectfont {Salt-and-pepper noise, $d = 0.02$}}  \hspace{-4mm} &
DnCNN~\cite{zhang2017beyond} \hspace{-4mm} &
RIDNet~\cite{anwar2019real} \hspace{-4mm} &
RNAN~\cite{zhang2019residual} \hspace{-4mm}
\\
\includegraphics[width=0.18\textwidth]{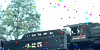} \hspace{-4mm} &
\includegraphics[width=0.18\textwidth]{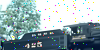} \hspace{-4mm} &
\includegraphics[width=0.18\textwidth]{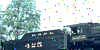} \hspace{-4mm} &
\includegraphics[width=0.18\textwidth]{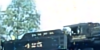} \hspace{-4mm}  
\\ 
Restormer~\cite{zamir2022restormer} \hspace{-4mm} &
SwinIR~\cite{liang2021swinir} \hspace{-4mm} &
baseline  \hspace{-4mm} &
Masked Training \hspace{-4mm}
\\
\end{tabular}
\end{adjustbox}

\end{tabular}}
\vspace{-3mm}
\caption{Visual comparison on out-of-distribution noise. When all other methods fail completely, our method is still able to denoise effectively.
Please refer to the supplementary material to see more visual results.}
\label{fig:main_compare}
\vspace{-1mm}
\end{figure*}

\begin{figure*}
\begin{minipage}{.63\textwidth}
\centering
\includegraphics[width=\linewidth]{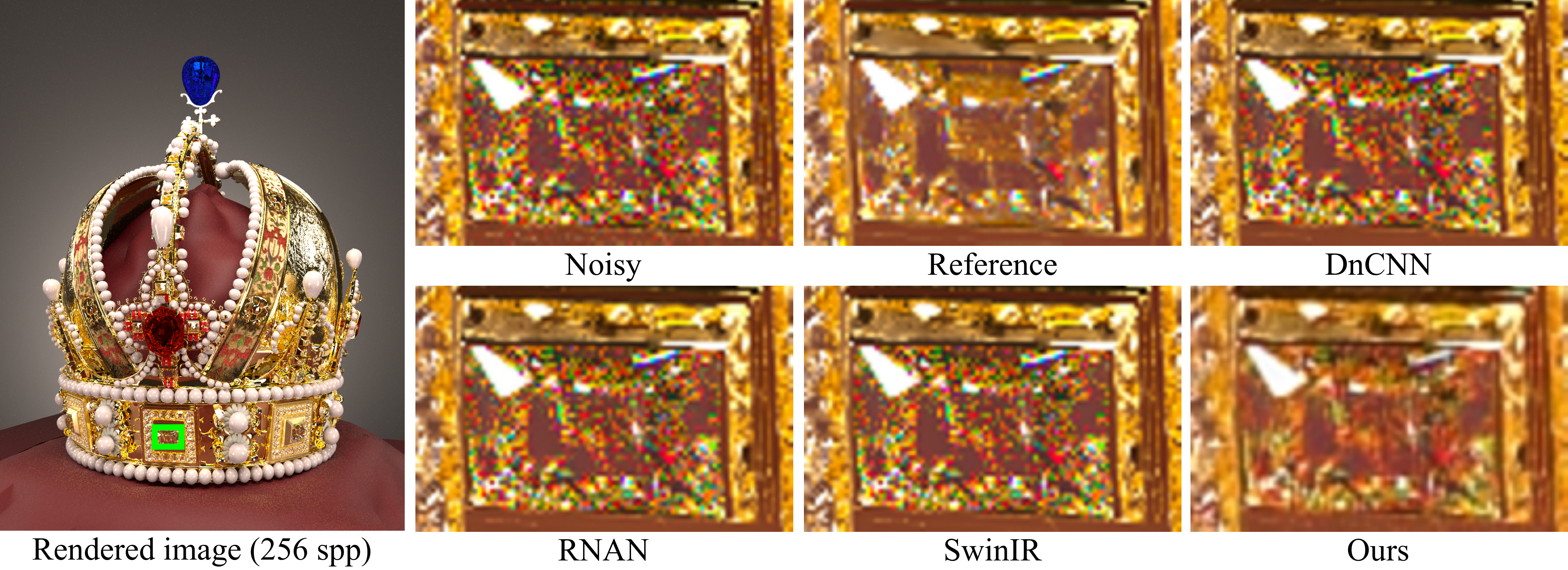}
\vspace{-7mm}
\caption{Visual results of denoising a Monte Carlo rendered image.}\label{fig:mcr}
\end{minipage}%
\hfill
\begin{minipage}{0.34\textwidth}
\centering
\includegraphics[width=\linewidth]{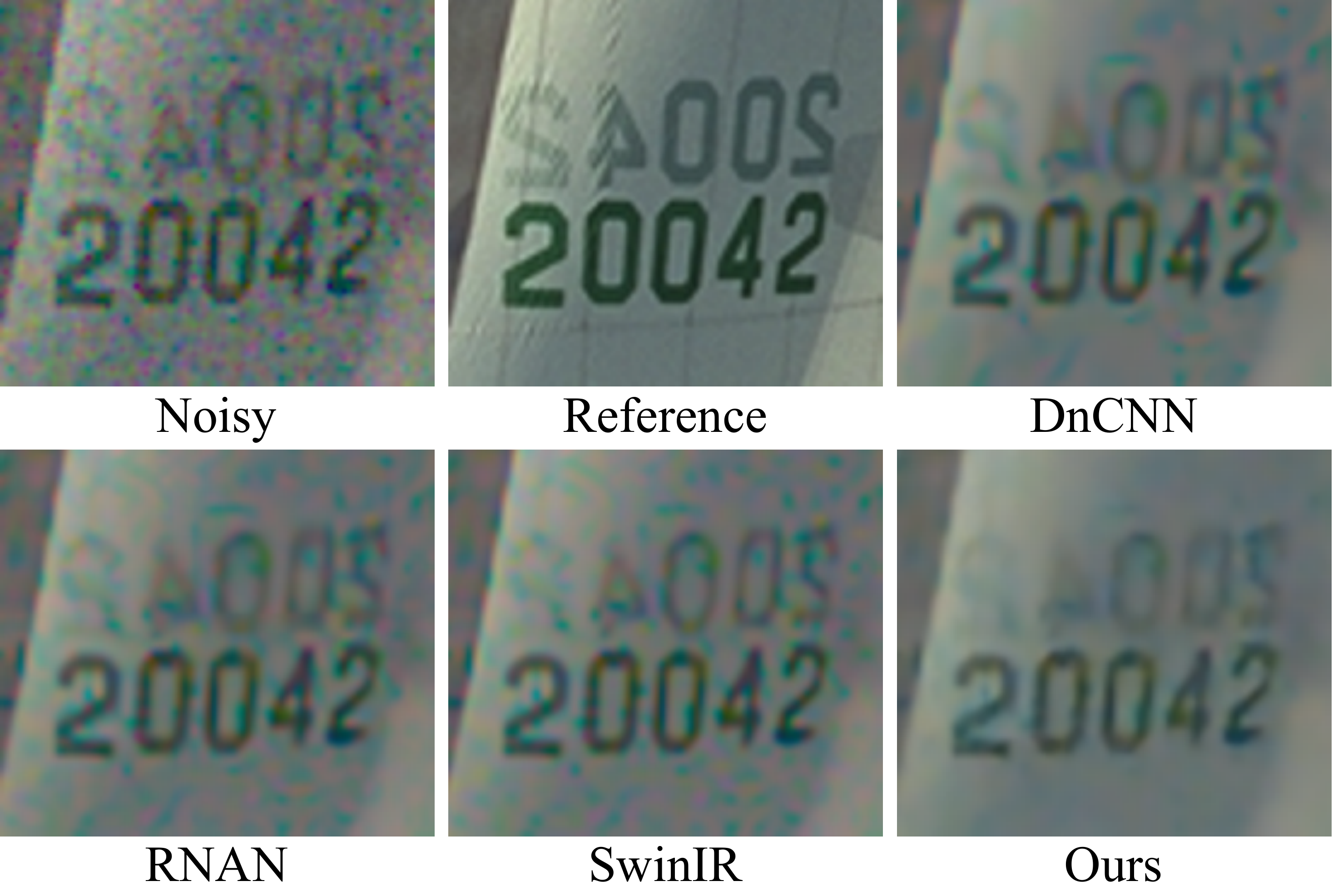}
\vspace{-7mm}
\caption{Results of ISP noise removal.}\label{fig:isp}
\end{minipage}
\vspace{-6mm}
\end{figure*}

\section{Experiments}
\label{Experiments}

\paragraph{Training Settings.}
For synthesizing training data, we sample the clean images from DIV2K~\cite{timofte2017ntire}, Flickr2K~\cite{lim2017enhanced}, BSD500~\cite{arbelaez2010contour}, and WED~\cite{ma2016waterloo} during training.
In our work, all the networks are trained using Gaussian noise with standard deviation $\sigma=15$.
Each input image is randomly cropped to a spatial resolution of 64$\times$64, and the number of the total training iteration is 200K.
We adopt the Adam optimizer~\cite{kingma2014adam} with $\beta_1$=$0.9$ and ${\beta}_2$=$0.99$ to minimize the $L_1$ pixel loss.
The initial learning rate is set as 1$\times$10$^{-4}$ and reduced by half at the milestone of 100K iterations and 150K iterations.
The batch size is set to 64.

\vspace{-4mm}
\paragraph{Testing Noise.}
Since the training process utilizes the Gaussian noise, we evaluate the generalization performance of the models on six other synthetic noise:
(1) Speckle noise, a type of noise that occurs during the acquisition of medical images or tomography images. 
(2) Poisson noise, a type of signal-dependent noise that occurs during the acquisition of digital images.
(3) Spatially-correlated noise. This is to synthesize the complex artifact after denoising using a flawed algorithm.
It is produced by filtering Gaussian noise with a $3\times3$ average kernel.
Different standard deviations of the Gaussian noise indicate different noise levels.
(4) Salt \& pepper noise.
(5) Image signal processing (ISP) noise. \cite{brooks2019unprocessing} proposes a method to synthesize realistic ISP noise during digital imaging.
(6) Mixture noise obtained by mixing the above different types of noise with different levels \cite{zhang2022practical}.
The clean images are sampled from the benchmark datasets, including CBSD68~\cite{martin2001database}, Kodak24~\cite{kodak24}, McMaster~\cite{zhang2011color}, and Urban100~\cite{huang2015single}.
We also include two real noise types in this work: the Smartphone Image Denoising Dataset (SIDD)~\cite{abdelhamed2018high} and Monte Carlo (MC) rendered image noise.
For evaluation, we follow \cite{gu2020pipal,gu2020image} and use the metrics PSNR, SSIM~\cite{ma2016waterloo}, and LPIPS~\cite{zhang2018perceptual} to evaluate the results.
Since PSNR and SSIM are questioned in assessing the perceptual quality of images \cite{gu2020pipal,gu2020image}, we also use the LPIPS as an additional metric.

\begin{table*}[t]
\centering
\subfloat[Abl. of input mask ratios.\label{tab:inputmaskratio}
]{
\scalebox{0.76}{
\begin{tabular}{ccc}
        \toprule
        \rowcolor{color3}\multicolumn{3}{c}{\textbf{Mix. noise on CBSD68~\cite{martin2001database}}} \\
           \rowcolor{color3} Ratio (\%) & PSNR & SSIM \\
        \midrule
            75       & 29.17 & 0.8132         \\
            85       & 29.44 & 0.8545         \\
            95       & 19.60 & 0.7273         \\ 
        \midrule
            70--80     &29.86    & 0.8593           \\
            75--85     & \textbf{30.04}   & \textbf{0.8756}      \\ 
            75--90     & 29.87   & 0.8728      \\ 
            75--95     & 29.26   & 0.8607      \\         
            80--90     & 29.74   & 0.8672      \\
        \bottomrule
        \end{tabular} 
}}
\subfloat[Quantitative comparison on Monte Carlo rendered image denoising.\label{tab:mcr}]{
\scalebox{0.76}{
\begin{tabular}{lccc|ccc}
\toprule 
         \rowcolor{color3} & \multicolumn{3}{c}{\textbf{128 samples per pixel}} & \multicolumn{3}{c}{\textbf{64 samples per pixel}}                  \\ 
    \rowcolor{color3} \textbf{Method} & PSNR    & SSIM     & LPIPS   & PSNR                      & SSIM   & LPIPS  \\ 
\midrule
DnCNN~\cite{zhang2017beyond}         & 29.94   & 0.7883   & 0.2671  & 26.28                     & 0.6779 & 0.4216 \\
RIDNet~\cite{anwar2019real}          & 29.96   & 0.7921   & 0.2548  & 26.27                     & 0.6788 & 0.4122 \\
RNAN~\cite{zhang2019residual}        & 29.86   & 0.7825   & 0.2702  & 26.26                     & 0.6743 & 0.4290 \\
SwinIR~\cite{liang2021swinir}        & 29.32   & 0.7627   & 0.2943  & 26.14                     & 0.6651 & 0.4485 \\
Restormer~\cite{zamir2022restormer}  & 24.98   & 0.6598   & 0.4575  & 24.59                     & 0.5880 & 0.5375 \\
Dropout~\cite{kong2022reflash}       & 28.85   & 0.7753   & 0.2941  & 26.10                     & 0.6696 & 0.4443 \\
baseline    & 29.68   & 0.7738   & 0.2851  & 25.91                     & 0.6535 & 0.4564 \\ 
\midrule
\textbf{Ours} & \cellcolor{green!15}\textbf{30.62}   & \cellcolor{green!15}\textbf{0.8500}   & \cellcolor{green!15}\textbf{0.2254}  & \cellcolor{green!15}\textbf{28.25}                     & \cellcolor{green!15}\textbf{0.7694} & \cellcolor{green!15}\textbf{0.3348} \\ 
\bottomrule
\end{tabular}
}}
\subfloat[Comparison on synthetic ISP noise.\label{tab:isp}]{
\scalebox{0.76}{
\begin{tabular}{lccc}
\toprule
\rowcolor{color3}   \multicolumn{4}{c}{\textbf{Synthetic ISP noise \cite{brooks2019unprocessing}}}  \\
    \rowcolor{color3} \textbf{Method} & PSNR  & SSIM   & LPIPS \\
\midrule
DnCNN~\cite{zhang2017beyond}       & \textbf{29.44} & 0.7857  & 0.3083 \\
RIDNet~\cite{anwar2019real}        & 28.75 & 0.7446  & 0.3696 \\
RNAN~\cite{zhang2019residual}      & 28.47 & 0.7243  & 0.3601 \\
SwinIR~\cite{liang2021swinir}      & 28.39 & 0.7079  & 0.3346 \\
Restormer~\cite{zamir2022restormer}& 19.31 & 0.4982  & 0.6556 \\
Dropout~\cite{kong2022reflash}     & 28.39 & 0.7816  & 0.2621 \\
baseline & 28.89 & 0.7595  & 0.2917 \\ 
\midrule
\textbf{Ours}     & \cellcolor{green!15}\textbf{29.44} & \cellcolor{green!15}\textbf{0.7920}  & \cellcolor{green!15}\textbf{0.2368} \\ %
\bottomrule
\end{tabular} 
}}
\vspace{-2mm}
\caption{We train all the models on Gaussian noise, $\sigma=15$. All the testing noise is out of the training set, therefore the results can show the models' generalization performance on different unseen noise.}
\vspace{-2mm}
\end{table*}

\begin{figure*}[t]
  \centering
  \includegraphics[width=\linewidth]{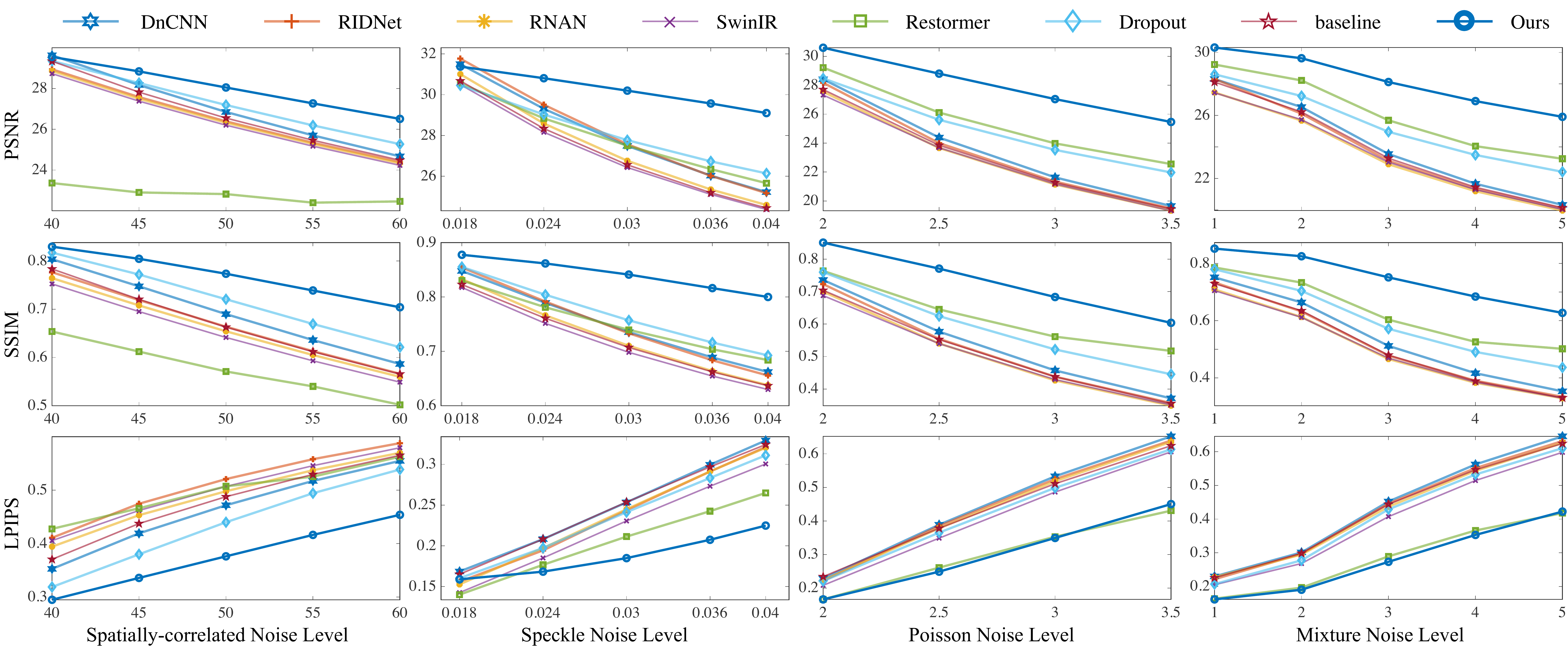}
  \vspace{-7mm}
  \caption{Performance comparisons on four noise types with different levels on the Kodak24 dataset \cite{kodak24}. All models are trained only on Gaussian noise. Our masked training approach demonstrates good generalization performance across different noise types. We involve multiple types and levels of noise in testing, the results cannot be shown here. More results are shown in the supplementary material.}
  \label{fig:performance}
  \vspace{-5mm}
\end{figure*}

\subsection{Resutls}
\paragraph{Ablation Study.}
\tablename~\ref{tab:differentmask} and \figurename~\ref{fig:ablation} show the effectiveness of using different mask operations.
As we can see, without the input mask, the model will lose its generalization ability, and cannot effectively remove the noise outside the training set.
Without the attention mask, due to the training-testing inconsistency, the quantitative performance degrades significantly, and the output image will have the wrong brightness. 
In addition, even without the attention mask, the generalization ability of the model is not significantly affected, and most of the noise is still effectively removed.
The input mask is the crucial factor in improving the model the generalization ability.

\tablename~\ref{tab:inputmaskratio} shows the impact of the different input mask ratios.
We test fixed ratios and random ratios from a uniform distribution.
From our experiments, fixed ratios are less stable for training than randomly chosen from a range, and the performance is also worse.
The best quantitative performance is achieved with random sampling ratios between 75\% $\sim$ 85\%.
This is a trade-off between denoising generalization ability and the preservation of image details.
As shown in \figurename~\ref{fig:rebuttal_ratio}, smaller ratios are not enough for the network to learn the distribution of images because more noise patterns are preserved. 
The larger ratio improves the model generalization, as the model focuses more on reconstruction. But at the same time, some image details may be lost.
For attention mask ratio, we show the effects in \tablename~\ref{tab:attentionmask}.
The optimal ratios are around 75\%.

\vspace{-4mm}
\paragraph{The Generalization Performance.}
We evaluate our deep denoising method on synthetic noise, where our training noise follows a Gaussian distribution with a single noise level, but we test on multiple types of non-Gaussian noise to assess the model's generalization performance.
In \figurename~\ref{fig:performance}, we compare our method with other state-of-the-art models based on their PSNR and SSIM scores.
The results show that our model outperforms all the other models in terms of generalization performance.
Particularly, as the noise level increases, our model exhibits a slower performance degradation and thus demonstrates better generalization.
In contrast, other models suffer from significant performance drops when dealing with more severe noise.
We also provide visual comparisons in \figurename~\ref{fig:main_compare}, where our model achieves remarkable denoising results even though it is trained only on Gaussian noise with a fixed standard deviation.
In contrast, existing models tend to overfit the training noise and fail when facing unseen noise.
More quantitative and qualitative results can be found in the supplementary material.

\vspace{-4mm}
\paragraph{Evaluation on ISP noise.}
The removal of the ISP noise is of great application value.
Brooks \etal \cite{brooks2019unprocessing} present a systematic approach for generating realistic raw data with ISP noise that can
 facilitate our research.
We use the default parameter settings of the method proposed in \cite{brooks2019unprocessing} to synthesize ISP noise on the Kodak24~\cite{kodak24} dataset for testing.
The results are shown in \figurename~\ref{fig:isp} and \tablename~\ref{tab:isp}.
Our method achieves superior results compared to all other methods. Notably, our method achieves a significant lead in LPIPS, indicating that our results exhibit better perceptual quality. 
Although DnCNN and our method obtain the same PSNR, our method still outperforms DnCNN in terms of SSIM and LPIPS.
Furthermore, as evident from \figurename~\ref{fig:isp}, DnCNN's results still contain visible noise, while our method effectively removes the noise.

\vspace{-4mm}
\paragraph{Evaluation on Monte Carlo rendering noise.}
Monte Carlo denoising is a vital component of the rendering process since the widespread use in the industry of Monte Carlo rendering algorithms \cite{christensen2018renderman,burley2018design,kulla2018sony}.
We use the test dataset proposed by \cite{firmino2022progressive} for Monte Carlo rendered image denoising. 
The test images were rendered in 128 samples-per-pixel (spp) and 64 spp. The lower the spp, the more severe the noise of the image.
In order to adapt the test set to our model, we first convert the data set to sRGB color space by tone mapping.
\figurename~\ref{fig:mcr} and \tablename~\ref{tab:mcr} show the denoising results. 
Our method outperforms all methods on both 128spp and 64spp settings.
In \figurename~\ref{fig:mcr}, the existing methods fail completely because of poor generalization.
Our model is still able to remove this noise, demonstrating the wide applicability of our method.


\begin{figure*}[!t]
  \centering
   \includegraphics[width=\linewidth]{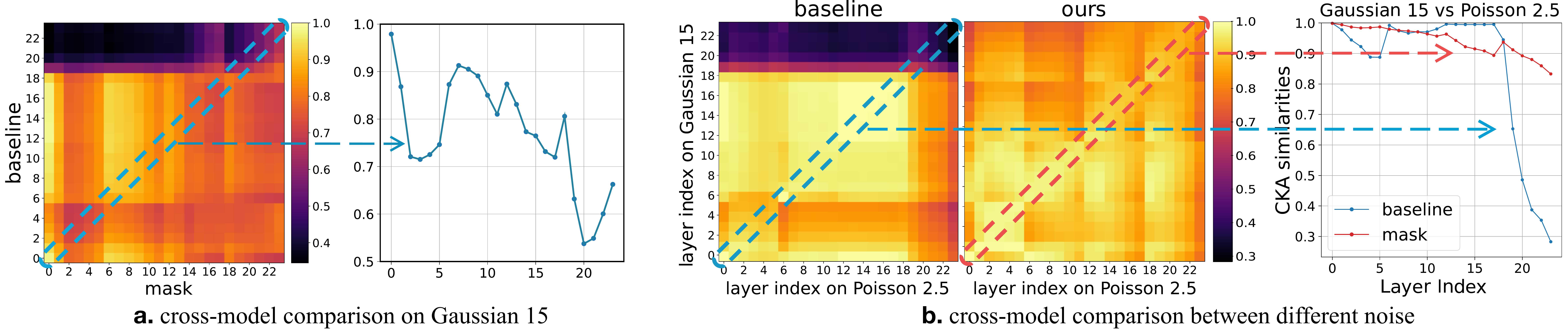}
   \vspace{-6mm}
   \caption{CKA similarity to analyze the representation similarity of network layers.}
   \label{fig:cka}
   \vspace{-4mm}
\end{figure*}

\begin{figure}[t]
  \centering
   \includegraphics[width=\linewidth]{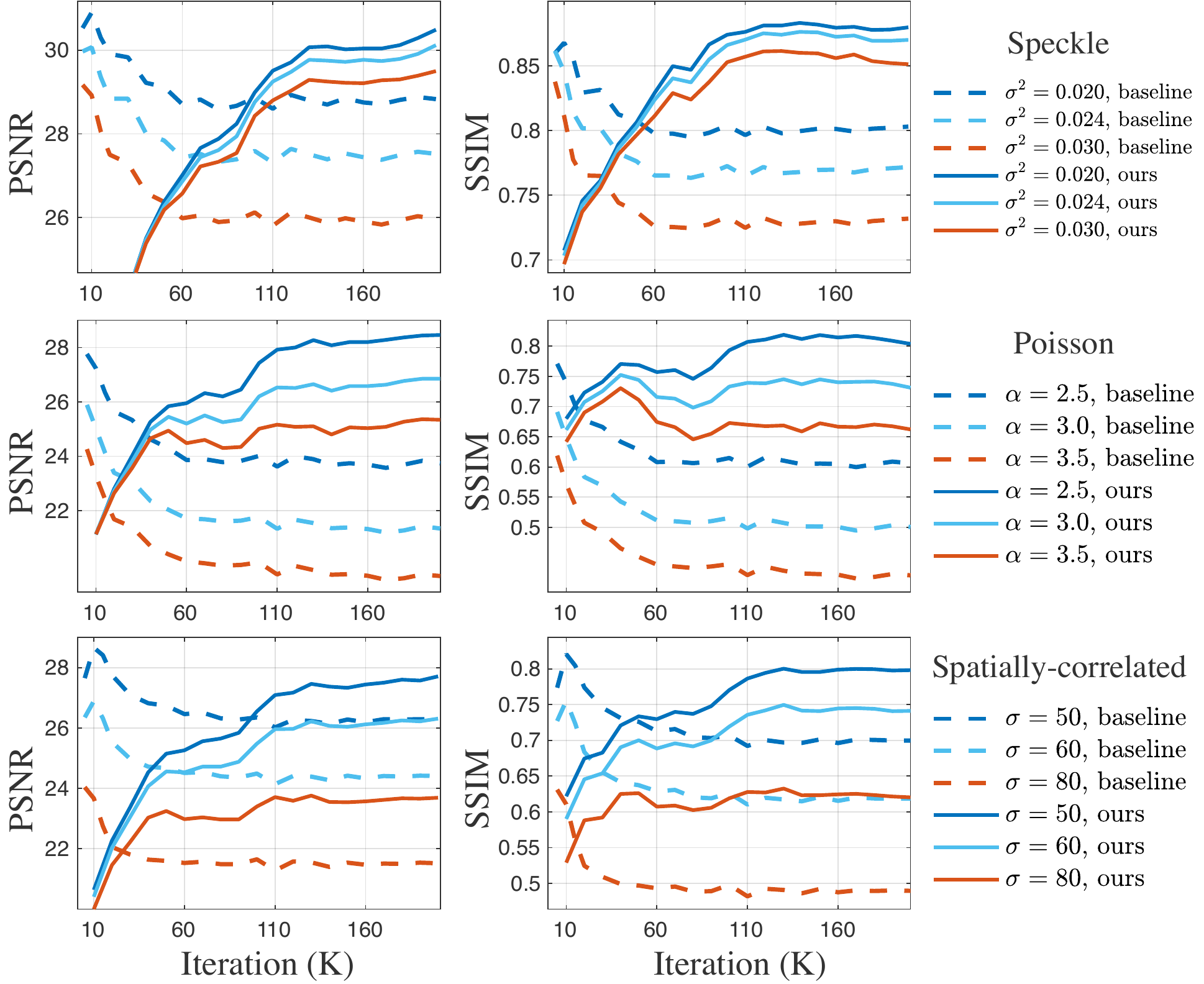}
   \vspace{-6mm}
   \caption{The testing curves on different noise types and levels.}
   \label{fig:train}
   \vspace{-1mm}
\end{figure}

\begin{figure}[t]
  \centering
   \includegraphics[width=\linewidth]{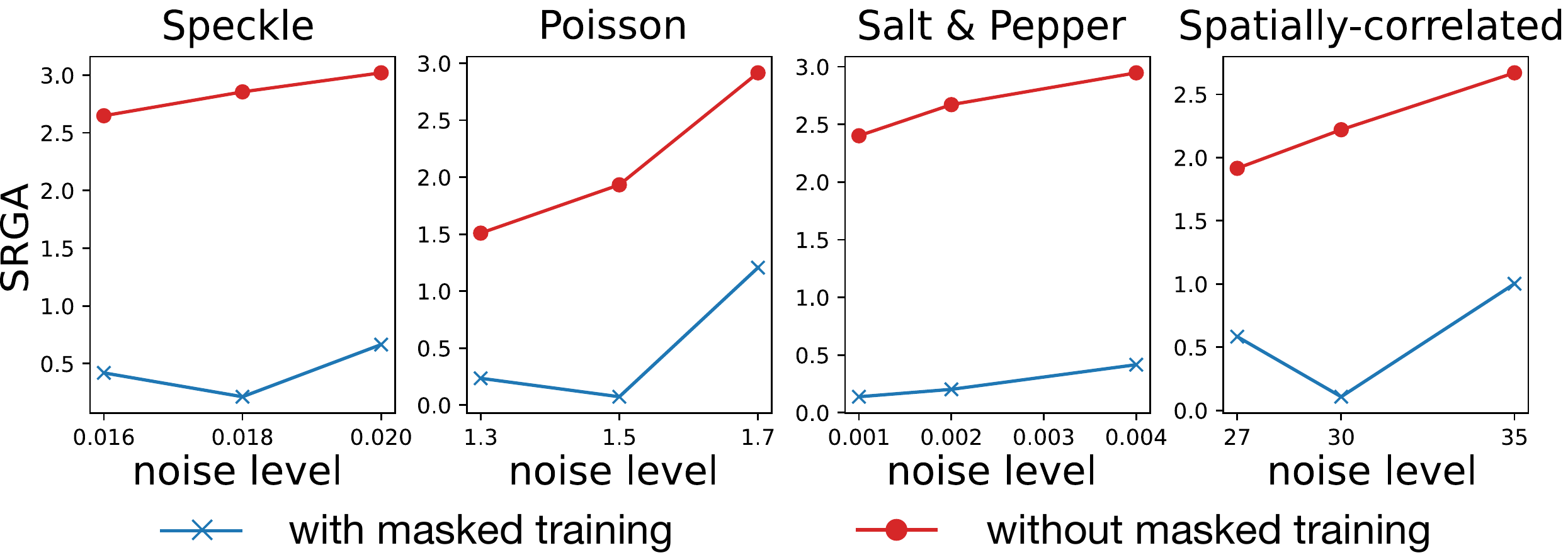}
   \vspace{-6mm}
   \caption{Comparing generalization ability with the SRGA metric. A lower SRGA value indicates better generalization ability.}
   \label{fig:srga}
   \vspace{-1mm}
\end{figure}

\begin{figure}[t]
  \centering
   \includegraphics[width=0.95\linewidth]{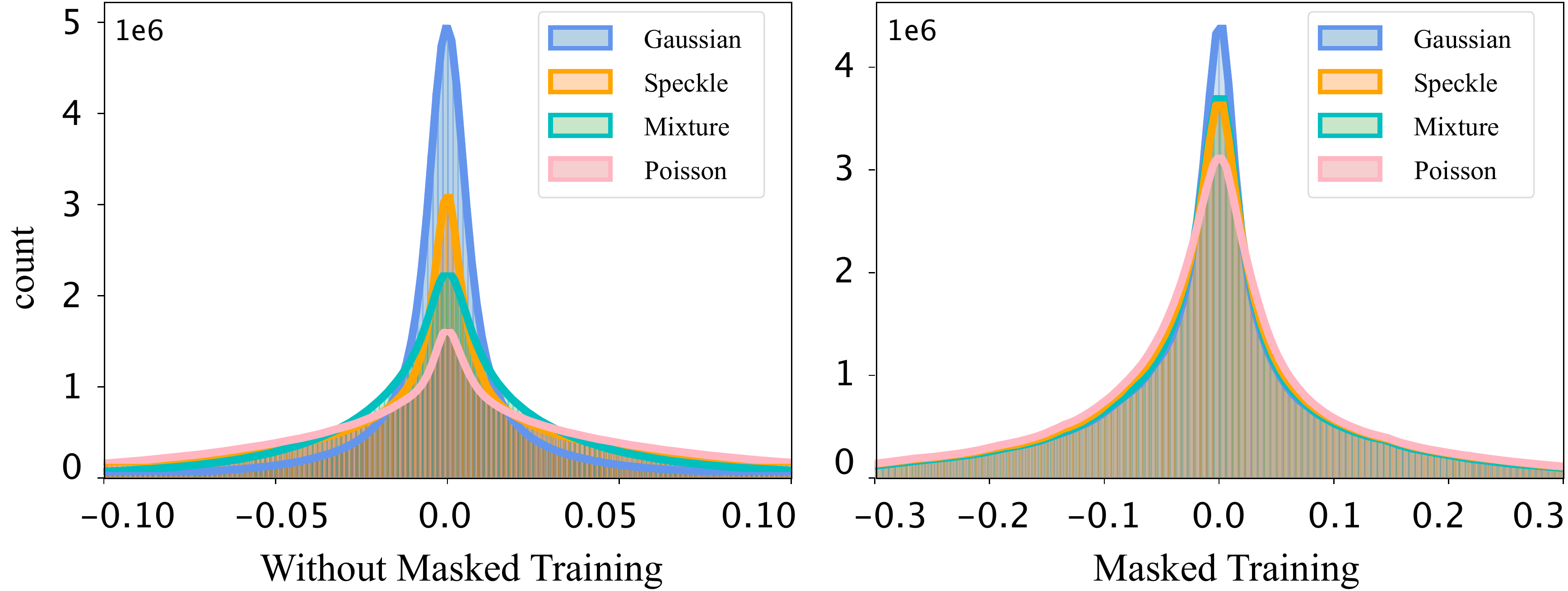}
   \vspace{-2mm}
   \caption{
   The distribution of baseline model features is biased across different noise types. Our method produces similar feature distributions across different noise.}
   \label{fig:distribution}
   \vspace{-3mm}
\end{figure}


\subsection{Generalization Analysis}

\paragraph{Training curve.}
\figurename~\ref{fig:train} shows the training curves of the model with and without the proposed masked training.
The models are trained using only Gaussian noise. The baseline method has a significant overfitting problem. The performance of our method gradually improves with training without overfitting.

\vspace{-4mm}
\paragraph{CKA analysis.}
To investigate how masked training differs from normal training strategy, we utilize the centered kernel alignment (CKA) \cite{cortes2012algorithms,raghu2021vision} to analyze the differences between network representations obtained from those two training methods.
Due to the limited space, we describe the detail of CKA in supplementary.
In \figurename~\ref{fig:cka}, we present our key findings. Specifically, \figurename~\ref{fig:cka} (a) shows the cross-model comparison between the baseline model and our masked training model.
We observe a significant difference between the two models in terms of their feature correlations in the deeper layers.
Specifically, the features of the deeper layers of the baseline model exhibit low correlations with all layers of our model. This finding suggests that these two training methods exhibit inconsistent learning patterns for features, especially for the deeper layers.

To explore how the models perform on different noise types, \figurename~\ref{fig:cka} (b) shows the cross-noise comparison between in-distribution noise and out-of-distribution noise, such as Gaussian and Poisson noise.
For the baseline model, we observe a low correlation between different noise types in the deep layers, indicating that the network processes these two types of noise in different ways for the deep layers.
This trend holds for other types of noise as well.
This phenomenon may be due to the baseline approach causing the deep layers of the model to overfit to the patterns of the training set, thereby limiting their generalization capabilities to handle different noise types.
In contrast, the high correlation between adjacent layers in our masked training model suggests that the model's representation of the two different noise types is similar. The proposed masked training forces the network to learn the underlying distribution of the images themselves, which makes the model more robust to different types of noise and enhances its generalization capability.

\vspace{-4mm}
\paragraph{Quantification of generalization performance.}
Liu \etal \cite{liu2022evaluating,liu2021discovering} suggest that model generalization ability can be measured by measuring the consistency of the model’s representations across different types of noise.
They also propose a generalization assessment index for low-level vision networks called SRGA \cite{liu2022evaluating}.
It is a non-parametric and non-learning metric which exploits the statistical characteristics of internal features of deep networks.
The lower the value of SRGA, the better the generalization ability.
In our case, we use Gaussian noise as the reference and other types of noise for testing.
\figurename~\ref{fig:srga} shows the SRGA results.
Inspired by \cite{liu2022evaluating}, we visualize the distributions of deep features on different noise types, shown in \figurename~\ref{fig:distribution}.
We can see that for the baseline model, the feature distributions under different noise types deviate from each other significantly.
For the model \textit{w/} masked training, the deep feature distributions of different noise types are close to each other.
This confirms the effectiveness of our method.

\section{Conclusion and Limitations}
In summary, our masked training method provides a promising approach to improving the generalization performance of deep learning-based image denoising models.
The limitation of our method is that the mask operation inevitably loses information. How to preserve more details needs to be explored in future work.
Our approach is a step towards developing more robust models for real-world applications.


\paragraph{Acknowledgment.}
This work is supported in part by Guangzhou Municipal Science and Technology Project (Grant No. 2023A03J0671), the National Natural Science Foundation of China under Grant (62276251), the Joint Lab of CAS-HK, and the Youth Innovation Promotion Association of Chinese Academy of Sciences (No. 2020356).

{\small
\bibliographystyle{ieee_fullname}
\bibliography{egbib}

\begin{thebibliography}{10}\itemsep=-1pt

\bibitem{abdelhamed2018high}
Abdelrahman Abdelhamed, Stephen Lin, and Michael~S Brown.
\newblock A high-quality denoising dataset for smartphone cameras.
\newblock In {\em Proceedings of the IEEE Conference on Computer Vision and
  Pattern Recognition}, pages 1692--1700, 2018.

\bibitem{anwar2019real}
Saeed Anwar and Nick Barnes.
\newblock Real image denoising with feature attention.
\newblock In {\em Proceedings of the IEEE/CVF international conference on
  computer vision}, pages 3155--3164, 2019.

\bibitem{arbelaez2010contour}
Pablo Arbelaez, Michael Maire, Charless Fowlkes, and Jitendra Malik.
\newblock Contour detection and hierarchical image segmentation.
\newblock {\em TPAMI}, 2010.

\bibitem{bao2021beit}
Hangbo Bao, Li Dong, and Furu Wei.
\newblock Beit: Bert pre-training of image transformers.
\newblock {\em arXiv preprint arXiv:2106.08254}, 2021.

\bibitem{brooks2019unprocessing}
Tim Brooks, Ben Mildenhall, Tianfan Xue, Jiawen Chen, Dillon Sharlet, and
  Jonathan~T Barron.
\newblock Unprocessing images for learned raw denoising.
\newblock In {\em Proceedings of the IEEE/CVF Conference on Computer Vision and
  Pattern Recognition}, pages 11036--11045, 2019.

\bibitem{brown2020language}
Tom Brown, Benjamin Mann, Nick Ryder, Melanie Subbiah, Jared~D Kaplan, Prafulla
  Dhariwal, Arvind Neelakantan, Pranav Shyam, Girish Sastry, Amanda Askell,
  et~al.
\newblock Language models are few-shot learners.
\newblock {\em Advances in neural information processing systems},
  33:1877--1901, 2020.

\bibitem{buades2005non}
Antoni Buades, Bartomeu Coll, and J-M Morel.
\newblock A non-local algorithm for image denoising.
\newblock In {\em 2005 IEEE computer society conference on computer vision and
  pattern recognition (CVPR'05)}, volume~2, pages 60--65. Ieee, 2005.

\bibitem{burley2018design}
Brent Burley, David Adler, Matt Jen-Yuan Chiang, Hank Driskill, Ralf Habel,
  Patrick Kelly, Peter Kutz, Yining~Karl Li, and Daniel Teece.
\newblock The design and evolution of disney’s hyperion renderer.
\newblock {\em ACM Transactions on Graphics (TOG)}, 37(3):1--22, 2018.

\bibitem{chen2018deep}
Chang Chen, Zhiwei Xiong, Xinmei Tian, and Feng Wu.
\newblock Deep boosting for image denoising.
\newblock In {\em Proceedings of the European Conference on Computer Vision
  (ECCV)}, pages 3--18, 2018.

\bibitem{chen2021pre}
Hanting Chen, Yunhe Wang, Tianyu Guo, Chang Xu, Yiping Deng, Zhenhua Liu, Siwei
  Ma, Chunjing Xu, Chao Xu, and Wen Gao.
\newblock Pre-trained image processing transformer.
\newblock In {\em CVPR}, 2021.

\bibitem{chen2018image}
Jingwen Chen, Jiawei Chen, Hongyang Chao, and Ming Yang.
\newblock Image blind denoising with generative adversarial network based noise
  modeling.
\newblock In {\em Proceedings of the IEEE conference on computer vision and
  pattern recognition}, pages 3155--3164, 2018.

\bibitem{chen2020generative}
Mark Chen, Alec Radford, Rewon Child, Jeffrey Wu, Heewoo Jun, David Luan, and
  Ilya Sutskever.
\newblock Generative pretraining from pixels.
\newblock In {\em International conference on machine learning}, pages
  1691--1703. PMLR, 2020.

\bibitem{chen2016trainable}
Yunjin Chen and Thomas Pock.
\newblock Trainable nonlinear reaction diffusion: A flexible framework for fast
  and effective image restoration.
\newblock {\em IEEE transactions on pattern analysis and machine intelligence},
  39(6):1256--1272, 2016.

\bibitem{chen2023recursive}
Zheng Chen, Yulun Zhang, Jinjin Gu, Linghe Kong, and Xiaokang Yang.
\newblock Recursive generalization transformer for image super-resolution.
\newblock {\em arXiv preprint arXiv:2303.06373}, 2023.

\bibitem{chen2022cross}
Zheng Chen, Yulun Zhang, Jinjin Gu, Yongbing Zhang, Linghe Kong, and Xin Yuan.
\newblock Cross aggregation transformer for image restoration.
\newblock In {\em NIPS}, 2022.

\bibitem{cheng2021nbnet}
Shen Cheng, Yuzhi Wang, Haibin Huang, Donghao Liu, Haoqiang Fan, and Shuaicheng
  Liu.
\newblock Nbnet: Noise basis learning for image denoising with subspace
  projection.
\newblock In {\em Proceedings of the IEEE/CVF Conference on Computer Vision and
  Pattern Recognition}, pages 4896--4906, 2021.

\bibitem{christensen2018renderman}
Per Christensen, Julian Fong, Jonathan Shade, Wayne Wooten, Brenden Schubert,
  Andrew Kensler, Stephen Friedman, Charlie Kilpatrick, Cliff Ramshaw, Marc
  Bannister, et~al.
\newblock Renderman: An advanced path-tracing architecture for movie rendering.
\newblock {\em ACM Transactions on Graphics (TOG)}, 37(3):1--21, 2018.

\bibitem{cortes2012algorithms}
Corinna Cortes, Mehryar Mohri, and Afshin Rostamizadeh.
\newblock Algorithms for learning kernels based on centered alignment.
\newblock {\em The Journal of Machine Learning Research}, 13:795--828, 2012.

\bibitem{dabov2007image}
Kostadin Dabov, Alessandro Foi, Vladimir Katkovnik, and Karen Egiazarian.
\newblock Image denoising by sparse 3-d transform-domain collaborative
  filtering.
\newblock {\em IEEE Transactions on image processing}, 16(8):2080--2095, 2007.

\bibitem{devlin2018bert}
Jacob Devlin, Ming-Wei Chang, Kenton Lee, and Kristina Toutanova.
\newblock Bert: Pre-training of deep bidirectional transformers for language
  understanding.
\newblock {\em arXiv preprint arXiv:1810.04805}, 2018.

\bibitem{divakar2017image}
Nithish Divakar and R Venkatesh~Babu.
\newblock Image denoising via cnns: An adversarial approach.
\newblock In {\em Proceedings of the IEEE Conference on Computer Vision and
  Pattern Recognition Workshops}, pages 80--87, 2017.

\bibitem{dosovitskiy2020image}
Alexey Dosovitskiy, Lucas Beyer, Alexander Kolesnikov, Dirk Weissenborn,
  Xiaohua Zhai, Thomas Unterthiner, Mostafa Dehghani, Matthias Minderer, Georg
  Heigold, Sylvain Gelly, et~al.
\newblock An image is worth 16x16 words: Transformers for image recognition at
  scale.
\newblock {\em arXiv preprint arXiv:2010.11929}, 2020.

\bibitem{elad2006image}
Michael Elad and Michal Aharon.
\newblock Image denoising via sparse and redundant representations over learned
  dictionaries.
\newblock {\em IEEE Transactions on Image processing}, 15(12):3736--3745, 2006.

\bibitem{feng2019suppressing}
Ruicheng Feng, Jinjin Gu, Yu Qiao, and Chao Dong.
\newblock Suppressing model overfitting for image super-resolution networks.
\newblock In {\em Proceedings of the IEEE/CVF Conference on Computer Vision and
  Pattern Recognition Workshops}, pages 0--0, 2019.

\bibitem{firmino2022progressive}
Arthur Firmino, Jeppe~Revall Frisvad, and Henrik~Wann Jensen.
\newblock Progressive denoising of monte carlo rendered images.
\newblock In {\em Computer Graphics Forum}, volume~41, pages 1--11. Wiley
  Online Library, 2022.

\bibitem{kodak24}
Rich Franzen.
\newblock Kodak lossless true color image suite.
\newblock source: http://r0k.us/graphics/kodak/, 1999.

\bibitem{gu2020image}
Jinjin Gu, Haoming Cai, Haoyu Chen, Xiaoxing Ye, Jimmy Ren, and Chao Dong.
\newblock Image quality assessment for perceptual image restoration: A new
  dataset, benchmark and metric.
\newblock {\em arXiv preprint arXiv:2011.15002}, 2020.

\bibitem{gu2020pipal}
Jinjin Gu, Haoming Cai, Haoyu Chen, Xiaoxing Ye, Jimmy Ren, and Chao Dong.
\newblock Pipal: a large-scale image quality assessment dataset for perceptual
  image restoration.
\newblock In {\em European Conference on Computer Vision}, pages 633--651.
  Springer, 2020.

\bibitem{gu2021interpreting}
Jinjin Gu and Chao Dong.
\newblock Interpreting super-resolution networks with local attribution maps.
\newblock In {\em Proceedings of the IEEE/CVF Conference on Computer Vision and
  Pattern Recognition}, pages 9199--9208, 2021.

\bibitem{gu2019blind}
Jinjin Gu, Hannan Lu, Wangmeng Zuo, and Chao Dong.
\newblock Blind super-resolution with iterative kernel correction.
\newblock In {\em Proceedings of the IEEE/CVF Conference on Computer Vision and
  Pattern Recognition}, pages 1604--1613, 2019.

\bibitem{gu2019self}
Shuhang Gu, Yawei Li, Luc~Van Gool, and Radu Timofte.
\newblock Self-guided network for fast image denoising.
\newblock In {\em Proceedings of the IEEE/CVF International Conference on
  Computer Vision}, pages 2511--2520, 2019.

\bibitem{gu2014weighted}
Shuhang Gu, Lei Zhang, Wangmeng Zuo, and Xiangchu Feng.
\newblock Weighted nuclear norm minimization with application to image
  denoising.
\newblock In {\em Proceedings of the IEEE conference on computer vision and
  pattern recognition}, pages 2862--2869, 2014.

\bibitem{guo2019toward}
Shi Guo, Zifei Yan, Kai Zhang, Wangmeng Zuo, and Lei Zhang.
\newblock Toward convolutional blind denoising of real photographs.
\newblock In {\em Proceedings of the IEEE/CVF conference on computer vision and
  pattern recognition}, pages 1712--1722, 2019.

\bibitem{he2022masked}
Kaiming He, Xinlei Chen, Saining Xie, Yanghao Li, Piotr Doll{\'a}r, and Ross
  Girshick.
\newblock Masked autoencoders are scalable vision learners.
\newblock In {\em Proceedings of the IEEE/CVF Conference on Computer Vision and
  Pattern Recognition}, pages 16000--16009, 2022.

\bibitem{hu2021pseudo}
Xiaowan Hu, Ruijun Ma, Zhihong Liu, Yuanhao Cai, Xiaole Zhao, Yulun Zhang, and
  Haoqian Wang.
\newblock Pseudo 3d auto-correlation network for real image denoising.
\newblock In {\em Proceedings of the IEEE/CVF Conference on Computer Vision and
  Pattern Recognition}, pages 16175--16184, 2021.

\bibitem{huang2015single}
Jia-Bin Huang, Abhishek Singh, and Narendra Ahuja.
\newblock Single image super-resolution from transformed self-exemplars.
\newblock In {\em CVPR}, 2015.

\bibitem{jia2019focnet}
Xixi Jia, Sanyang Liu, Xiangchu Feng, and Lei Zhang.
\newblock Focnet: A fractional optimal control network for image denoising.
\newblock In {\em Proceedings of the IEEE/CVF Conference on Computer Vision and
  Pattern Recognition}, pages 6054--6063, 2019.

\bibitem{kingma2014adam}
Diederik Kingma and Jimmy Ba.
\newblock Adam: A method for stochastic optimization.
\newblock In {\em ICLR}, 2015.

\bibitem{kokkinos2018deep}
Filippos Kokkinos and Stamatios Lefkimmiatis.
\newblock Deep image demosaicking using a cascade of convolutional residual
  denoising networks.
\newblock In {\em Proceedings of the European conference on computer vision
  (ECCV)}, pages 303--319, 2018.

\bibitem{kong2022reflash}
Xiangtao Kong, Xina Liu, Jinjin Gu, Yu Qiao, and Chao Dong.
\newblock Reflash dropout in image super-resolution.
\newblock In {\em Proceedings of the IEEE/CVF Conference on Computer Vision and
  Pattern Recognition}, pages 6002--6012, 2022.

\bibitem{kornblith2019similarity}
Simon Kornblith, Mohammad Norouzi, Honglak Lee, and Geoffrey Hinton.
\newblock Similarity of neural network representations revisited.
\newblock In {\em International Conference on Machine Learning}, pages
  3519--3529. PMLR, 2019.

\bibitem{krull2019noise2void}
Alexander Krull, Tim-Oliver Buchholz, and Florian Jug.
\newblock Noise2void-learning denoising from single noisy images.
\newblock In {\em Proceedings of the IEEE/CVF conference on computer vision and
  pattern recognition}, pages 2129--2137, 2019.

\bibitem{kulla2018sony}
Christopher Kulla, Alejandro Conty, Clifford Stein, and Larry Gritz.
\newblock Sony pictures imageworks arnold.
\newblock {\em ACM Transactions on Graphics (TOG)}, 37(3):1--18, 2018.

\bibitem{lefkimmiatis2017non}
Stamatios Lefkimmiatis.
\newblock Non-local color image denoising with convolutional neural networks.
\newblock In {\em Proceedings of the IEEE conference on computer vision and
  pattern recognition}, pages 3587--3596, 2017.

\bibitem{lefkimmiatis2018universal}
Stamatios Lefkimmiatis.
\newblock Universal denoising networks: a novel cnn architecture for image
  denoising.
\newblock In {\em Proceedings of the IEEE conference on computer vision and
  pattern recognition}, pages 3204--3213, 2018.

\bibitem{liang2021swinir}
Jingyun Liang, Jiezhang Cao, Guolei Sun, Kai Zhang, Luc Van~Gool, and Radu
  Timofte.
\newblock Swinir: Image restoration using swin transformer.
\newblock In {\em CVPR}, 2021.

\bibitem{lim2017enhanced}
Bee Lim, Sanghyun Son, Heewon Kim, Seungjun Nah, and Kyoung~Mu Lee.
\newblock Enhanced deep residual networks for single image super-resolution.
\newblock In {\em CVPRW}, 2017.

\bibitem{liu2022blind}
Anran Liu, Yihao Liu, Jinjin Gu, Yu Qiao, and Chao Dong.
\newblock Blind image super-resolution: A survey and beyond.
\newblock {\em IEEE Transactions on Pattern Analysis and Machine Intelligence},
  2022.

\bibitem{liu2018non}
Ding Liu, Bihan Wen, Yuchen Fan, Chen~Change Loy, and Thomas~S Huang.
\newblock Non-local recurrent network for image restoration.
\newblock {\em Advances in neural information processing systems}, 31, 2018.

\bibitem{liu2021discovering}
Yihao Liu, Anran Liu, Jinjin Gu, Zhipeng Zhang, Wenhao Wu, Yu Qiao, and Chao
  Dong.
\newblock Discovering" semantics" in super-resolution networks.
\newblock {\em arXiv preprint arXiv:2108.00406}, 2021.

\bibitem{liu2022evaluating}
Yihao Liu, Hengyuan Zhao, Jinjin Gu, Yu Qiao, and Chao Dong.
\newblock Evaluating the generalization ability of super-resolution networks.
\newblock {\em arXiv preprint arXiv:2205.07019}, 2022.

\bibitem{ma2016waterloo}
Kede Ma, Zhengfang Duanmu, Qingbo Wu, Zhou Wang, Hongwei Yong, Hongliang Li,
  and Lei Zhang.
\newblock Waterloo exploration database: New challenges for image quality
  assessment models.
\newblock {\em TIP}, 2016.

\bibitem{magid2022texture}
Salma~Abdel Magid, Zudi Lin, Donglai Wei, Yulun Zhang, Jinjin Gu, and Hanspeter
  Pfister.
\newblock Texture-based error analysis for image super-resolution.
\newblock In {\em Proceedings of the IEEE/CVF Conference on Computer Vision and
  Pattern Recognition}, pages 2118--2127, 2022.

\bibitem{mairal2009non}
Julien Mairal, Francis Bach, Jean Ponce, Guillermo Sapiro, and Andrew
  Zisserman.
\newblock Non-local sparse models for image restoration.
\newblock In {\em 2009 IEEE 12th international conference on computer vision},
  pages 2272--2279. IEEE, 2009.

\bibitem{mao2016image}
Xiaojiao Mao, Chunhua Shen, and Yu-Bin Yang.
\newblock Image restoration using very deep convolutional encoder-decoder
  networks with symmetric skip connections.
\newblock {\em Advances in neural information processing systems}, 29, 2016.

\bibitem{martin2001database}
David Martin, Charless Fowlkes, Doron Tal, and Jitendra Malik.
\newblock A database of human segmented natural images and its application to
  evaluating segmentation algorithms and measuring ecological statistics.
\newblock In {\em Proceedings Eighth IEEE International Conference on Computer
  Vision. ICCV 2001}, volume~2, pages 416--423. IEEE, 2001.

\bibitem{pathak2016context}
Deepak Pathak, Philipp Krahenbuhl, Jeff Donahue, Trevor Darrell, and Alexei~A
  Efros.
\newblock Context encoders: Feature learning by inpainting.
\newblock In {\em Proceedings of the IEEE conference on computer vision and
  pattern recognition}, pages 2536--2544, 2016.

\bibitem{plotz2017benchmarking}
Tobias Plotz and Stefan Roth.
\newblock Benchmarking denoising algorithms with real photographs.
\newblock In {\em Proceedings of the IEEE conference on computer vision and
  pattern recognition}, pages 1586--1595, 2017.

\bibitem{plotz2018neural}
Tobias Pl{\"o}tz and Stefan Roth.
\newblock Neural nearest neighbors networks.
\newblock {\em Advances in Neural information processing systems}, 31, 2018.

\bibitem{radford2018improving}
Alec Radford, Karthik Narasimhan, Tim Salimans, Ilya Sutskever, et~al.
\newblock Improving language understanding by generative pre-training.
\newblock 2018.

\bibitem{radford2019language}
Alec Radford, Jeffrey Wu, Rewon Child, David Luan, Dario Amodei, Ilya
  Sutskever, et~al.
\newblock Language models are unsupervised multitask learners.
\newblock {\em OpenAI blog}, 1(8):9, 2019.

\bibitem{raghu2021vision}
Maithra Raghu, Thomas Unterthiner, Simon Kornblith, Chiyuan Zhang, and Alexey
  Dosovitskiy.
\newblock Do vision transformers see like convolutional neural networks?
\newblock {\em Advances in Neural Information Processing Systems},
  34:12116--12128, 2021.

\bibitem{shi2022rethinking}
Shuwei Shi, Jinjin Gu, Liangbin Xie, Xintao Wang, Yujiu Yang, and Chao Dong.
\newblock Rethinking alignment in video super-resolution transformers.
\newblock {\em arXiv preprint arXiv:2207.08494}, 2022.

\bibitem{tai2017memnet}
Ying Tai, Jian Yang, Xiaoming Liu, and Chunyan Xu.
\newblock Memnet: A persistent memory network for image restoration.
\newblock In {\em Proceedings of the IEEE international conference on computer
  vision}, pages 4539--4547, 2017.

\bibitem{timofte2017ntire}
Radu Timofte, Eirikur Agustsson, Luc Van~Gool, Ming-Hsuan Yang, Lei Zhang, Bee
  Lim, Sanghyun Son, Heewon Kim, Seungjun Nah, Kyoung~Mu Lee, et~al.
\newblock Ntire 2017 challenge on single image super-resolution: Methods and
  results.
\newblock In {\em CVPRW}, 2017.

\bibitem{uhlen2010towards}
Mathias Uhlen, Per Oksvold, Linn Fagerberg, Emma Lundberg, Kalle Jonasson,
  Mattias Forsberg, Martin Zwahlen, Caroline Kampf, Kenneth Wester, Sophia
  Hober, et~al.
\newblock Towards a knowledge-based human protein atlas.
\newblock {\em Nature biotechnology}, 28(12):1248--1250, 2010.

\bibitem{vincent2010stacked}
Pascal Vincent, Hugo Larochelle, Isabelle Lajoie, Yoshua Bengio, Pierre-Antoine
  Manzagol, and L{\'e}on Bottou.
\newblock Stacked denoising autoencoders: Learning useful representations in a
  deep network with a local denoising criterion.
\newblock {\em Journal of machine learning research}, 11(12), 2010.

\bibitem{wang2021uformer}
Zhendong Wang, Xiaodong Cun, Jianmin Bao, and Jianzhuang Liu.
\newblock Uformer: A general u-shaped transformer for image restoration.
\newblock {\em arXiv preprint arXiv:2106.03106}, 2021.

\bibitem{wei2020physics}
Kaixuan Wei, Ying Fu, Jiaolong Yang, and Hua Huang.
\newblock A physics-based noise formation model for extreme low-light raw
  denoising.
\newblock In {\em Proceedings of the IEEE/CVF Conference on Computer Vision and
  Pattern Recognition}, pages 2758--2767, 2020.

\bibitem{xie2022simmim}
Zhenda Xie, Zheng Zhang, Yue Cao, Yutong Lin, Jianmin Bao, Zhuliang Yao, Qi
  Dai, and Han Hu.
\newblock Simmim: A simple framework for masked image modeling.
\newblock In {\em Proceedings of the IEEE/CVF Conference on Computer Vision and
  Pattern Recognition}, pages 9653--9663, 2022.

\bibitem{yang2020learning}
Fuzhi Yang, Huan Yang, Jianlong Fu, Hongtao Lu, and Baining Guo.
\newblock Learning texture transformer network for image super-resolution.
\newblock In {\em CVPR}, 2020.

\bibitem{yuan2018unsupervised}
Yuan Yuan, Siyuan Liu, Jiawei Zhang, Yongbing Zhang, Chao Dong, and Liang Lin.
\newblock Unsupervised image super-resolution using cycle-in-cycle generative
  adversarial networks.
\newblock In {\em Proceedings of the IEEE Conference on Computer Vision and
  Pattern Recognition Workshops}, pages 701--710, 2018.

\bibitem{yue2019variational}
Zongsheng Yue, Hongwei Yong, Qian Zhao, Deyu Meng, and Lei Zhang.
\newblock Variational denoising network: Toward blind noise modeling and
  removal.
\newblock {\em Advances in neural information processing systems}, 32, 2019.

\bibitem{yue2020dual}
Zongsheng Yue, Qian Zhao, Lei Zhang, and Deyu Meng.
\newblock Dual adversarial network: Toward real-world noise removal and noise
  generation.
\newblock In {\em European Conference on Computer Vision}, pages 41--58.
  Springer, 2020.

\bibitem{zamir2021restormer}
Syed~Waqas Zamir, Aditya Arora, Salman Khan, Munawar Hayat, Fahad~Shahbaz Khan,
  and Ming-Hsuan Yang.
\newblock Restormer: Efficient transformer for high-resolution image
  restoration.
\newblock In {\em CVPR}, 2022.

\bibitem{zamir2022restormer}
Syed~Waqas Zamir, Aditya Arora, Salman Khan, Munawar Hayat, Fahad~Shahbaz Khan,
  and Ming-Hsuan Yang.
\newblock Restormer: Efficient transformer for high-resolution image
  restoration.
\newblock In {\em Proceedings of the IEEE/CVF Conference on Computer Vision and
  Pattern Recognition}, pages 5728--5739, 2022.

\bibitem{zamir2021multi}
Syed~Waqas Zamir, Aditya Arora, Salman Khan, Munawar Hayat, Fahad~Shahbaz Khan,
  Ming-Hsuan Yang, and Ling Shao.
\newblock Multi-stage progressive image restoration.
\newblock In {\em Proceedings of the IEEE/CVF conference on computer vision and
  pattern recognition}, pages 14821--14831, 2021.

\bibitem{zhang2023xformer}
Jiale Zhang, Yulun Zhang, Jinjin Gu, Jiahua Dong, Linghe Kong, and Xiaokang
  Yang.
\newblock Xformer: Hybrid x-shaped transformer for image denoising.
\newblock {\em arXiv preprint arXiv:2303.06440}, 2023.

\bibitem{zhang2022accurate}
Jiale Zhang, Yulun Zhang, Jinjin Gu, Yongbing Zhang, Linghe Kong, and Xin Yuan.
\newblock Accurate image restoration with attention retractable transformer.
\newblock {\em arXiv preprint arXiv:2210.01427}, 2022.

\bibitem{zhang2022practical}
Kai Zhang, Yawei Li, Jingyun Liang, Jiezhang Cao, Yulun Zhang, Hao Tang, Radu
  Timofte, and Luc Van~Gool.
\newblock Practical blind denoising via swin-conv-unet and data synthesis.
\newblock {\em arXiv preprint arXiv:2203.13278}, 2022.

\bibitem{zhang2017beyond}
Kai Zhang, Wangmeng Zuo, Yunjin Chen, Deyu Meng, and Lei Zhang.
\newblock Beyond a gaussian denoiser: Residual learning of deep cnn for image
  denoising.
\newblock {\em IEEE transactions on image processing}, 26(7):3142--3155, 2017.

\bibitem{zhang2018ffdnet}
Kai Zhang, Wangmeng Zuo, and Lei Zhang.
\newblock Ffdnet: Toward a fast and flexible solution for cnn-based image
  denoising.
\newblock {\em IEEE Transactions on Image Processing}, 27(9):4608--4622, 2018.

\bibitem{zhang2011color}
Lei Zhang, Xiaolin Wu, Antoni Buades, and Xin Li.
\newblock Color demosaicking by local directional interpolation and nonlocal
  adaptive thresholding.
\newblock {\em Journal of Electronic imaging}, 20(2):023016, 2011.

\bibitem{zhang2018perceptual}
Richard Zhang, Phillip Isola, Alexei~A Efros, Eli Shechtman, and Oliver Wang.
\newblock The unreasonable effectiveness of deep features as a perceptual
  metric.
\newblock In {\em CVPR}, 2018.

\bibitem{zhang2021accurate}
Yulun Zhang, Kunpeng Li, Kai Li, Gan Sun, Yu Kong, and Yun Fu.
\newblock Accurate and fast image denoising via attention guided scaling.
\newblock {\em IEEE Transactions on Image Processing}, 30:6255--6265, 2021.

\bibitem{zhang2019residual}
Yulun Zhang, Kunpeng Li, Kai Li, Bineng Zhong, and Yun Fu.
\newblock Residual non-local attention networks for image restoration.
\newblock {\em arXiv preprint arXiv:1903.10082}, 2019.

\bibitem{zhang2020residual}
Yulun Zhang, Yapeng Tian, Yu Kong, Bineng Zhong, and Yun Fu.
\newblock Residual dense network for image restoration.
\newblock {\em IEEE Transactions on Pattern Analysis and Machine Intelligence},
  43(7):2480--2495, 2020.

\end{thebibliography}
}

\clearpage

\appendix

\section*{Appendix}

\section{Details of the Test Noise}
\label{sec:testnoise}
We evaluate the generalization performance of the models on six different synthetic noise types to evaluate the generalization performance on the noise out of the training set:

\noindent (1) \textbf{Speckle noise} is a kind of noise that can occur during the acquisition of medical images or tomography images. 
We use different variances $\sigma^2$ to obtain different levels of noise. The \textit{imnoise} function in MATLAB is used for generating Speckle noise. 
We add multiplicative noise according to the equation $J = I+n*I$, where $n$ is uniformly distributed random noise with mean 0 and variance $\sigma^2$, $J$ is the noisy image.

\noindent (2) \textbf{Poisson noise} is a kind of signal-dependent noise that occurs during the acquisition of digital images.
We amplified the noise using different scaling factor $\alpha$ using the equation $J = I+n*\alpha$, where we generate Poisson noise $n$ first, then multiply it by a scaling factor $\alpha$.

\noindent (3) \textbf{Spatially-correlated noise} indicates additive Gaussian noise filtered with an average kernel of size $3\times3$. Different levels indicate different standard deviations $\sigma$ for the used Gaussian noise. This is to synthesize the complex artifact after denoising using a flawed algorithm.

\noindent (4) \textbf{Salt \& pepper noise}. Different noise levels represent different noise densities, denoted by $d$. The \textit{imnoise} function in MATLAB is used for generating Salt \& pepper noise. This noise can appear during image acquisition as a result of camera imaging pipeline errors.

\noindent (5) \textbf{Image signal processing (ISP) noise}. 
Modern digital cameras aim to produce visually pleasing and accurate images that match human perception. The raw sensor data captured by the camera cannot directly produce a usable image, and several post-processing stages are required to convert its linear intensities into the final image \cite{brooks2019unprocessing}. As the original raw image contains noise, the post-processed image exhibits more complex noise. Since there are no adequate real noisy and noise-free image pairs, many denoising algorithms perform poorly on real data due to the gap between synthetic and real noise. In our experiments, we use the default parameter settings of \cite{brooks2019unprocessing} to synthesize ISP noise on RGB images.

\noindent (6) \textbf{Mixture noise} is obtained by mixing the above different types of noise with different levels.
We consider the real-world case where the image suffers from multiple degradations.
The order of noise adding is Gaussian noise (variances $\sigma^2_g$), speckle noise (variances $\sigma^2_{s1}$), Poisson noise (scale $\alpha$), Salt \& pepper noise (density $d$), speckle noise (variances $\sigma^2_{s2}$).
Since speckle noise is a multiplicative noise, it will have different effects when used in different positions.
It will be multiplied by the noise already existing in the image to obtain complex noise degradation.
There are 4 levels: 
\begin{enumerate}
    \item $\sigma^2_g=0.003$, $\sigma^2_{s1}=0.003$, $\alpha=1$,  $d=0.002$, $\sigma^2_{s2}=0.003$;
    \item $\sigma^2_g=0.004$, $\sigma^2_{s1}=0.004$, $\alpha=1$,  $d=0.002$, $\sigma^2_{s2}=0.004$;
    \item $\sigma^2_g=0.006$, $\sigma^2_{s1}=0.006$, $\alpha=1$,  $d=0.003$, $\sigma^2_{s2}=0.006$;
    \item $\sigma^2_g=0.008$, $\sigma^2_{s1}=0.008$, $\alpha=1$,  $d=0.004$, $\sigma^2_{s2}=0.008$;
\end{enumerate}
%
The noise patterns produced by these four settings are completely different from existing studies.

We also include two real noise types in this work: the Smartphone Image Denoising Dataset (SIDD)~\cite{abdelhamed2018high} and Monte Carlo (MC) rendered image noise \cite{firmino2022progressive}.

\begin{figure}[t]
  \centering
   \includegraphics[width=\linewidth]{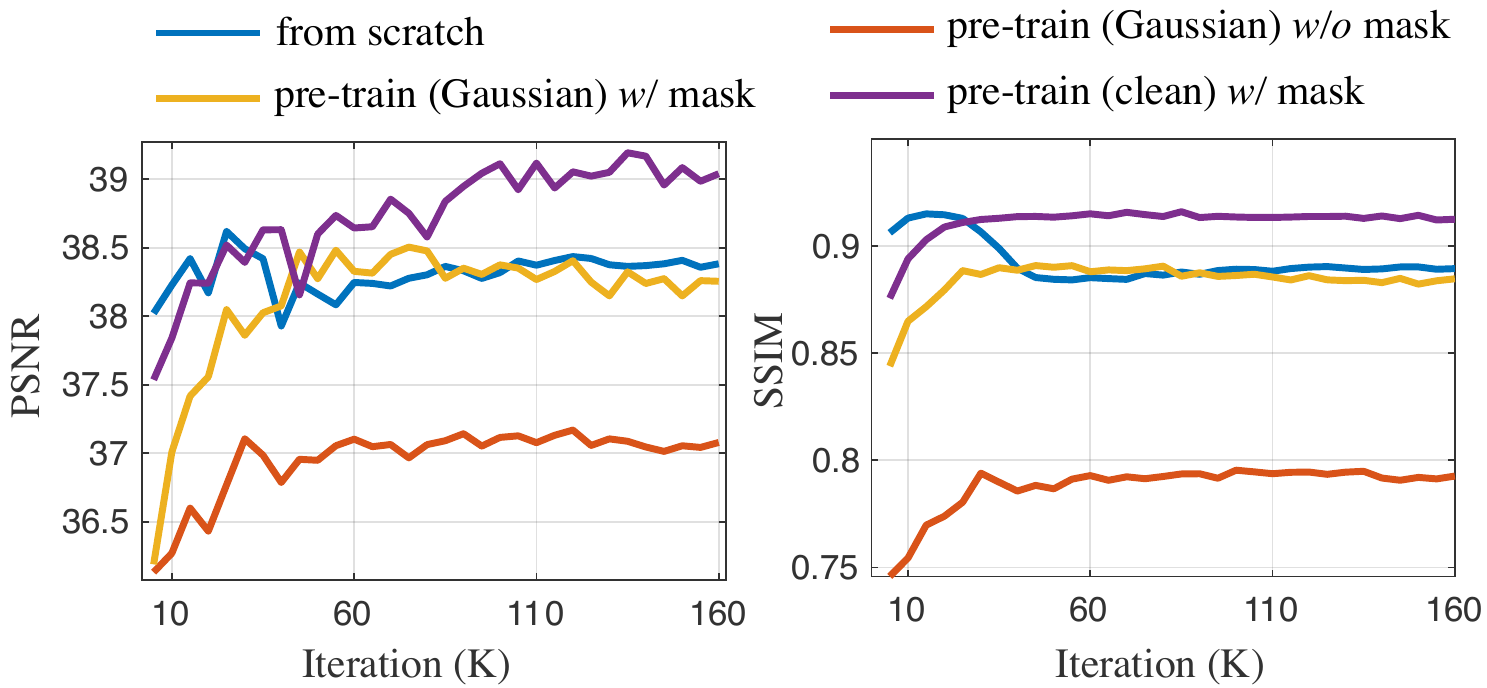}
   \caption{Training curve of different methods validated using our SIDD testset.}
   \label{fig:siddcurve}
\end{figure}

\section{Additional Comparisons}
\label{sec:comparisons}

\paragraph{Methods for Comparison.}
We compare our method with several classical methods: 
DnCNN~\cite{zhang2017beyond},       
RIDNet~\cite{anwar2019real},         
RNAN~\cite{zhang2019residual},        
SwinIR~\cite{liang2021swinir},       
Restormer~\cite{zamir2022restormer},  
Dropout~\cite{kong2022reflash}.   
Among them, Dropout~\cite{kong2022reflash} was proposed to improve the generalization ability and relieve the overfitting problem.
Following ~\cite{kong2022reflash}, we apply the dropout layer with a dropout probability of 0.7 before the output convolutional layer of the baseline model.

\begin{table}[t]
  \centering
  \footnotesize 
\resizebox{\columnwidth}{!}{%
\begin{tabular}{cccccccc}
\toprule
\rowcolor{color3} ID & Pre-train  & \begin{tabular}[c]{@{}c@{}}SIDD\\ Fine-tune\end{tabular} & \begin{tabular}[c]{@{}c@{}}Masked \\ Traning\end{tabular} & PSNR & SSIM & LPIPS \\ \midrule
1 & Gaus. 15       &              &              & 32.11 & 0.6606 & 0.5434 \\ 
2 & Gaus. 15       &              & $\checkmark$ & 33.01 & 0.6999 & 0.4626 \\ \midrule
3 &     None       & $\checkmark$ &              & 38.36 & 0.8879 & 0.3555 \\
4 & Gaus. 15       & $\checkmark$ &              & 37.08 & 0.7920 & 0.3622 \\
5 & Gaus. 15       & $\checkmark$ & $\checkmark$ & 38.15 & 0.8822 & 0.3237 \\ \midrule
6 & Clean          & $\checkmark$ & $\checkmark$ & \textbf{39.11} & \textbf{0.9135} & \textbf{0.2614} \\ 
\bottomrule
\end{tabular}
}
\vspace{-3mm}
  \caption{Masked pre-training for limited paired data. 
  Our method of pre-training on clean images by masked training first and then fine-tuning on target limited dataset yields the best results.}
  \label{tab:siddexpt}
\vspace{-3mm}
\end{table}

\begin{figure*}[t]
\scriptsize
\centering
\resizebox{0.95\textwidth}{!}{
\begin{tabular}{cc}
\\
\hspace{-0.4cm}
\begin{adjustbox}{valign=t}
\begin{tabular}{c}
\includegraphics[width=0.206\textwidth]{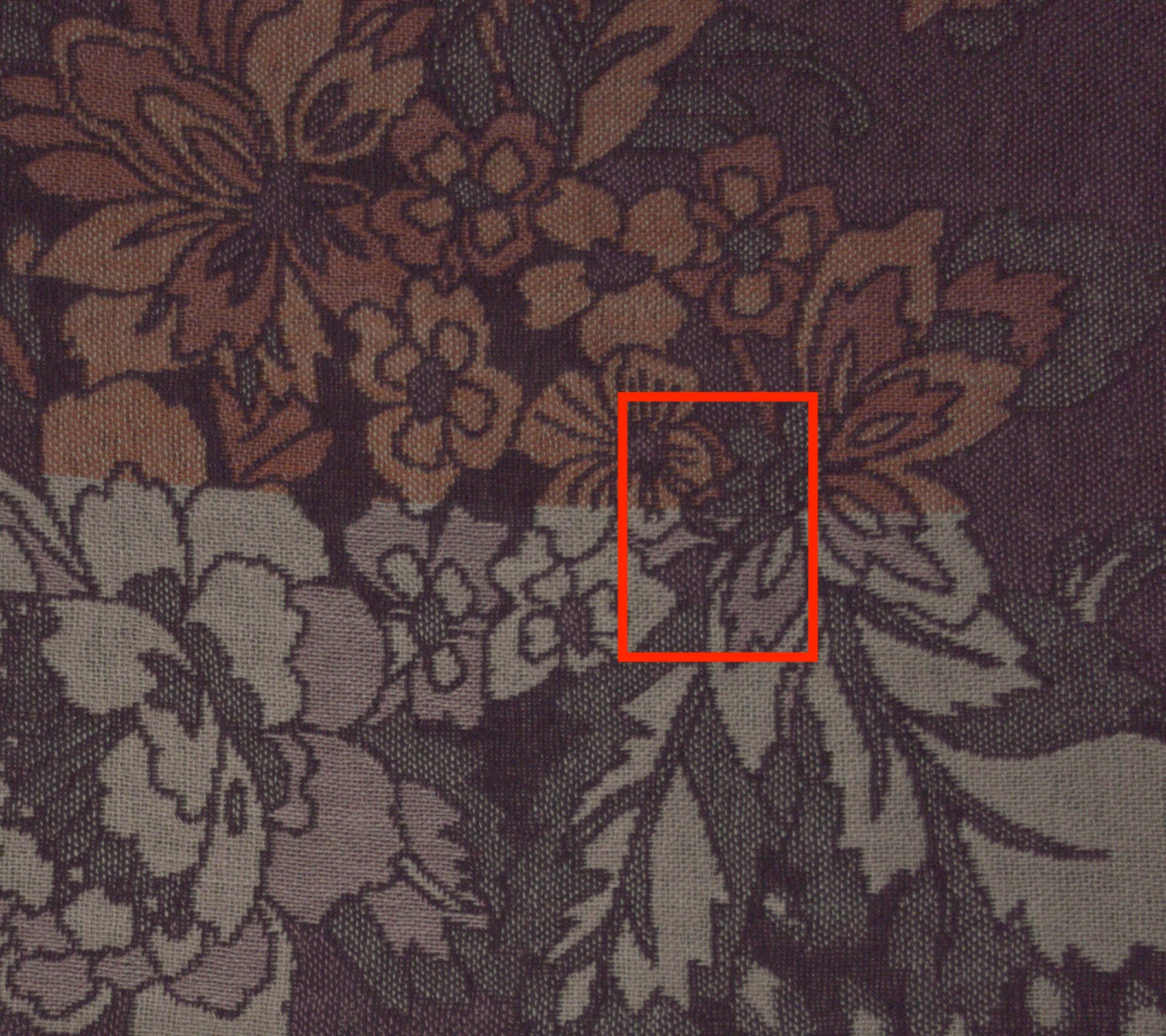}
\\
{\fontsize{6pt}{0}\selectfont {010\_0179\_008\_S6\_03200\_00800\_5500\_L}}
\end{tabular}
\end{adjustbox}
\hspace{-0.46cm}
\begin{adjustbox}{valign=t}
\begin{tabular}{cccccc}
\includegraphics[width=0.124\textwidth]{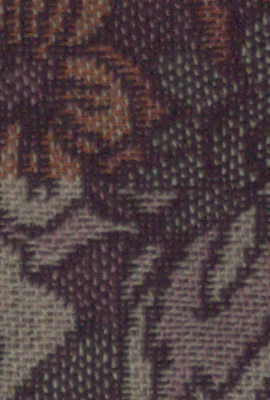} \hspace{-3.5mm} &
\includegraphics[width=0.124\textwidth]{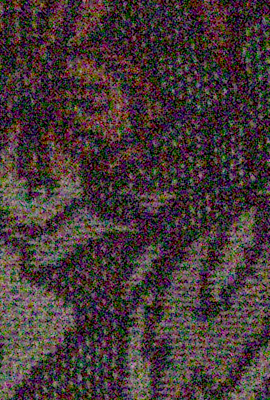} \hspace{-3.5mm} &
\includegraphics[width=0.124\textwidth]{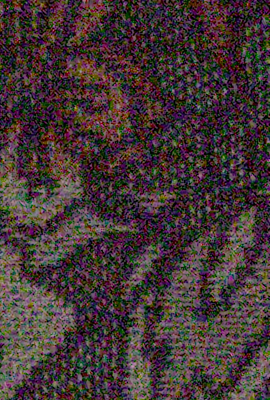} \hspace{-3.5mm} &
\includegraphics[width=0.124\textwidth]{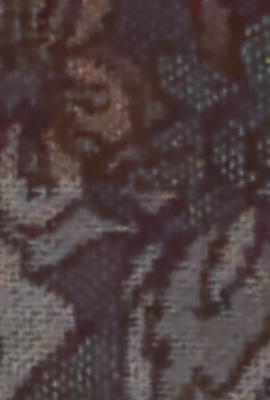} \hspace{-3.5mm} &
\includegraphics[width=0.124\textwidth]{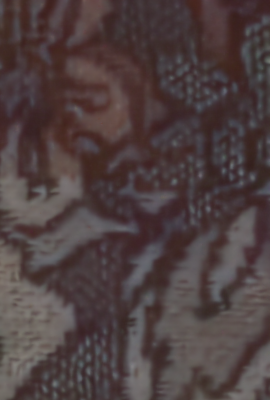} \hspace{-3.5mm} &
\includegraphics[width=0.124\textwidth]{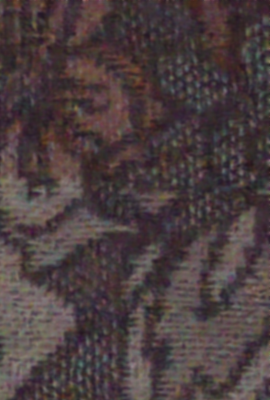} \hspace{-3.5mm} 
\\
HQ \hspace{-4mm} &
Noisy \hspace{-4mm} &
SwinIR \hspace{-4mm} &
from scratch  \hspace{-4mm} &
pre-train \textit{w/o} mask\hspace{-4mm} &
pre-train \textit{w/} mask \hspace{-4mm}
\\
\end{tabular}
\end{adjustbox}
\end{tabular}}
\vspace{-3mm}
\caption{Visual comparison of different methods on real smartphone noise dataset SIDD \cite{abdelhamed2018high}. 
``SwinIR'' is trained on Gaussian noise, $\sigma=15$. "from scratch" is trained directly on the target two SIDD training samples.
``pre-train \textit{w/o} mask'' is pre-trained on Gaussian noise, $\sigma=15$, and fine-tuned without mask.
``pre-train \textit{w/} mask'' is pre-trained on clean images and fine-tuned by masked training.
}
\label{fig:sidd}
\vspace{-2mm}
\end{figure*}

\begin{figure*}[t]
  \centering
  \includegraphics[width=\linewidth]{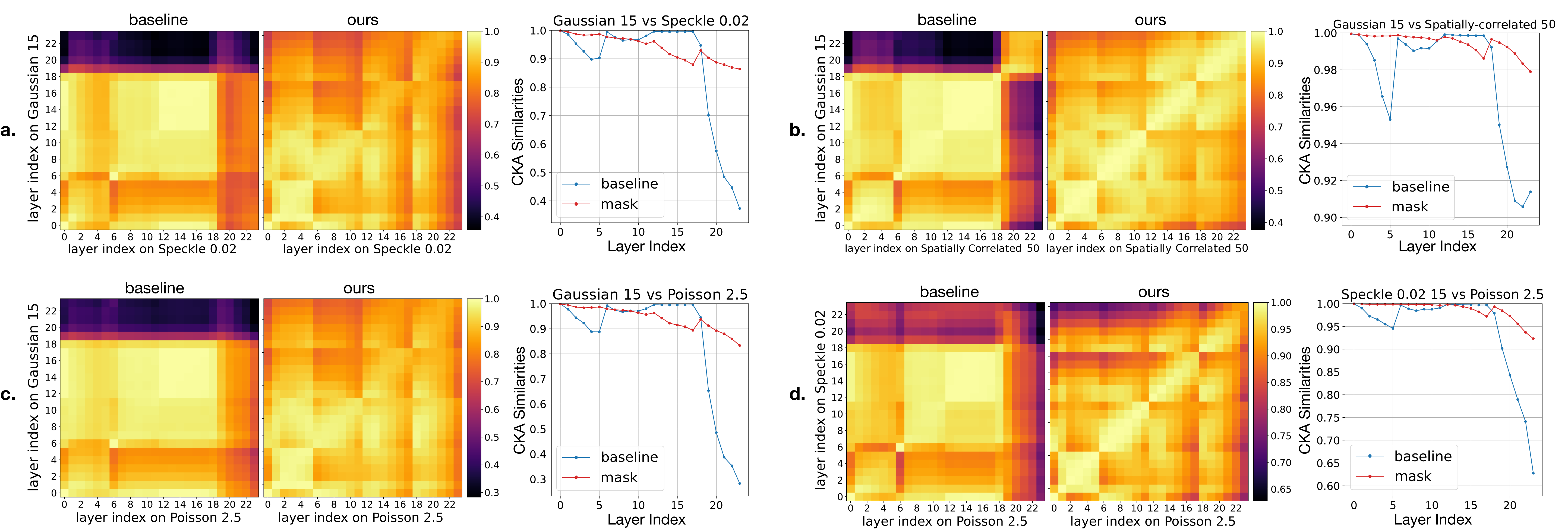}
  \caption{CKA similarity to analyze the representation similarity of network layers.}
  \label{fig:ckasupp}
\end{figure*}

\paragraph{Masked Training as Pre-training.}
In many real-world scenarios, we can only access very limited image pairs for training. It is not enough to adequately train a denoising network because the network can easily overfit the training data.
The performance of the network will be limited if it is trained only on limited data.
The pre-training and fine-tuning paradigm may be helpful in this case.
One approach is to train the network on the synthetic data first and then fine-tune it on the target data~\cite{zhang2017beyond}, but the performance may also be unsatisfactory because of the gap between the pre-train data and the target data.
In this paragraph, we will introduce a practical approach that uses the masked training method for pre-training.
We first pre-train the model on clean images with the masked training strategy, and then fine-tune the model on the limited real training samples with the mask.
This allows the model to obtain generalization ability even when trained on extremely limited training data.
Pre-training on clean images enables the network to learn the content representation of natural images and thus benefits the fine-tuning of target noise.
To conduct such experiments, we use images from the SIDD dataset \cite{abdelhamed2018high}. SIDD contains real noisy images with high-quality clean references.
Due to different lighting and different cameras, the noise of the image is also different. It is consistent with the complex noise situation in the real world.
In order to simulate a scenario with extremely limited training samples, the training set only contains two 4K noisy -- clean image pairs from SIDD.
We also selected one image from each of the ten scenes, for a total of ten images as a test set.
\tablename~\ref{tab:siddexpt} shows the experiment settings and results.
For experiment 3, we directly train the model on the limited training samples.
For experiment 4 and 5, we first pre-train the models using Gaussian noise with $\sigma=15$ and then fine-tune them on target noise.
While for experiment 6, we pre-trained the model on clean (noise-free) images with the proposed masked training strategy, and then fine-tuned it on the target training samples.
The model pre-trained on clean images using the proposed masked training achieves the best results.
This demonstrates the potential of our approach as a new low-level pre-training method.
In addition, our method pre-trained on noisy images is not as effective as pre-trained on clean images, which illustrates that our method benefits from learning information about the image's distribution. 
Visual results are shown in \figurename~\ref{fig:sidd}. Our method preserves the most texture detail.
\figurename~\ref{fig:siddcurve} shows the training curves for different experiments.
The numerical performance of the model pre-trained on Gaussian noise and fine-tuned without masking (red line) is generally low and does not increase with training.
For the model trained from scratch directly on SIDD (blue line), its PSNR starts to fluctuate at the beginning of training and does not improve any further. Its SSIM even drops with training.
This indicates a severe overfitting problem.
In contrast, the method using the proposed masked training (purple and yellow lines) can continue to improve the performance during the training process. This indicates that the model has not yet had an overfitting problem.
The method pre-trained with clean images (purple line) performs better.

\paragraph{Quantitative Comparison.}
We provide full numerical results in \tablename~\ref{tab:kodak24}, \tablename~\ref{tab:cbsd68}, \tablename~\ref{tab:mcm}, and \tablename~\ref{tab:urban100}, where we evaluate our method on four benchmark datasets, namely CBSD68~\cite{martin2001database}, Kodak24~\cite{kodak24}, McMaster~\cite{zhang2011color}, and Urban100~\cite{huang2015single}.
Our method outperforms other state-of-the-art models significantly across all noise types. Particularly, we obtain a significant lead in LPIPS performance, suggesting that our results have better human visual perceptual quality.

\begin{figure*}[!th]
\scriptsize
\centering
\begin{tabular}{ccc}
\hspace{-0.45cm}
\begin{adjustbox}{valign=t}
\begin{tabular}{c}
\includegraphics[width=0.253\textwidth]{figs/visual/sp002/CBSD68_0067.png}
\\
CBSD68: img\_0067
\end{tabular}
\end{adjustbox}
\hspace{-0.46cm}
\begin{adjustbox}{valign=t}
\begin{tabular}{cccccc}
\includegraphics[width=0.18\textwidth]{figs/visual/sp002/LR.png} \hspace{-4mm} &
\includegraphics[width=0.18\textwidth]{figs/visual/sp002/dncnn.png} \hspace{-4mm} &
\includegraphics[width=0.18\textwidth]{figs/visual/sp002/ridnet.png} \hspace{-4mm} &
\includegraphics[width=0.18\textwidth]{figs/visual/sp002/rnan.png} \hspace{-4mm} 
\\
{\fontsize{6pt}{0}\selectfont {Salt-and-pepper noise, $d = 0.02$}}  \hspace{-4mm} &
DnCNN~\cite{zhang2017beyond} \hspace{-4mm} &
RIDNet~\cite{anwar2019real} \hspace{-4mm} &
RNAN~\cite{zhang2019residual} \hspace{-4mm}
\\
\includegraphics[width=0.18\textwidth]{figs/visual/sp002/Restormer.png} \hspace{-4mm} &
\includegraphics[width=0.18\textwidth]{figs/visual/sp002/SwinIR.png} \hspace{-4mm} &
\includegraphics[width=0.18\textwidth]{figs/visual/sp002/baseline.png} \hspace{-4mm} &
\includegraphics[width=0.18\textwidth]{figs/visual/sp002/mask.png} \hspace{-4mm}  
\\ 
Restormer~\cite{zamir2022restormer} \hspace{-4mm} &
SwinIR~\cite{liang2021swinir} \hspace{-4mm} &
baseline  \hspace{-4mm} &
Masked Training \hspace{-4mm}
\\
\end{tabular}
\end{adjustbox}
\\
\hspace{-0.45cm}
\begin{adjustbox}{valign=t}
\begin{tabular}{c}
\includegraphics[width=0.253\textwidth]{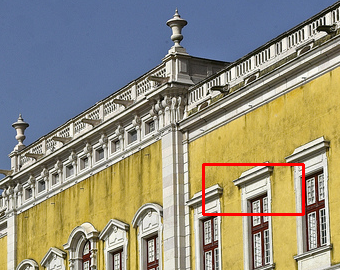}
\\
urban100: img\_054
\end{tabular}
\end{adjustbox}
\hspace{-0.46cm}
\begin{adjustbox}{valign=t}
\begin{tabular}{cccccc}
\includegraphics[width=0.18\textwidth]{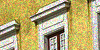} \hspace{-4mm} &
\includegraphics[width=0.18\textwidth]{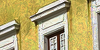} \hspace{-4mm} &
\includegraphics[width=0.18\textwidth]{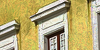} \hspace{-4mm} &
\includegraphics[width=0.18\textwidth]{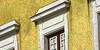} \hspace{-4mm} 
\\
Speckle noise, $\sigma^2=0.016$ \hspace{-4mm} &
DnCNN~\cite{zhang2017beyond} \hspace{-4mm} &
RIDNet~\cite{anwar2019real} \hspace{-4mm} &
RNAN~\cite{zhang2019residual} \hspace{-4mm}
\\
\includegraphics[width=0.18\textwidth]{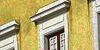} \hspace{-4mm} &
\includegraphics[width=0.18\textwidth]{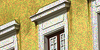} \hspace{-4mm} &
\includegraphics[width=0.18\textwidth]{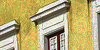} \hspace{-4mm} &
\includegraphics[width=0.18\textwidth]{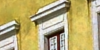} \hspace{-4mm}  
\\ 
Restormer~\cite{zamir2022restormer} \hspace{-4mm} &
SwinIR~\cite{liang2021swinir} \hspace{-4mm} &
baseline  \hspace{-4mm} &
Masked Training \hspace{-4mm}
\\
\end{tabular}
\end{adjustbox}
\\
\hspace{-0.45cm}
\begin{adjustbox}{valign=t}
\begin{tabular}{c}
\includegraphics[width=0.253\textwidth]{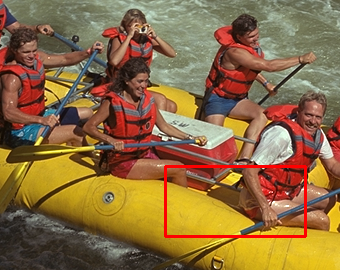}
\\
kodak24: img\_14
\end{tabular}
\end{adjustbox}
\hspace{-0.46cm}
\begin{adjustbox}{valign=t}
\begin{tabular}{cccccc}
\includegraphics[width=0.18\textwidth]{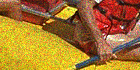} \hspace{-4mm} &
\includegraphics[width=0.18\textwidth]{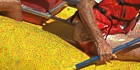} \hspace{-4mm} &
\includegraphics[width=0.18\textwidth]{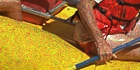} \hspace{-4mm} &
\includegraphics[width=0.18\textwidth]{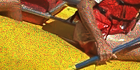} \hspace{-4mm} 
\\
Poisson noise 2 \hspace{-4mm} &
DnCNN~\cite{zhang2017beyond} \hspace{-4mm} &
RIDNet~\cite{anwar2019real} \hspace{-4mm} &
RNAN~\cite{zhang2019residual} \hspace{-4mm}
\\
\includegraphics[width=0.18\textwidth]{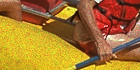} \hspace{-4mm} &
\includegraphics[width=0.18\textwidth]{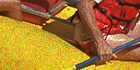} \hspace{-4mm} &
\includegraphics[width=0.18\textwidth]{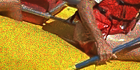} \hspace{-4mm} &
\includegraphics[width=0.18\textwidth]{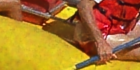} \hspace{-4mm}  
\\ 
Restormer~\cite{zamir2022restormer} \hspace{-4mm} &
SwinIR~\cite{liang2021swinir} \hspace{-4mm} &
baseline  \hspace{-4mm} &
Masked Training \hspace{-4mm}
\\
\end{tabular}
\end{adjustbox}
\\
\hspace{-0.45cm}
\begin{adjustbox}{valign=t}
\begin{tabular}{c}
\includegraphics[width=0.253\textwidth]{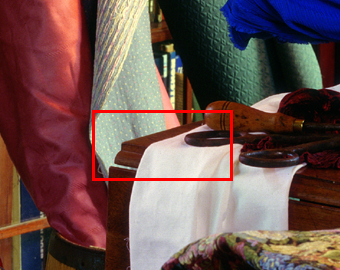}
\\
McM: img\_2
\end{tabular}
\end{adjustbox}
\hspace{-0.46cm}
\begin{adjustbox}{valign=t}
\begin{tabular}{cccccc}
\includegraphics[width=0.18\textwidth]{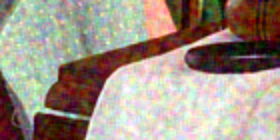} \hspace{-4mm} &
\includegraphics[width=0.18\textwidth]{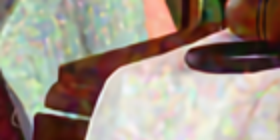} \hspace{-4mm} &
\includegraphics[width=0.18\textwidth]{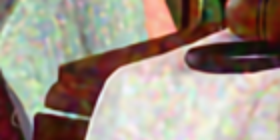} \hspace{-4mm} &
\includegraphics[width=0.18\textwidth]{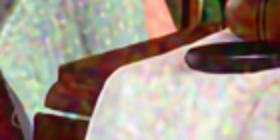} \hspace{-4mm} 
\\
{\fontsize{5pt}{0}\selectfont {Spatially-correlated noise, $\sigma=45$}}    \hspace{-4mm} &
DnCNN~\cite{zhang2017beyond} \hspace{-4mm} &
RIDNet~\cite{anwar2019real} \hspace{-4mm} &
RNAN~\cite{zhang2019residual} \hspace{-4mm}
\\
\includegraphics[width=0.18\textwidth]{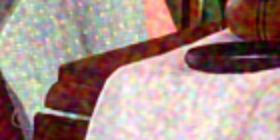} \hspace{-4mm} &
\includegraphics[width=0.18\textwidth]{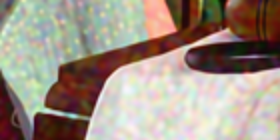} \hspace{-4mm} &
\includegraphics[width=0.18\textwidth]{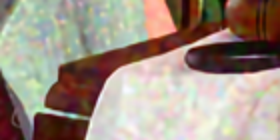} \hspace{-4mm} &
\includegraphics[width=0.18\textwidth]{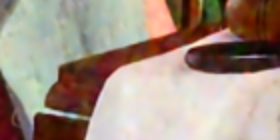} \hspace{-4mm}  
\\ 
Restormer~\cite{zamir2022restormer} \hspace{-4mm} &
SwinIR~\cite{liang2021swinir} \hspace{-4mm} &
baseline  \hspace{-4mm} &
Masked Training \hspace{-4mm}
\\
\end{tabular}
\end{adjustbox}
\\
\hspace{-0.45cm}
\begin{adjustbox}{valign=t}
\begin{tabular}{c}
\includegraphics[width=0.253\textwidth]{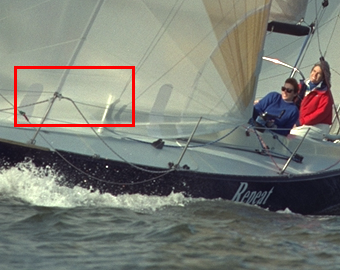}
\\
kodak24: img\_10
\end{tabular}
\end{adjustbox}
\hspace{-0.46cm}
\begin{adjustbox}{valign=t}
\begin{tabular}{cccccc}
\includegraphics[width=0.18\textwidth]{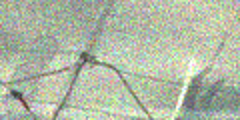} \hspace{-4mm} &
\includegraphics[width=0.18\textwidth]{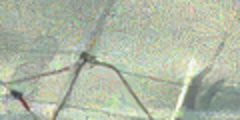} \hspace{-4mm} &
\includegraphics[width=0.18\textwidth]{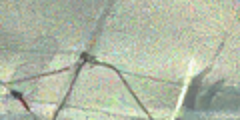} \hspace{-4mm} &
\includegraphics[width=0.18\textwidth]{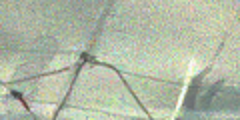} \hspace{-4mm} 
\\
Poisson noise, $\alpha=1.7$  \hspace{-4mm} &
DnCNN~\cite{zhang2017beyond} \hspace{-4mm} &
RIDNet~\cite{anwar2019real} \hspace{-4mm} &
RNAN~\cite{zhang2019residual} \hspace{-4mm}
\\
\includegraphics[width=0.18\textwidth]{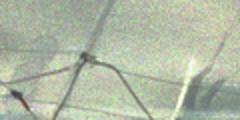} \hspace{-4mm} &
\includegraphics[width=0.18\textwidth]{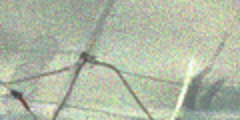} \hspace{-4mm} &
\includegraphics[width=0.18\textwidth]{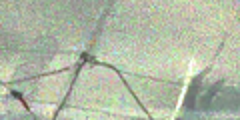} \hspace{-4mm} &
\includegraphics[width=0.18\textwidth]{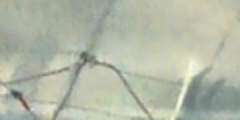} \hspace{-4mm}  
\\ 
Restormer~\cite{zamir2022restormer} \hspace{-4mm} &
SwinIR~\cite{liang2021swinir} \hspace{-4mm} &
baseline  \hspace{-4mm} &
Masked Training \hspace{-4mm}
\\
\end{tabular}
\end{adjustbox}
\\
\hspace{-0.45cm}
\begin{adjustbox}{valign=t}
\begin{tabular}{c}
\includegraphics[width=0.253\textwidth]{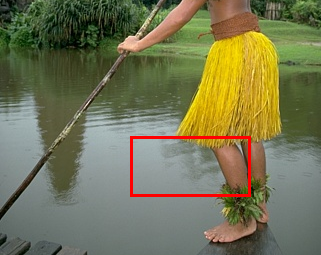}
\\
CBSD68: img\_0009
\end{tabular}
\end{adjustbox}
\hspace{-0.46cm}
\begin{adjustbox}{valign=t}
\begin{tabular}{cccccc}
\includegraphics[width=0.18\textwidth]{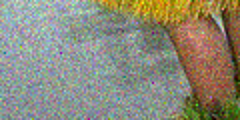} \hspace{-4mm} &
\includegraphics[width=0.18\textwidth]{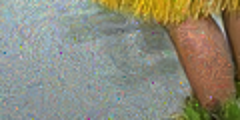} \hspace{-4mm} &
\includegraphics[width=0.18\textwidth]{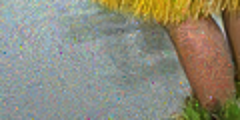} \hspace{-4mm} &
\includegraphics[width=0.18\textwidth]{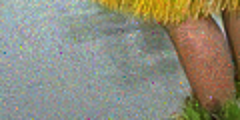} \hspace{-4mm} 
\\
Mixture noise, level $1$  \hspace{-4mm} &
DnCNN~\cite{zhang2017beyond} \hspace{-4mm} &
RIDNet~\cite{anwar2019real} \hspace{-4mm} &
RNAN~\cite{zhang2019residual} \hspace{-4mm}
\\
\includegraphics[width=0.18\textwidth]{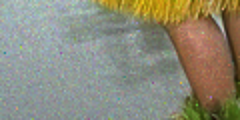} \hspace{-4mm} &
\includegraphics[width=0.18\textwidth]{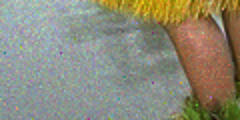} \hspace{-4mm} &
\includegraphics[width=0.18\textwidth]{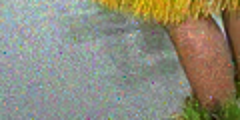} \hspace{-4mm} &
\includegraphics[width=0.18\textwidth]{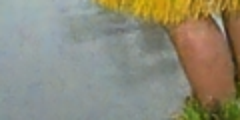} \hspace{-4mm}  
\\ 
Restormer~\cite{zamir2022restormer} \hspace{-4mm} &
SwinIR~\cite{liang2021swinir} \hspace{-4mm} &
baseline  \hspace{-4mm} &
Masked Training \hspace{-4mm}
\\
\end{tabular}
\end{adjustbox}

\end{tabular}
\vspace{-1mm}
\caption{Visual comparison.}
\label{fig:visual}
\vspace{-5mm}
\end{figure*}

\paragraph{Additional Visual Results.}
\figurename~\ref{fig:visual} shows more visual comparisons. 
The model's performance without masked training is significantly limited over the various noise types.
Our model still effectively removes noise when dealing with a variety of noise outside the training set.

\section{Additional Analyses of CKA}
\label{sec:cka}
In the main text, in order to investigate how masked training differs from normal training strategy, we utilize the centered kernel alignment (CKA) \cite{cortes2012algorithms,raghu2021vision} to analyze the differences between network representations obtained from those two training methods.
In detail, we calculate the representations of two layers $\mathbf{X} \in \mathbb{R}^{m \times p_1}$ and $\mathbf{Y} \in \mathbb{R}^{m \times p_2}$ on the same $m$ data points, with $p_1$ and $p_2$ neurons respectively. Gram matrices $\mathbf{K}=\mathbf{X X}^{\top}$ and $\mathbf{L}=\mathbf{Y} \mathbf{Y}^{\top}$ are used to compute CKA:
$$
\operatorname{CKA}(\mathbf{K}, \mathbf{L})=\frac{\operatorname{HSIC}(\mathbf{K}, \mathbf{L})}{\sqrt{\operatorname{HSIC}(\mathbf{K}, \mathbf{K}) \mathrm{HSIC}(\mathbf{L}, \mathbf{L})}}
$$
where HSIC is the Hilbert-Schmidt independence criterion \cite{kornblith2019similarity}. Given the centering matrix $\mathbf{H}=\mathbf{I}_n-\frac{1}{n} \mathbf{1 1}^{\top}$, and centered Gram matrices $\mathbf{K}^{\prime}=\mathbf{H K H}$ and  $\mathbf{L}^{\prime}=\mathbf{H L H}$, we have $\operatorname{HSIC}(\mathbf{K}, \mathbf{L})=\operatorname{vec}\left(\mathbf{K}^{\prime}\right) \cdot \operatorname{vec}\left(\mathbf{L}^{\prime}\right) /(m-1)^2$. 
More CKA results are shown in \figurename~\ref{fig:ckasupp}.
We first compare the correlation of the features between different noise types.
For the baseline model, the correlation between the features of Gaussian noise and other different noises at the deep level is relatively low (a, b, c).
Besides, the feature correlation between the noise outside the training set is also low (d).
The model using the proposed masked training is able to have a high correlation in all cases.
\figurename~\ref{fig:ckasupp} (a) shows the cross-model comparison between baseline and masked training models.
We find that a significant difference between the two is that the features of the deeper layers of the baseline model have low correlations with all layers of our model.
This indicates that these two training methods have inconsistent learning patterns for features, especially for the deeper layers.
To explore how the model performs on different noise, \figurename~\ref{fig:ckasupp} (b) shows the cross-noise comparison between in-distribution noise and out-of-distribution noise (Gaussian and Poisson noise).
For the baseline model, there is a low correlation between the different noise in the deep layers.
It shows that the network processes these two types of noise differently for the deep layers.
The other types of noise share a similar phenomenon.
We suggest that this is because the baseline approach makes the deep layer of the model focus on overfitting the patterns of the training set, which leads to the poor generalization of the deep layers to handle different noise.
In our model, the correlation between adjacent layers in our model is high.
The proposed masked training forces the network to learn the distribution of the images themselves, which is similar to different types of noise.
This allows our method to have a stronger generalization capability.


\begin{table*}[b]
\centering
\footnotesize 
\label{tab:my-table}

\begin{tabular}{lccc|ccc|ccc|ccc}
\toprule 
    \rowcolor{color3} {\textbf{Speckle noise}} 
    & \multicolumn{3}{c}{\textbf{$\sigma^2=0.02$}}
    & \multicolumn{3}{c}{\textbf{$\sigma^2=0.024$}} 
    & \multicolumn{3}{c}{\textbf{$\sigma^2=0.03$}} 
    & \multicolumn{3}{c}{\textbf{$\sigma^2=0.04$}}     \\ 
    \rowcolor{color3} \textbf{Method}        & PSNR    & SSIM     & LPIPS   & PSNR   & SSIM   & LPIPS  & PSNR   & SSIM   & LPIPS  & PSNR   & SSIM   & LPIPS \\ 
\midrule
DnCNN~\cite{zhang2017beyond}         & 30.74  & 0.8281  & 0.1806     & 29.31  & 0.7891  & 0.2082     & 27.49  & 0.7353  & 0.2533     & 25.22  & 0.6620  & 0.3292 \\
RIDNet~\cite{anwar2019real}          & 31.01  & 0.8337  & 0.1665     & 29.51  & 0.7916  & 0.1944     & 27.57  & 0.7331  & 0.2436     & 25.17  & 0.6554  & 0.3212 \\
RNAN~\cite{zhang2019residual}        & 30.15  & 0.8101  & 0.1660     & 28.59  & 0.7662  & 0.1972     & 26.76  & 0.7101  & 0.2449     & 24.59  & 0.6377  & 0.3203 \\
SwinIR~\cite{liang2021swinir}        & 29.64  & 0.7939  & 0.1555     & 28.16  & 0.7514  & 0.1851     & 26.43  & 0.6981  & 0.2305     & 24.37  & 0.6298  & 0.3004 \\
Restormer~\cite{zamir2022restormer}  & 29.95  & 0.8135  & 0.1521     & 28.84  & 0.7810  & 0.1767     & 27.50  & 0.7395  & 0.2113     & 25.66  & 0.6839  & 0.2649 \\
Dropout~\cite{kong2022reflash}       & 29.97  & 0.8382  & 0.1709     & 29.03  & 0.8041  & 0.1974     & 27.77  & 0.7570  & 0.2413     & 26.14  & 0.6925  & 0.3110 \\
baseline                             & 29.84  & 0.8016  & 0.1778     & 28.34  & 0.7608  & 0.2082     & 26.56  & 0.7071  & 0.2536     & 24.44  & 0.6367  & 0.3242 \\ 
\midrule
\textbf{Ours} & \cellcolor{green!15}\textbf{31.22}   & \cellcolor{green!15}\textbf{0.8739}   & \cellcolor{green!15}\textbf{0.1594}  
              & \cellcolor{green!15}\textbf{30.81}   & \cellcolor{green!15}\textbf{0.8617}   & \cellcolor{green!15}\textbf{0.1683} 
              & \cellcolor{green!15}\textbf{30.20}   & \cellcolor{green!15}\textbf{0.8412}   & \cellcolor{green!15}\textbf{0.1849} 
              & \cellcolor{green!15}\textbf{29.10}   & \cellcolor{green!15}\textbf{0.8000}   & \cellcolor{green!15}\textbf{0.2248} \\ 
\end{tabular}

\begin{tabular}{lccc|ccc|ccc|ccc}
\toprule 
    \rowcolor{color3} {\textbf{Poisson noise}} 
    & \multicolumn{3}{c}{\textbf{$\alpha=2$}} 
    & \multicolumn{3}{c}{\textbf{$\alpha=2.5$}} 
    & \multicolumn{3}{c}{\textbf{$\alpha=3$}} 
    & \multicolumn{3}{c}{\textbf{$\alpha=3.5$}}     \\ 
    \rowcolor{color3} \textbf{Method}        & PSNR    & SSIM     & LPIPS   & PSNR   & SSIM   & LPIPS  & PSNR   & SSIM   & LPIPS  & PSNR   & SSIM   & LPIPS \\ 
\midrule
DnCNN~\cite{zhang2017beyond}         	&	28.41	  &	0.7359	  &	0.2284	&	24.38	  &	0.5767	  &	0.3887	&	21.63	  &	0.4571	  &	0.5330	&	19.65	  &	0.3711	  &	0.6521	\\
RIDNet~\cite{anwar2019real}          	&	28.17	  &	0.7231	  &	0.2215	&	24.00	  &	0.5546	  &	0.3849	&	21.34	  &	0.4379	  &	0.5246	&	19.48	  &	0.3567	  &	0.6397	\\
RNAN~\cite{zhang2019residual}        	&	27.55	  &	0.7000	  &	0.2231	&	23.66	  &	0.5402	  &	0.3783	&	21.14	  &	0.4263	  &	0.5184	&	19.33	  &	0.3486	  &	0.6355	\\
SwinIR~\cite{liang2021swinir}        	&	27.32	  &	0.6877	  &	0.2081	&	23.68	  &	0.5398	  &	0.3487	&	21.17	  &	0.4294	  &	0.4860	&	19.32	  &	0.3506	  &	0.6059	\\
Restormer~\cite{zamir2022restormer}  	&	29.22	  &	0.7639	  &	0.1662	&	26.11	  &	0.6452	  &	0.2608	&	23.98	  &	0.5613	  &	0.3530	&	22.55	  &	0.5174	  &	0.4306	\\
Dropout~\cite{kong2022reflash}       	&	28.47	  &	0.7601	  &	0.2209	&	25.61	  &	0.6245	  &	0.3652	&	23.53	  &	0.5218	  &	0.4986	&	21.97	  &	0.4454	  &	0.6136	\\
baseline                             	&	27.70	  &	0.7040	  &	0.2339	&	23.85	  &	0.5524	  &	0.3782	&	21.27	  &	0.4377	  &	0.5109	&	19.45	  &	0.3550	  &	0.6241	\\ 
\midrule
\textbf{Ours} & \cellcolor{green!15}\textbf{30.59}   & \cellcolor{green!15}\textbf{0.8510}   & \cellcolor{green!15}\textbf{0.1662}  
              & \cellcolor{green!15}\textbf{28.80}   & \cellcolor{green!15}\textbf{0.7709}   & \cellcolor{green!15}\textbf{0.2488} 
              & \cellcolor{green!15}\textbf{27.04}   & \cellcolor{green!15}\textbf{0.6834}   & \cellcolor{green!15}\textbf{0.3493} 
              & \cellcolor{green!15}\textbf{25.46}   & \cellcolor{green!15}\textbf{0.6039}   & \cellcolor{green!15}\textbf{0.4502} \\ 
\end{tabular}

\begin{tabular}{lccc|ccc|ccc|ccc}
\toprule 
    \rowcolor{color3} {\textbf{{\fontsize{6pt}{0}\selectfont {Spatially-correlated}}}}  
    & \multicolumn{3}{c}{\textbf{$\sigma=40$}} 
    & \multicolumn{3}{c}{\textbf{$\sigma=45$}} 
    & \multicolumn{3}{c}{\textbf{$\sigma=50$}} 
    & \multicolumn{3}{c}{\textbf{$\sigma=55$}}     \\ 
    \rowcolor{color3} \textbf{Method} & PSNR & SSIM & LPIPS & PSNR & SSIM & LPIPS & PSNR & SSIM & LPIPS & PSNR & SSIM & LPIPS \\ 
\midrule
DnCNN~\cite{zhang2017beyond}          &\cellcolor{green!15}29.63&0.8036&0.3527&28.17&0.7474&0.4192&26.85&0.6898&0.4718&25.70&0.6360&0.5173\\
RIDNet~\cite{anwar2019real}           &28.94&0.7766 & 0.4109 & 27.58 & 0.7189 & 0.4746 & 26.39 & 0.6637 & 0.5208 & 25.34 & 0.6131 & 0.5580 \\
RNAN~\cite{zhang2019residual}         & 28.86 & 0.7644 & 0.3943 & 27.50 & 0.7078 & 0.4532 & 26.32 & 0.6542 & 0.4980 & 25.28 & 0.6050 & 0.5373 \\
SwinIR~\cite{liang2021swinir}         & 28.73 & 0.7524 & 0.4056 & 27.38 & 0.6951 & 0.4620 & 26.20 & 0.6414 & 0.5070 & 25.17 & 0.5930 & 0.5458 \\
Restormer~\cite{zamir2022restormer}   & 23.42 & 0.6533 & 0.4412 & 23.06 & 0.6109 & 0.4783 & 22.82 & 0.5709 & 0.5072 & 22.59 & 0.5353 & 0.5356 \\
Dropout~\cite{kong2022reflash}        & 29.35 & 0.8173 & 0.3188 & 28.27 & 0.7719 & 0.3800 & 27.19 & 0.7206 & 0.4400 & 26.19 & 0.6694 & 0.4943 \\
baseline                              & 29.34 & 0.7834 & 0.3706 & 27.82 & 0.7205 & 0.4375 & 26.55 & 0.6628 & 0.4878 & 25.46 & 0.6118 & 0.5295 \\ 
\midrule
\textbf{Ours} & \textbf{29.55}   & \cellcolor{green!15}\textbf{0.8296}   & \cellcolor{green!15}\textbf{0.2949}  
              & \cellcolor{green!15}\textbf{28.84}   & \cellcolor{green!15}\textbf{0.8045}   & \cellcolor{green!15}\textbf{0.3358} 
              & \cellcolor{green!15}\textbf{28.05}   & \cellcolor{green!15}\textbf{0.7735}   & \cellcolor{green!15}\textbf{0.3762} 
              & \cellcolor{green!15}\textbf{27.27}   & \cellcolor{green!15}\textbf{0.7388}   & \cellcolor{green!15}\textbf{0.4163} \\ 
\end{tabular}

\begin{tabular}{lccc|ccc|ccc|ccc}
\toprule 
    \rowcolor{color3} \textbf{Salt \& pepper}
    & \multicolumn{3}{c}{\textbf{$d=0.002$}}
    & \multicolumn{3}{c}{\textbf{$d=0.004$}} 
    & \multicolumn{3}{c}{\textbf{$d=0.008$}} 
    & \multicolumn{3}{c}{\textbf{$d=0.012$}}     \\ 
    \rowcolor{color3} \textbf{Method} & PSNR & SSIM & LPIPS & PSNR & SSIM & LPIPS & PSNR & SSIM & LPIPS & PSNR & SSIM & LPIPS \\ 
\midrule
DnCNN~\cite{zhang2017beyond}         	&	24.75 &	0.6785 &	0.3639	&	21.15 &	0.4952 &	0.5626	&	17.55 &	0.2993 &	0.8196	&	15.47 &	0.2066 &	0.9779	\\
RIDNet~\cite{anwar2019real}          	&	25.19 &	0.6769 &	0.3617	&	21.38 &	0.4934 &	0.5498	&	17.65 &	0.2969 &	0.8029	&	15.60 &	0.2066 &	0.9598	\\
RNAN~\cite{zhang2019residual}        	&	23.59 &	0.6416 &	0.3829	&	20.42 &	0.4639 &	0.5599	&	17.21 &	0.2850 &	0.8048	&	15.31 &	0.2006 &	0.9644	\\
SwinIR~\cite{liang2021swinir}        	&	23.42 &	0.6329 &	0.3873	&	20.21 &	0.4511 &	0.5710	&	17.00 &	0.2688 &	0.8103	&	15.14 &	0.1875 &	0.9614	\\
Restormer~\cite{zamir2022restormer}  	&	23.81 &	0.6384 &	0.3919	&	20.99 &	0.4831 &	0.5551	&	19.79 &	0.3878 &	0.6512	&	19.25 &	0.3257 &	0.7574	\\
Dropout~\cite{kong2022reflash}       	&	27.44 &	0.7180 &	0.3041	&	24.36 &	0.5557 &	0.4898	&	21.01 &	0.3790 &	0.7415	&	19.03 &	0.2902 &	0.9047	\\
baseline                             	&	25.36 &	0.6510 &	0.3694	&	21.93 &	0.4747 &	0.5642	&	18.42 &	0.2939 &	0.8153	&	16.46 &	0.2106 &	0.9656	\\ 
\midrule
\textbf{Ours} & \cellcolor{green!15}\textbf{30.52}   & \cellcolor{green!15}\textbf{0.8477}   & \cellcolor{green!15}\textbf{0.1768}  
              & \cellcolor{green!15}\textbf{28.48}   & \cellcolor{green!15}\textbf{0.7681}   & \cellcolor{green!15}\textbf{0.2786} 
              & \cellcolor{green!15}\textbf{25.01}   & \cellcolor{green!15}\textbf{0.5958}   & \cellcolor{green!15}\textbf{0.5039} 
              & \cellcolor{green!15}\textbf{22.48}   & \cellcolor{green!15}\textbf{0.4622}   & \cellcolor{green!15}\textbf{0.6979} \\ 
\end{tabular}

\begin{tabular}{lccc|ccc|ccc|ccc}
\toprule 
    \rowcolor{color3} {\textbf{{\fontsize{6pt}{0}\selectfont {Mixture noise}}}}  
    & \multicolumn{3}{c}{\textbf{level 1}} 
    & \multicolumn{3}{c}{\textbf{level 2}} 
    & \multicolumn{3}{c}{\textbf{level 3}} 
    & \multicolumn{3}{c}{\textbf{level 4}}     \\ 
    \rowcolor{color3} \textbf{Method}        & PSNR    & SSIM     & LPIPS   & PSNR   & SSIM   & LPIPS  & PSNR   & SSIM   & LPIPS  & PSNR   & SSIM   & LPIPS \\ 
\midrule
DnCNN~\cite{zhang2017beyond}         	&	28.31 &	0.7514 &	0.2299	&	26.53 &	0.6636 &	0.3011	&	23.55 &	0.5117 &	0.4522	&	21.66 &	0.4162 &	0.5622	\\
RIDNet~\cite{anwar2019real}          	&	28.13 &	0.7335 &	0.2215	&	26.11 &	0.6320 &	0.2971	&	23.13 &	0.4776 &	0.4461	&	21.34 &	0.3899 &	0.5514	\\
RNAN~\cite{zhang2019residual}        	&	27.46 &	0.7090 &	0.2280	&	25.67 &	0.6126 &	0.2948	&	22.90 &	0.4657 &	0.4369	&	21.19 &	0.3826 &	0.5431	\\
SwinIR~\cite{liang2021swinir}        	&	27.44 &	0.7049 &	0.2051	&	25.73 &	0.6113 &	0.2682	&	23.03 &	0.4689 &	0.4073	&	21.29 &	0.3847 &	0.5145	\\
Restormer~\cite{zamir2022restormer}  	&	29.23 &	0.7859 &	0.1639	&	28.22 &	0.7330 &	0.1965	&	25.69 &	0.6034 &	0.2894	&	24.05 &	0.5257 &	0.3662	\\
Dropout~\cite{kong2022reflash}       	&	28.61 &	0.7797 &	0.2071	&	27.23 &	0.7039 &	0.2777	&	24.96 &	0.5715 &	0.4290	&	23.49 &	0.4906 &	0.5324	\\
baseline                             	&	28.12 &	0.7295 &	0.2259	&	26.22 &	0.6346 &	0.2985	&	23.28 &	0.4795 &	0.4441	&	21.44 &	0.3885 &	0.5463	\\ 
\midrule
\textbf{Ours} & \cellcolor{green!15}\textbf{30.31}   & \cellcolor{green!15}\textbf{0.8518}   & \cellcolor{green!15}\textbf{0.1617}  
              & \cellcolor{green!15}\textbf{29.63}   & \cellcolor{green!15}\textbf{0.8251}   & \cellcolor{green!15}\textbf{0.1903} 
              & \cellcolor{green!15}\textbf{28.12}   & \cellcolor{green!15}\textbf{0.7513}   & \cellcolor{green!15}\textbf{0.2732} 
              & \cellcolor{green!15}\textbf{26.91}   & \cellcolor{green!15}\textbf{0.6841}   & \cellcolor{green!15}\textbf{0.3530} \\ 
\bottomrule
\end{tabular}

    \caption{Quantitative comparison on Kodak24~\cite{kodak24}.}
    \label{tab:kodak24}
\end{table*}


\begin{table*}[h]
\centering
\footnotesize 
\label{tab:my-table}

\begin{tabular}{lccc|ccc|ccc|ccc}
\toprule 
    \rowcolor{color3} {\textbf{Speckle noise}} 
    & \multicolumn{3}{c}{\textbf{$\sigma^2=0.02$}}
    & \multicolumn{3}{c}{\textbf{$\sigma^2=0.024$}} 
    & \multicolumn{3}{c}{\textbf{$\sigma^2=0.03$}} 
    & \multicolumn{3}{c}{\textbf{$\sigma^2=0.04$}}     \\ 
    \rowcolor{color3} \textbf{Method}        & PSNR    & SSIM     & LPIPS   & PSNR   & SSIM   & LPIPS  & PSNR   & SSIM   & LPIPS  & PSNR   & SSIM   & LPIPS \\ 
\midrule
DnCNN~\cite{zhang2017beyond}         	&   30.67	  &	0.8254	  &	0.1506	&	29.24	  &	0.7927	  &	0.1840	&	27.54	  &	0.7551	  &	0.2269	&	25.49	  &	0.7095	  &	0.2856	\\
RIDNet~\cite{anwar2019real}          	&	\cellcolor{green!15}30.77	  &	\cellcolor{green!15}0.8261	  &	0.1444	&	29.31	  &	0.7934	  &	0.1757	&	27.58	  &	0.7551	  &	0.2168	&	25.49	  &	0.7081	  &	0.2750	\\
RNAN~\cite{zhang2019residual}        	&	29.77	  &	0.8066	  &	0.1492	&	28.32	  &	0.7745	  &	0.1814	&	26.67	  &	0.7377	  &	0.2224	&	24.75	  &	0.6932	  &	0.2796	\\
SwinIR~\cite{liang2021swinir}        	&	29.17	  &	0.7947	  &	0.1258	&	27.83	  &	0.7660	  &	0.1524	&	26.30	  &	0.7322	  &	0.1893	&	24.46	  &	0.6909	  &	0.2412	\\
Restormer~\cite{zamir2022restormer}  	&	28.89	  &	0.8005	  &	0.1300	&	27.95	  &	0.7790	  &	0.1515	&	26.81	  &	0.7523	  &	0.1807	&	25.30	  &	0.7173	  &	0.2213	\\
Dropout~\cite{kong2022reflash}       	&	28.64	  &	0.8153	  &	0.1416	&	27.85	  &	0.7852	  &	0.1688	&	26.89	  &	0.7501	  &	0.2032	&	25.64	  &	0.7062	  &	0.2525	\\
baseline                             	&	28.86	  &	0.7283	  &	0.1353	&	27.61	  &	0.7014	  &	0.1593	&	26.15	  &	0.6679	  &	0.1938	&	24.38	  &	0.6251	  &	0.2437	\\ 
\midrule
\textbf{Ours} & \textbf{30.33}   & \textbf{0.8157}   & \cellcolor{green!15}\textbf{0.1130}  
              & \cellcolor{green!15}\textbf{30.01}   & \cellcolor{green!15}\textbf{0.8016}   & \cellcolor{green!15}\textbf{0.1238} 
              & \cellcolor{green!15}\textbf{29.53}   & \cellcolor{green!15}\textbf{0.7800}   & \cellcolor{green!15}\textbf{0.1412} 
              & \cellcolor{green!15}\textbf{28.66}   & \cellcolor{green!15}\textbf{0.7463}   & \cellcolor{green!15}\textbf{0.1761} \\ 
\end{tabular}

\begin{tabular}{lccc|ccc|ccc|ccc}
\toprule 
    \rowcolor{color3} {\textbf{Poisson noise}} 
    & \multicolumn{3}{c}{\textbf{$\alpha=2$}} 
    & \multicolumn{3}{c}{\textbf{$\alpha=2.5$}} 
    & \multicolumn{3}{c}{\textbf{$\alpha=3$}} 
    & \multicolumn{3}{c}{\textbf{$\alpha=3.5$}}     \\ 
    \rowcolor{color3} \textbf{Method}        & PSNR    & SSIM     & LPIPS   & PSNR   & SSIM   & LPIPS  & PSNR   & SSIM   & LPIPS  & PSNR   & SSIM   & LPIPS \\ 
\midrule
DnCNN~\cite{zhang2017beyond}         	&	29.13	  &	0.7771	  &	0.1772	&	25.40	  &	0.6740	  &	0.2915	&	22.78	  &	0.5910	  &	0.3972	&	20.86	  &	0.5261	  &	0.4846	\\
RIDNet~\cite{anwar2019real}          	&	29.00	  &	0.7706	  &	0.1681	&	25.17	  &	0.6636	  &	0.2838	&	22.59	  &	0.5836	  &	0.3877	&	20.76	  &	0.5227	  &	0.4730	\\
RNAN~\cite{zhang2019residual}        	&	28.13	  &	0.7488	  &	0.1760	&	24.58	  &	0.6476	  &	0.2897	&	22.18	  &	0.5710	  &	0.3916	&	20.44	  &	0.5119	  &	0.4765	\\
SwinIR~\cite{liang2021swinir}        	&	27.85	  &	0.7419	  &	0.1468	&	24.48	  &	0.6459	  &	0.2472	&	22.12	  &	0.5710	  &	0.3419	&	20.35	  &	0.5122	  &	0.4229	\\
Restormer~\cite{zamir2022restormer}  	&	28.74	  &	0.7765	  &	0.1310	&	25.78	  &	0.6936	  &	0.2082	&	23.57	  &	0.6296	  &	0.2778	&	21.94	  &	0.5792	  &	0.3342	\\
Dropout~\cite{kong2022reflash}       	&	27.74	  &	0.7699	  &	0.1649	&	25.56	  &	0.6751	  &	0.2645	&	23.84	  &	0.5986	  &	0.3558	&	22.47	  &	0.5377	  &	0.4355	\\
baseline                             	&	27.89	  &	0.7024	  &	0.1557	&	24.51	  &	0.6025	  &	0.2522	&	22.19	  &	0.5361	  &	0.3427	&	20.49	  &	0.4761	  &	0.4207	\\ 
\midrule
\textbf{Ours} & \cellcolor{green!15}\textbf{30.01}   & \cellcolor{green!15}\textbf{0.8016}   & \cellcolor{green!15}\textbf{0.1120}  
              & \cellcolor{green!15}\textbf{28.67}   & \cellcolor{green!15}\textbf{0.7439}   & \cellcolor{green!15}\textbf{0.1683} 
              & \cellcolor{green!15}\textbf{27.23}   & \cellcolor{green!15}\textbf{0.6876}   & \cellcolor{green!15}\textbf{0.2329} 
              & \cellcolor{green!15}\textbf{25.99}   & \cellcolor{green!15}\textbf{0.6347}   & \cellcolor{green!15}\textbf{0.2976} \\ 
\end{tabular}

\begin{tabular}{lccc|ccc|ccc|ccc}
\toprule 
    \rowcolor{color3} {\textbf{{\fontsize{6pt}{0}\selectfont {Spatially-correlated}}}}  
    & \multicolumn{3}{c}{\textbf{$\sigma=40$}} 
    & \multicolumn{3}{c}{\textbf{$\sigma=45$}} 
    & \multicolumn{3}{c}{\textbf{$\sigma=50$}} 
    & \multicolumn{3}{c}{\textbf{$\sigma=55$}}     \\ 
    \rowcolor{color3} \textbf{Method} & PSNR & SSIM & LPIPS & PSNR & SSIM & LPIPS & PSNR & SSIM & LPIPS & PSNR & SSIM & LPIPS \\ 
\midrule
DnCNN~\cite{zhang2017beyond}         	&	\cellcolor{green!15}29.92 & \cellcolor{green!15}0.8159	  &	0.2221	&	\cellcolor{green!15}28.59	  &	0.7672	  &	0.2718	&	27.35	  &	0.7160	  &	0.3197	&	26.23	  &	0.6665	  &	0.3654	\\
RIDNet~\cite{anwar2019real}          	&	29.36	  &	0.7958	  &	0.2608	&	28.06	  &	0.7433	  &	0.3146	&	26.90	  &	0.6910	  &	0.3624	&	25.85	  &	0.6426	  &	0.4056	\\
RNAN~\cite{zhang2019residual}        	&	29.16	  &	0.7792	  &	0.2542	&	27.85	  &	0.7257	  &	0.3053	&	26.70	  &	0.6751	  &	0.3514	&	25.68	  &	0.6286	  &	0.3941	\\
SwinIR~\cite{liang2021swinir}        	&	29.10	  &	0.7710	  &	0.2498	&	27.77	  &	0.7165	  &	0.3005	&	26.61	  &	0.6658	  &	0.3446	&	25.59	  &	0.6193	  &	0.3876	\\
Restormer~\cite{zamir2022restormer}  	&	24.46	  &	0.6408	  &	0.2867	&	23.90	  &	0.6043	  &	0.3217	&	23.48	  &	0.5723	  &	0.3542	&	23.18	  &	0.5431	  &	0.3874	\\
Dropout~\cite{kong2022reflash}       	&	28.15	  &	0.7946	  &	0.2123	&	27.32	  &	0.7542	  &	0.2562	&	26.47	  &	0.7097	  &	0.3021	&	25.65	  &	0.6649	  &	0.3493	\\
baseline                             	&	29.43	  &	0.7731	  &	0.2365	&	28.05	  &	0.7191	  &	0.289	&	26.61	  &	0.6532	  &	0.3513	&	25.82	  &	0.6223	  &	0.3770	\\ 
\midrule
\textbf{Ours} & \textbf{28.96}   & \textbf{0.7996}   & \cellcolor{green!15}\textbf{0.1952}  
              & \textbf{28.36}   & \cellcolor{green!15}\textbf{0.7779}   & \cellcolor{green!15}\textbf{0.2216} 
              & \cellcolor{green!15}\textbf{27.65}   & \cellcolor{green!15}\textbf{0.7529}   & \cellcolor{green!15}\textbf{0.2507} 
              & \cellcolor{green!15}\textbf{27.01}   & \cellcolor{green!15}\textbf{0.7251}   & \cellcolor{green!15}\textbf{0.2827} \\ 
\end{tabular}

\begin{tabular}{lccc|ccc|ccc|ccc}
\toprule 
    \rowcolor{color3} \textbf{Salt \& pepper}
    & \multicolumn{3}{c}{\textbf{$d=0.002$}}
    & \multicolumn{3}{c}{\textbf{$d=0.004$}} 
    & \multicolumn{3}{c}{\textbf{$d=0.008$}} 
    & \multicolumn{3}{c}{\textbf{$d=0.012$}}     \\ 
    \rowcolor{color3} \textbf{Method} & PSNR & SSIM & LPIPS & PSNR & SSIM & LPIPS & PSNR & SSIM & LPIPS & PSNR & SSIM & LPIPS \\ 
\midrule
DnCNN~\cite{zhang2017beyond}         	&	23.53	  &	0.6675	  &	0.3607	&	20.13	  &	0.4878	  &	0.5403	&	16.72	  &	0.2966	  &	0.7748	&	14.73	  &	0.2057	  &	0.9320	\\
RIDNet~\cite{anwar2019real}          	&	24.01	  &	0.6639	  &	0.3581	&	20.48	  &	0.4864	  &	0.5288	&	16.93	  &	0.2960	  &	0.7584	&	14.92	  &	0.2065	  &	0.9131	\\
RNAN~\cite{zhang2019residual}        	&	22.62	  &	0.6428	  &	0.3731	&	19.54	  &	0.4651	  &	0.5374	&	16.43	  &	0.2854	  &	0.7626	&	14.59	  &	0.2007	  &	0.9193	\\
SwinIR~\cite{liang2021swinir}        	&	22.68	  &	0.6391	  &	0.3580	&	19.50	  &	0.4581	  &	0.5226	&	16.32	  &	0.2749	  &	0.7379	&	14.47	  &	0.1914	  &	0.8889	\\
Restormer~\cite{zamir2022restormer}  	&	23.04	  &	0.6398	  &	0.3667	&	20.10	  &	0.4829	  &	0.5207	&	18.64	  &	0.3555	  &	0.6163	&	18.34	  &	0.3156	  &	0.6797	\\
Dropout~\cite{kong2022reflash}       	&	25.83	  &	0.6771	  &	0.3082	&	23.04	  &	0.5197	  &	0.4693	&	19.89	  &	0.3536	  &	0.6918	&	17.96	  &	0.2709	  &	0.8487	\\
baseline                             	&	24.06	  &	0.6224	  &	0.3485	&	20.87	  &	0.4630	  &	0.5183	&	17.69	  &	0.2959	  &	0.7378	&	15.86	  &	0.2156	  &	0.8867	\\ 
\midrule
\textbf{Ours} & \cellcolor{green!15}\textbf{29.51}   & \cellcolor{green!15}\textbf{0.7929}   & \cellcolor{green!15}\textbf{0.1504}  
              & \cellcolor{green!15}\textbf{27.45}   & \cellcolor{green!15}\textbf{0.7117}   & \cellcolor{green!15}\textbf{0.2476} 
              & \cellcolor{green!15}\textbf{24.03}   & \cellcolor{green!15}\textbf{0.5508}   & \cellcolor{green!15}\textbf{0.4350} 
              & \cellcolor{green!15}\textbf{21.59}   & \cellcolor{green!15}\textbf{0.4313}   & \cellcolor{green!15}\textbf{0.5968} \\ 
\end{tabular}

\begin{tabular}{lccc|ccc|ccc|ccc}
\toprule 
    \rowcolor{color3} {\textbf{{\fontsize{6pt}{0}\selectfont {Mixture noise}}}}  
    & \multicolumn{3}{c}{\textbf{level 1}} 
    & \multicolumn{3}{c}{\textbf{level 2}} 
    & \multicolumn{3}{c}{\textbf{level 3}} 
    & \multicolumn{3}{c}{\textbf{level 4}}     \\ 
    \rowcolor{color3} \textbf{Method}        & PSNR    & SSIM     & LPIPS   & PSNR   & SSIM   & LPIPS  & PSNR   & SSIM   & LPIPS  & PSNR   & SSIM   & LPIPS \\ 
\midrule
DnCNN~\cite{zhang2017beyond}         	&	28.41 &	0.7627 &	0.1869	&	26.88 &	0.6989 &	0.2406	&	24.16 &	0.5781 &	0.3564	&	22.33 &	0.4877 &	0.4447	\\
RIDNet~\cite{anwar2019real}          	&	28.38 &	0.7509 &	0.1781	&	26.65 &	0.6811 &	0.2337	&	23.82 &	0.5558 &	0.3479	&	22.03 &	0.4659 &	0.4335	\\
RNAN~\cite{zhang2019residual}        	&	27.52 &	0.7285 &	0.1886	&	25.99 &	0.6616 &	0.2414	&	23.42 &	0.5412 &	0.3510	&	21.75 &	0.4533 &	0.4351	\\
SwinIR~\cite{liang2021swinir}        	&	27.57 &	0.7271 &	0.1601	&	26.07 &	0.6619 &	0.2050	&	23.56 &	0.5453 &	0.3059	&	21.86 &	0.4557 &	0.3869	\\
Restormer~\cite{zamir2022restormer}  	&	28.59 &	0.7674 &	0.1410	&	27.53 &	0.7210 &	0.1703	&	25.29 &	0.6263 &	0.2462	&	23.71 &	0.5578 &	0.2991	\\
Dropout~\cite{kong2022reflash}       	&	27.47 &	0.7515 &	0.1694	&	26.41 &	0.6924 &	0.2190	&	24.58 &	0.5856 &	0.3255	&	23.27 &	0.5086 &	0.4079	\\
baseline                             	&	28.05 &	0.7472 &	0.1665	&	26.40 &	0.6810 &	0.2148	&	23.70 &	0.5418 &	0.3229	&	21.91 &	0.4397 &	0.4061	\\ 
\midrule
\textbf{Ours} & \cellcolor{green!15}\textbf{29.91}   & \cellcolor{green!15}\textbf{0.8267}   & \cellcolor{green!15}\textbf{0.1094}  
              & \cellcolor{green!15}\textbf{29.44}   & \cellcolor{green!15}\textbf{0.8111}   & \cellcolor{green!15}\textbf{0.1312} 
              & \cellcolor{green!15}\textbf{28.24}   & \cellcolor{green!15}\textbf{0.7570}   & \cellcolor{green!15}\textbf{0.1870} 
              & \cellcolor{green!15}\textbf{27.15}   & \cellcolor{green!15}\textbf{0.7018}   & \cellcolor{green!15}\textbf{0.2452} \\ 
\bottomrule
\end{tabular}

    \caption{Quantitative comparison on McMaster~\cite{zhang2011color}.}
    \label{tab:mcm}
\end{table*}


\begin{table*}[h]
\centering
\footnotesize 
\label{tab:my-table}

\begin{tabular}{lccc|ccc|ccc|ccc}
\toprule 
    \rowcolor{color3} {\textbf{Speckle noise}} 
    & \multicolumn{3}{c}{\textbf{$\sigma^2=0.02$}}
    & \multicolumn{3}{c}{\textbf{$\sigma^2=0.024$}} 
    & \multicolumn{3}{c}{\textbf{$\sigma^2=0.03$}} 
    & \multicolumn{3}{c}{\textbf{$\sigma^2=0.04$}}     \\ 
    \rowcolor{color3} \textbf{Method}        & PSNR    & SSIM     & LPIPS   & PSNR   & SSIM   & LPIPS  & PSNR   & SSIM   & LPIPS  & PSNR   & SSIM   & LPIPS \\ 
\midrule
DnCNN~\cite{zhang2017beyond}         	&	29.90	  &	0.8380	  &	0.1699	&	28.57	  &	0.8044	  &	0.1982	&	26.90	  &	0.7610	  &	0.2374	&	24.84	  &	0.7035	  &	0.2996	\\
RIDNet~\cite{anwar2019real}          	&	30.11	  &	0.8404	  &	0.1597	&	28.75	  &	0.8044	  &	0.1884	&	27.03	  &	0.7590	  &	0.2305	&	24.87	  &	0.6999	  &	0.2927	\\
RNAN~\cite{zhang2019residual}        	&	29.36	  &	0.8228	  &	0.1593	&	27.95	  &	0.7883	  &	0.1872	&	26.28	  &	0.7451	  &	0.2276	&	24.28	  &	0.6870	  &	0.2893	\\
SwinIR~\cite{liang2021swinir}        	&	28.89	  &	0.8101	  &	0.1602	&	27.55	  &	0.7774	  &	0.1867	&	25.98	  &	0.7362	  &	0.2251	&	24.07	  &	0.6810	  &	0.2849	\\
Restormer~\cite{zamir2022restormer}  	&	29.16	  &	0.8279	  &	0.1518	&	28.13	  &	0.8015	  &	0.1742	&	26.84	  &	0.7667	  &	0.2049	&	25.17	  &	0.7202	  &	0.2523	\\
Dropout~\cite{kong2022reflash}       	&	29.13	  &	0.8447	  &	0.1684	&	28.28	  &	0.8171	  &	0.1953	&	27.16	  &	0.7804	  &	0.2347	&	25.69	  &	0.7311	  &	0.2936	\\
baseline                             	&	29.11	  &	0.8122	  &	0.1794	&	27.75	  &	0.7801	  &	0.2077	&	26.15	  &	0.7393	  &	0.2465	&	24.19	  &	0.6837	  &	0.3050	\\ 
\midrule
\textbf{Ours} & \cellcolor{green!15}\textbf{30.46}   & \cellcolor{green!15}\textbf{0.8777}   & \cellcolor{green!15}\textbf{0.1435}  
              & \cellcolor{green!15}\textbf{30.08}   & \cellcolor{green!15}\textbf{0.8697}   & \cellcolor{green!15}\textbf{0.1511} 
              & \cellcolor{green!15}\textbf{29.49}   & \cellcolor{green!15}\textbf{0.8502}   & \cellcolor{green!15}\textbf{0.1691} 
              & \cellcolor{green!15}\textbf{28.53}   & \cellcolor{green!15}\textbf{0.8169}   & \cellcolor{green!15}\textbf{0.2060} \\ 
\end{tabular}

\begin{tabular}{lccc|ccc|ccc|ccc}
\toprule 
    \rowcolor{color3} {\textbf{Poisson noise}} 
    & \multicolumn{3}{c}{\textbf{$\alpha=2$}} 
    & \multicolumn{3}{c}{\textbf{$\alpha=2.5$}} 
    & \multicolumn{3}{c}{\textbf{$\alpha=3$}} 
    & \multicolumn{3}{c}{\textbf{$\alpha=3.5$}}     \\ 
    \rowcolor{color3} \textbf{Method}        & PSNR    & SSIM     & LPIPS   & PSNR   & SSIM   & LPIPS  & PSNR   & SSIM   & LPIPS  & PSNR   & SSIM   & LPIPS \\ 
\midrule
DnCNN~\cite{zhang2017beyond}         	&	28.13	  &	0.7790	  &	0.1957	&	24.40	  &	0.6417	  &	0.3284	&	21.77	  &	0.5295	  &	0.4524	&	19.83	  &	0.4446	  &	0.5639	\\
RIDNet~\cite{anwar2019real}          	&	28.00	  &	0.7705	  &	0.1878	&	24.08	  &	0.6199	  &	0.3237	&	21.50	  &	0.5082	  &	0.4459	&	19.67	  &	0.4279	  &	0.5542	\\
RNAN~\cite{zhang2019residual}        	&	27.38	  &	0.7505	  &	0.1902	&	23.73	  &	0.6081	  &	0.3201	&	21.29	  &	0.5003	  &	0.4405	&	19.51	  &	0.4220	  &	0.5498	\\
SwinIR~\cite{liang2021swinir}        	&	27.12	  &	0.7392	  &	0.1849	&	23.69	  &	0.6049	  &	0.3094	&	21.27	  &	0.4992	  &	0.4282	&	19.46	  &	0.4200	  &	0.5393	\\
Restormer~\cite{zamir2022restormer}  	&	28.68	  &	0.7973	  &	0.1506	&	25.67	  &	0.6951	  &	0.2361	&	23.54	  &	0.6167	  &	0.3139	&	22.25	  &	0.5598	  &	0.3831	\\
Dropout~\cite{kong2022reflash}       	&	28.03	  &	0.7953	  &	0.1975	&	25.42	  &	0.6823	  &	0.3220	&	23.45	  &	0.5901	  &	0.4366	&	21.94	  &	0.5182	  &	0.5418	\\
baseline                             	&	27.55	  &	0.7517	  &	0.2085	&	23.92	  &	0.6173	  &	0.3346	&	21.42	  &	0.5087	  &	0.4510	&	19.63	  &	0.4259	  &	0.5572	\\ 
\midrule
\textbf{Ours} & \cellcolor{green!15}\textbf{30.01}   & \cellcolor{green!15}\textbf{0.8656}   & \cellcolor{green!15}\textbf{0.1390}  
              & \cellcolor{green!15}\textbf{28.48}   & \cellcolor{green!15}\textbf{0.8053}   & \cellcolor{green!15}\textbf{0.2072} 
              & \cellcolor{green!15}\textbf{26.84}   & \cellcolor{green!15}\textbf{0.7318}   & \cellcolor{green!15}\textbf{0.2974} 
              & \cellcolor{green!15}\textbf{25.33}   & \cellcolor{green!15}\textbf{0.6616}   & \cellcolor{green!15}\textbf{0.3937} \\ 
\end{tabular}

\begin{tabular}{lccc|ccc|ccc|ccc}
\toprule 
    \rowcolor{color3} {\textbf{{\fontsize{6pt}{0}\selectfont {Spatially-correlated}}}}  
    & \multicolumn{3}{c}{\textbf{$\sigma=40$}} 
    & \multicolumn{3}{c}{\textbf{$\sigma=45$}} 
    & \multicolumn{3}{c}{\textbf{$\sigma=50$}} 
    & \multicolumn{3}{c}{\textbf{$\sigma=55$}}     \\ 
    \rowcolor{color3} \textbf{Method} & PSNR & SSIM & LPIPS & PSNR & SSIM & LPIPS & PSNR & SSIM & LPIPS & PSNR & SSIM & LPIPS \\ 
\midrule
DnCNN~\cite{zhang2017beyond}         	&	\cellcolor{green!15}29.38	  &	0.8304	  &	0.2819	&	28.02	  &	0.7839	  &	0.3379	&	26.78	  &	0.7349	  &	0.3864	&	25.68	  &	0.6880	  &	0.4290	\\
RIDNet~\cite{anwar2019real}          	&	28.74	  &	0.8092	  &	0.3306	&	27.45	  &	0.7603	  &	0.3865	&	26.32	  &	0.7122	  &	0.4300	&	25.31	  &	0.6670	  &	0.4672	\\
RNAN~\cite{zhang2019residual}        	&	28.68	  &	0.7983	  &	0.3192	&	27.39	  &	0.7499	  &	0.3703	&	26.25	  &	0.7029	  &	0.4122	&	25.25	  &	0.6591	  &	0.4500	\\
SwinIR~\cite{liang2021swinir}        	&	28.56	  &	0.7883	  &	0.3353	&	27.26	  &	0.7389	  &	0.3853	&	26.13	  &	0.6918	  &	0.4298	&	25.13	  &	0.6484	  &	0.4664	\\
Restormer~\cite{zamir2022restormer}  	&	24.54	  &	0.7076	  &	0.3661	&	24.17	  &	0.6689	  &	0.4007	&	23.70	  &	0.6320	  &	0.4348	&	23.35	  &	0.5978	  &	0.4640	\\
Dropout~\cite{kong2022reflash}       	&	28.89	  &	0.8383	  &	0.2580	&	27.89	  &	0.7999	  &	0.3109	&	26.90	  &	0.7563	  &	0.3656	&	25.96	  &	0.7123	  &	0.4135	\\
baseline                             	&	29.11	  &	0.8109	  &	0.3071	&	27.69	  &	0.7578	  &	0.3658	&	26.48	  &	0.7078	  &	0.4147	&	25.42	  &	0.6625	  &	0.4537	\\ 
\midrule
\textbf{Ours} & \textbf{29.08}   & \cellcolor{green!15}\textbf{0.8445}   & \cellcolor{green!15}\textbf{0.2431}  
              & \cellcolor{green!15}\textbf{28.43}   & \cellcolor{green!15}\textbf{0.8242}   & \cellcolor{green!15}\textbf{0.2765} 
              & \cellcolor{green!15}\textbf{27.71}   & \cellcolor{green!15}\textbf{0.7985}   & \cellcolor{green!15}\textbf{0.3127} 
              & \cellcolor{green!15}\textbf{27.03}   & \cellcolor{green!15}\textbf{0.7719}   & \cellcolor{green!15}\textbf{0.3476} \\ 
\end{tabular}

\begin{tabular}{lccc|ccc|ccc|ccc}
\toprule 
    \rowcolor{color3} \textbf{Salt \& pepper}
    & \multicolumn{3}{c}{\textbf{$d=0.002$}}
    & \multicolumn{3}{c}{\textbf{$d=0.004$}} 
    & \multicolumn{3}{c}{\textbf{$d=0.008$}} 
    & \multicolumn{3}{c}{\textbf{$d=0.012$}}     \\ 
    \rowcolor{color3} \textbf{Method} & PSNR & SSIM & LPIPS & PSNR & SSIM & LPIPS & PSNR & SSIM & LPIPS & PSNR & SSIM & LPIPS \\ 
\midrule
DnCNN~\cite{zhang2017beyond}         	&	24.39	  &	0.7102	  &	0.3205	&	20.88	  &	0.5423	  &	0.5032	&	17.33	  &	0.3499	  &	0.7615	&	15.27	  &	0.2510	  &	0.9304	\\
RIDNet~\cite{anwar2019real}          	&	24.83	  &	0.7065	  &	0.3165	&	21.12	  &	0.5400	  &	0.4912	&	17.44	  &	0.3470	  &	0.7459	&	15.41	  &	0.2510	  &	0.9096	\\
RNAN~\cite{zhang2019residual}        	&	23.32	  &	0.6768	  &	0.3312	&	20.19	  &	0.5127	  &	0.4970	&	16.99	  &	0.3343	  &	0.7464	&	15.12	  &	0.2443	  &	0.9133	\\
SwinIR~\cite{liang2021swinir}        	&	23.21	  &	0.6724	  &	0.3416	&	20.04	  &	0.5035	  &	0.5123	&	16.84	  &	0.3206	  &	0.7541	&	14.97	  &	0.2320	  &	0.9190	\\
Restormer~\cite{zamir2022restormer}  	&	23.58	  &	0.6779	  &	0.3429	&	20.77	  &	0.5292	  &	0.5016	&	19.13	  &	0.4143	  &	0.6322	&	18.37	  &	0.3500	  &	0.7409	\\
Dropout~\cite{kong2022reflash}       	&	26.92	  &	0.7433	  &	0.2739	&	23.97	  &	0.5999	  &	0.4380	&	20.70	  &	0.4330	  &	0.6832	&	18.75	  &	0.3431	  &	0.8508	\\
baseline                             	&	25.09	  &	0.6879	  &	0.3289	&	21.71	  &	0.5261	  &	0.5088	&	18.25	  &	0.3480	  &	0.7621	&	16.30	  &	0.2594	  &	0.9216	\\ 
\midrule
\textbf{Ours} & \cellcolor{green!15}\textbf{29.96}   & \cellcolor{green!15}\textbf{0.8558}   & \cellcolor{green!15}\textbf{0.1512}  
              & \cellcolor{green!15}\textbf{28.01}   & \cellcolor{green!15}\textbf{0.7893}   & \cellcolor{green!15}\textbf{0.2295} 
              & \cellcolor{green!15}\textbf{24.69}   & \cellcolor{green!15}\textbf{0.6391}   & \cellcolor{green!15}\textbf{0.4408} 
              & \cellcolor{green!15}\textbf{22.23}   & \cellcolor{green!15}\textbf{0.5174}   & \cellcolor{green!15}\textbf{0.6331} \\ 
\end{tabular}

\begin{tabular}{lccc|ccc|ccc|ccc}
\toprule 
    \rowcolor{color3} {\textbf{{\fontsize{6pt}{0}\selectfont {Mixture noise}}}}  
    & \multicolumn{3}{c}{\textbf{level 1}} 
    & \multicolumn{3}{c}{\textbf{level 2}} 
    & \multicolumn{3}{c}{\textbf{level 3}} 
    & \multicolumn{3}{c}{\textbf{level 4}}     \\ 
    \rowcolor{color3} \textbf{Method}        & PSNR    & SSIM     & LPIPS   & PSNR   & SSIM   & LPIPS  & PSNR   & SSIM   & LPIPS  & PSNR   & SSIM   & LPIPS \\ 
\midrule
DnCNN~\cite{zhang2017beyond}         	&	27.91	  &	0.7876	  &	0.1955	&	26.28	  &	0.7151	  &	0.2561	&	23.52	  &	0.5791	  &	0.3825	&	21.70	  &	0.4867	  &	0.4833	\\
RIDNet~\cite{anwar2019real}          	&	27.80	  &	0.7740	  &	0.1888	&	25.97	  &	0.6885	  &	0.2510	&	23.14	  &	0.5463	  &	0.3777	&	21.38	  &	0.4589	  &	0.4752	\\
RNAN~\cite{zhang2019residual}        	&	27.16	  &	0.7543	  &	0.1946	&	25.52	  &	0.6718	  &	0.2515	&	22.89	  &	0.5366	  &	0.3711	&	21.22	  &	0.4532	  &	0.4683	\\
SwinIR~\cite{liang2021swinir}        	&	27.10	  &	0.7477	  &	0.1827	&	25.51	  &	0.6668	  &	0.2378	&	22.96	  &	0.5363	  &	0.3563	&	21.29	  &	0.4523	  &	0.4533	\\
Restormer~\cite{zamir2022restormer}  	&	28.54	  &	0.8091	  &	0.1493	&	27.50	  &	0.7625	  &	0.1796	&	25.17	  &	0.6509	  &	0.2599	&	23.52	  &	0.5729	  &	0.3270	\\
Dropout~\cite{kong2022reflash}       	&	28.01	  &	0.8076	  &	0.1841	&	26.78	  &	0.7455	  &	0.2455	&	24.70	  &	0.6296	  &	0.3722	&	23.29	  &	0.5532	  &	0.4672	\\
baseline                             	&	27.81	  &	0.7717	  &	0.2022	&	26.06	  &	0.6916	  &	0.2659	&	23.27	  &	0.5476	  &	0.3927	&	21.48	  &	0.4563	  &	0.4886	\\ 
\midrule
\textbf{Ours} & \cellcolor{green!15}\textbf{29.74}   & \cellcolor{green!15}\textbf{0.8672}   & \cellcolor{green!15}\textbf{0.1342}  
              & \cellcolor{green!15}\textbf{29.14}   & \cellcolor{green!15}\textbf{0.8466}   & \cellcolor{green!15}\textbf{0.1551} 
              & \cellcolor{green!15}\textbf{27.80}   & \cellcolor{green!15}\textbf{0.7900}   & \cellcolor{green!15}\textbf{0.2231} 
              & \cellcolor{green!15}\textbf{26.62}   & \cellcolor{green!15}\textbf{0.7305}   & \cellcolor{green!15}\textbf{0.2964} \\ 
\bottomrule
\end{tabular}

    \caption{Quantitative comparison on CBSD68~\cite{martin2001database}.}
    \label{tab:cbsd68}
\end{table*}


\begin{table*}[h]
\centering
\footnotesize 
\label{tab:my-table}

\begin{tabular}{lccc|ccc|ccc|ccc}
\toprule 
    \rowcolor{color3} {\textbf{Speckle noise}} 
    & \multicolumn{3}{c}{\textbf{$\sigma^2=0.02$}}
    & \multicolumn{3}{c}{\textbf{$\sigma^2=0.024$}} 
    & \multicolumn{3}{c}{\textbf{$\sigma^2=0.03$}} 
    & \multicolumn{3}{c}{\textbf{$\sigma^2=0.04$}}     \\ 
    \rowcolor{color3} \textbf{Method}        & PSNR    & SSIM     & LPIPS   & PSNR   & SSIM   & LPIPS  & PSNR   & SSIM   & LPIPS  & PSNR   & SSIM   & LPIPS \\ 
\midrule
DnCNN~\cite{zhang2017beyond}         	&	28.66 &	0.8207 &	0.1456	&	27.28 &	0.7880 &	0.1745	&	25.64 &	0.7478 &	0.2138	&	23.67 &	0.6962 &	0.2716	\\
RIDNet~\cite{anwar2019real}          	&	28.73 &	0.8218 &	0.1386	&	27.31 &	0.7874 &	0.1683	&	25.63 &	0.7457 &	0.2086	&	23.63 &	0.6933 &	0.2662	\\
RNAN~\cite{zhang2019residual}        	&	27.99 &	0.8047 &	0.1414	&	26.60 &	0.7726 &	0.1697	&	25.01 &	0.7333 &	0.2085	&	23.14 &	0.6826 &	0.2652	\\
SwinIR~\cite{liang2021swinir}        	&	27.50 &	0.7931 &	0.1408	&	26.19 &	0.7626 &	0.1683	&	24.68 &	0.7256 &	0.2059	&	22.88 &	0.6772 &	0.2609	\\
Restormer~\cite{zamir2022restormer}  	&	28.22 &	0.8100 &	0.1370	&	27.17 &	0.7851 &	0.1578	&	25.86 &	0.7529 &	0.1874	&	24.15 &	0.7106 &	0.2302	\\
Dropout~\cite{kong2022reflash}       	&	27.69 &	0.8258 &	0.1516	&	26.83 &	0.7981 &	0.1797	&	25.78 &	0.7639 &	0.2167	&	24.42 &	0.7200 &	0.2693	\\
baseline                             	&	27.66 &	0.7916 &	0.1611	&	26.33 &	0.7617 &	0.1877	&	24.80 &	0.7242 &	0.2241	&	22.98 &	0.6753 &	0.2772	\\ 
\midrule
\textbf{Ours} & \cellcolor{green!15}\textbf{28.97}   & \cellcolor{green!15}\textbf{0.8771}   & \cellcolor{green!15}\textbf{0.1062}  
              & \cellcolor{green!15}\textbf{28.60}   & \cellcolor{green!15}\textbf{0.8642}   & \cellcolor{green!15}\textbf{0.1180} 
              & \cellcolor{green!15}\textbf{28.04}   & \cellcolor{green!15}\textbf{0.8421}   & \cellcolor{green!15}\textbf{0.1421} 
              & \cellcolor{green!15}\textbf{27.12}   & \cellcolor{green!15}\textbf{0.8055}   & \cellcolor{green!15}\textbf{0.1832} \\ 
\end{tabular}

\begin{tabular}{lccc|ccc|ccc|ccc}
\toprule 
    \rowcolor{color3} {\textbf{Poisson noise}} 
    & \multicolumn{3}{c}{\textbf{$\alpha=2$}} 
    & \multicolumn{3}{c}{\textbf{$\alpha=2.5$}} 
    & \multicolumn{3}{c}{\textbf{$\alpha=3$}} 
    & \multicolumn{3}{c}{\textbf{$\alpha=3.5$}}     \\ 
    \rowcolor{color3} \textbf{Method}        & PSNR    & SSIM     & LPIPS   & PSNR   & SSIM   & LPIPS  & PSNR   & SSIM   & LPIPS  & PSNR   & SSIM   & LPIPS \\ 
\midrule
DnCNN~\cite{zhang2017beyond}         	&	27.72 &	0.7814 &	0.1656	&	24.06 &	0.6682 &	0.2738	&	21.52 &	0.5807 &	0.3740	&	19.65 &	0.5128 &	0.4638	\\
RIDNet~\cite{anwar2019real}          	&	27.51 &	0.7728 &	0.1600	&	23.75 &	0.6536 &	0.2697	&	21.27 &	0.5675 &	0.3686	&	19.51 &	0.5025 &	0.4561	\\
RNAN~\cite{zhang2019residual}        	&	26.88 &	0.7550 &	0.1634	&	23.37 &	0.6428 &	0.2682	&	21.02 &	0.5593 &	0.3662	&	19.30 &	0.4953 &	0.4544	\\
SwinIR~\cite{liang2021swinir}        	&	26.59 &	0.7451 &	0.1586	&	23.27 &	0.6392 &	0.2575	&	20.95 &	0.5575 &	0.3533	&	19.21 &	0.4929 &	0.4426	\\
Restormer~\cite{zamir2022restormer}  	&	28.39 &	0.7964 &	0.1326	&	25.34 &	0.7049 &	0.2043	&	22.89 &	0.6266 &	0.2802	&	21.25 &	0.5684 &	0.3524	\\
Dropout~\cite{kong2022reflash}       	&	27.19 &	0.7928 &	0.1722	&	24.82 &	0.6989 &	0.2706	&	22.98 &	0.6269 &	0.3607	&	21.55 &	0.5698 &	0.4437	\\
baseline                             	&	26.94 &	0.7511 &	0.1790	&	23.45 &	0.6425 &	0.2788	&	21.09 &	0.5593 &	0.3712	&	19.40 &	0.4936 &	0.4556	\\ 
\midrule
\textbf{Ours} & \cellcolor{green!15}\textbf{28.72}   & \cellcolor{green!15}\textbf{0.8710}   & \cellcolor{green!15}\textbf{0.1051}  
              & \cellcolor{green!15}\textbf{27.48}   & \cellcolor{green!15}\textbf{0.8142}   & \cellcolor{green!15}\textbf{0.1668} 
              & \cellcolor{green!15}\textbf{26.04}   & \cellcolor{green!15}\textbf{0.7446}   & \cellcolor{green!15}\textbf{0.2464} 
              & \cellcolor{green!15}\textbf{24.71}   & \cellcolor{green!15}\textbf{0.6845}   & \cellcolor{green!15}\textbf{0.3232} \\ 
\end{tabular}

\begin{tabular}{lccc|ccc|ccc|ccc}
\toprule 
    \rowcolor{color3} {\textbf{{\fontsize{6pt}{0}\selectfont {Spatially-correlated}}}}  
    & \multicolumn{3}{c}{\textbf{$\sigma=40$}} 
    & \multicolumn{3}{c}{\textbf{$\sigma=45$}} 
    & \multicolumn{3}{c}{\textbf{$\sigma=50$}} 
    & \multicolumn{3}{c}{\textbf{$\sigma=55$}}     \\ 
    \rowcolor{color3} \textbf{Method} & PSNR & SSIM & LPIPS & PSNR & SSIM & LPIPS & PSNR & SSIM & LPIPS & PSNR & SSIM & LPIPS \\ 
\midrule
DnCNN~\cite{zhang2017beyond}         	&	\cellcolor{green!15}29.87 &	0.8526 &	0.1912	&	\cellcolor{green!15}28.50 &	0.8110 &	0.2371	&	\cellcolor{green!15}27.23 &	0.7677 &	0.2795	&	26.09 &	0.7258 &	0.3173	\\
RIDNet~\cite{anwar2019real}          	&	29.24 &	0.8364 &	0.2216	&	27.89 &	0.7908 &	0.2702	&	26.68 &	0.7464 &	0.3116	&	25.62 &	0.7051 &	0.3464	\\
RNAN~\cite{zhang2019residual}        	&	29.07 &	0.8203 &	0.2248	&	27.72 &	0.7767 &	0.2674	&	26.54 &	0.7351 &	0.3052	&	25.50 &	0.6961 &	0.3385	\\
SwinIR~\cite{liang2021swinir}        	&	28.99 &	0.8116 &	0.2360	&	27.64 &	0.7678 &	0.2769	&	26.46 &	0.7265 &	0.3131	&	25.43 &	0.6882 &	0.3455	\\
Restormer~\cite{zamir2022restormer}  	&	26.38 &	0.7360 &	0.2593	&	25.56 &	0.7011 &	0.2902	&	24.77 &	0.6686 &	0.3189	&	24.06 &	0.6384 &	0.3455	\\
Dropout~\cite{kong2022reflash}       	&	28.68 &	0.8529 &	0.1797	&	27.78 &	0.8191 &	0.2204	&	26.86 &	0.7808 &	0.2635	&	25.96 &	0.7411 &	0.3046	\\
baseline                             	&	29.58 &	0.8440 &	0.2092	&	28.11 &	0.7950 &	0.2567	&	26.84 &	0.7492 &	0.2974	&	25.74 &	0.7076 &	0.3323	\\ 
\midrule
\textbf{Ours} & \textbf{28.06}   & \cellcolor{green!15}\textbf{0.8586}   & \cellcolor{green!15}\textbf{0.1720}  
              & \textbf{27.55}   & \cellcolor{green!15}\textbf{0.8410}   & \cellcolor{green!15}\textbf{0.1976} 
              & \textbf{26.98}   & \cellcolor{green!15}\textbf{0.8196}   & \cellcolor{green!15}\textbf{0.2266} 
              & \cellcolor{green!15}\textbf{26.40}   & \cellcolor{green!15}\textbf{0.7951}   & \cellcolor{green!15}\textbf{0.2562} \\ 
\end{tabular}

\begin{tabular}{lccc|ccc|ccc|ccc}
\toprule 
    \rowcolor{color3} \textbf{Salt \& pepper}
    & \multicolumn{3}{c}{\textbf{$d=0.002$}}
    & \multicolumn{3}{c}{\textbf{$d=0.004$}} 
    & \multicolumn{3}{c}{\textbf{$d=0.008$}} 
    & \multicolumn{3}{c}{\textbf{$d=0.012$}}     \\ 
    \rowcolor{color3} \textbf{Method} & PSNR & SSIM & LPIPS & PSNR & SSIM & LPIPS & PSNR & SSIM & LPIPS & PSNR & SSIM & LPIPS \\ 
\midrule
DnCNN~\cite{zhang2017beyond}         	&	24.01 &	0.7372 &	0.2643	&	20.55 &	0.5828 &	0.4143	&	17.05 &	0.4029 &	0.6335	&	15.01 &	0.3062 &	0.7973	\\
RIDNet~\cite{anwar2019real}          	&	24.56 &	0.7372 &	0.2613	&	20.88 &	0.5835 &	0.4062	&	17.20 &	0.4023 &	0.6220	&	15.16 &	0.3072 &	0.7824	\\
RNAN~\cite{zhang2019residual}        	&	23.01 &	0.7132 &	0.2744	&	19.87 &	0.5582 &	0.4137	&	16.71 &	0.3892 &	0.6223	&	14.86 &	0.2999 &	0.7840	\\
SwinIR~\cite{liang2021swinir}        	&	22.90 &	0.7075 &	0.2823	&	19.74 &	0.5507 &	0.4215	&	16.56 &	0.3790 &	0.6231	&	14.71 &	0.2910 &	0.7773	\\
Restormer~\cite{zamir2022restormer}  	&	23.42 &	0.7145 &	0.2799	&	20.53 &	0.5772 &	0.4086	&	18.65 &	0.4571 &	0.5308	&	17.81 &	0.3967 &	0.6311	\\
Dropout~\cite{kong2022reflash}       	&	26.33 &	0.7591 &	0.2326	&	23.48 &	0.6279 &	0.3647	&	20.29 &	0.4781 &	0.5635	&	18.35 &	0.3943 &	0.7181	\\
baseline                             	&	24.92 &	0.7224 &	0.2667	&	21.56 &	0.5752 &	0.4130	&	18.11 &	0.4103 &	0.6263	&	16.15 &	0.3225 &	0.7840	\\ 
\midrule
\textbf{Ours} & \cellcolor{green!15}\textbf{28.58}   & \cellcolor{green!15}\textbf{0.8655}   & \cellcolor{green!15}\textbf{0.1158}  
              & \cellcolor{green!15}\textbf{26.93}   & \cellcolor{green!15}\textbf{0.8074}   & \cellcolor{green!15}\textbf{0.1850} 
              & \cellcolor{green!15}\textbf{24.01}   & \cellcolor{green!15}\textbf{0.6780}   & \cellcolor{green!15}\textbf{0.3530} 
              & \cellcolor{green!15}\textbf{21.75}   & \cellcolor{green!15}\textbf{0.5652}   & \cellcolor{green!15}\textbf{0.5140} \\ 
\end{tabular}

\begin{tabular}{lccc|ccc|ccc|ccc}
\toprule 
    \rowcolor{color3} {\textbf{{\fontsize{6pt}{0}\selectfont {Mixture noise}}}}  
    & \multicolumn{3}{c}{\textbf{level 1}} 
    & \multicolumn{3}{c}{\textbf{level 2}} 
    & \multicolumn{3}{c}{\textbf{level 3}} 
    & \multicolumn{3}{c}{\textbf{level 4}}     \\ 
    \rowcolor{color3} \textbf{Method}        & PSNR    & SSIM     & LPIPS   & PSNR   & SSIM   & LPIPS  & PSNR   & SSIM   & LPIPS  & PSNR   & SSIM   & LPIPS \\ 
\midrule
DnCNN~\cite{zhang2017beyond}         	&	27.62 &	0.7842 &	0.1656	&	26.08 &	0.7221 &	0.2120	&	23.41 &	0.6112 &	0.3116	&	21.64 &	0.5332 &	0.3907	\\
RIDNet~\cite{anwar2019real}          	&	27.51 &	0.7725 &	0.1592	&	25.75 &	0.7011 &	0.2076	&	23.01 &	0.5844 &	0.3080	&	21.31 &	0.5099 &	0.3851	\\
RNAN~\cite{zhang2019residual}        	&	26.85 &	0.7535 &	0.1651	&	25.28 &	0.6866 &	0.2092	&	22.75 &	0.5759 &	0.3046	&	21.13 &	0.5041 &	0.3813	\\
SwinIR~\cite{liang2021swinir}        	&	26.79 &	0.7475 &	0.1566	&	25.26 &	0.6816 &	0.1973	&	22.81 &	0.5751 &	0.2878	&	21.19 &	0.5040 &	0.3634	\\
Restormer~\cite{zamir2022restormer}  	&	28.45 &	0.8085 &	0.1269	&	27.39 &	0.7665 &	0.1517	&	25.03 &	0.6716 &	0.2171	&	23.26 &	0.5984 &	0.2749	\\
Dropout~\cite{kong2022reflash}       	&	27.22 &	0.7976 &	0.1608	&	26.11 &	0.7431 &	0.2077	&	24.22 &	0.6484 &	0.3035	&	22.91 &	0.5849 &	0.3770	\\
baseline                             	&	27.47 &	0.7795 &	0.1718	&	25.79 &	0.7136 &	0.2191	&	23.12 &	0.5931 &	0.3170	&	21.38 &	0.5131 &	0.3925	\\ 
\midrule
\textbf{Ours} & \cellcolor{green!15}\textbf{28.57}   & \cellcolor{green!15}\textbf{0.8749}   & \cellcolor{green!15}\textbf{0.0995}  
              & \cellcolor{green!15}\textbf{28.08}   & \cellcolor{green!15}\textbf{0.8566}   & \cellcolor{green!15}\textbf{0.1186} 
              & \cellcolor{green!15}\textbf{26.97}   & \cellcolor{green!15}\textbf{0.8053}   & \cellcolor{green!15}\textbf{0.1747} 
              & \cellcolor{green!15}\textbf{25.97}   & \cellcolor{green!15}\textbf{0.7516}   & \cellcolor{green!15}\textbf{0.2337} \\ 
\bottomrule
\end{tabular}

    \caption{Quantitative comparison on Urban100~\cite{huang2015single}.}
    \label{tab:urban100}
\end{table*}

\end{document}